%% file: main.tex
\documentclass[nohyperref]{article}

\usepackage{microtype}
\usepackage{graphicx}
\usepackage{booktabs} % for professional tables

\usepackage{hyperref}

\usepackage[accepted]{icml2022}

\usepackage{amsmath}
\usepackage{amssymb}
\usepackage{mathtools}
\usepackage{amsthm}

\usepackage[capitalize,noabbrev]{cleveref}

\theoremstyle{plain}

\theoremstyle{definition}

\theoremstyle{remark}

\usepackage[textsize=tiny]{todonotes}

\icmltitlerunning{On the Practicality of Deterministic Epistemic Uncertainty}

\input{custom/custom_packages}
\input{custom/custom_commands}
\input{custom/tikz_styles}

\begin{document}

\twocolumn[
\icmltitle{On the Practicality of Deterministic Epistemic Uncertainty}



\icmlsetsymbol{equal}{*}

\begin{icmlauthorlist}
\icmlauthor{Janis Postels}{equal,xxx}
\icmlauthor{Mattia Segu}{equal,xxx}
\icmlauthor{Tao Sun}{xxx}
\icmlauthor{Luca Sieber}{xxx}
\icmlauthor{Luc Van Gool}{xxx}
\icmlauthor{Fisher Yu}{xxx}
\icmlauthor{Federico Tombari}{yyy,zzz}
\end{icmlauthorlist}

\icmlaffiliation{xxx}{ETH Zurich}
\icmlaffiliation{yyy}{Technical University Munich}
\icmlaffiliation{zzz}{Google}

\icmlcorrespondingauthor{Janis, Postels}{jpostels@ethz.ch}
\icmlcorrespondingauthor{Mattia, Segu}{segum@ethz.ch}

\icmlkeywords{Machine Learning, ICML}

\vskip 0.3in
]



\printAffiliationsAndNotice{\icmlEqualContribution} % otherwise use the standard text.

\input{sections/0_abstract}

\section{Introduction} \label{sec:introduction}

\input{sections/1_introduction}

\section{Related Work} \label{sec:related_work}

\input{sections/2_related_work}

\section{Taxonomy for Deterministic Uncertainty Quantification} \label{sec:taxonomy}

\input{sections/3_taxonomy}

\section{Evaluation of Deterministic Epistemic Uncertainty}\label{sec:experiments} 

\input{sections/4_experiments}

\section{Conclusion \& Discussion} \label{sec:conclusion}

\input{sections/5_conclusion}

\section{Acknowledgement}

\input{sections/6_acknowledgement}

\clearpage
\bibliography{bibliography}
\bibliographystyle{icml2022}

%%%%%%%%%%%%%%%%%%%%%%%%%%%%%%%%%%%%%%%%%%%%%%%%%%%%%%%%%%%%%%%%%%%%%%%%%%%%%%%
%%%%%%%%%%%%%%%%%%%%%%%%%%%%%%%%%%%%%%%%%%%%%%%%%%%%%%%%%%%%%%%%%%%%%%%%%%%%%%%
% APPENDIX
%%%%%%%%%%%%%%%%%%%%%%%%%%%%%%%%%%%%%%%%%%%%%%%%%%%%%%%%%%%%%%%%%%%%%%%%%%%%%%%
%%%%%%%%%%%%%%%%%%%%%%%%%%%%%%%%%%%%%%%%%%%%%%%%%%%%%%%%%%%%%%%%%%%%%%%%%%%%%%%
\newpage
\appendix
\onecolumn
\section{Appendix}
\input{sections/7_supplement}
%%%%%%%%%%%%%%%%%%%%%%%%%%%%%%%%%%%%%%%%%%%%%%%%%%%%%%%%%%%%%%%%%%%%%%%%%%%%%%%
%%%%%%%%%%%%%%%%%%%%%%%%%%%%%%%%%%%%%%%%%%%%%%%%%%%%%%%%%%%%%%%%%%%%%%%%%%%%%%%

\end{document}

%% file: custom/custom_packages.tex
\usepackage{acro}
\usepackage{float}
\usepackage{graphicx}
\usepackage[caption=false]{subfig}
\usepackage{amsmath}
\usepackage{amssymb}
\usepackage{pifont}
\usepackage{multirow}
\usepackage{gensymb}

\usepackage[utf8]{inputenc}
\usepackage{pgfplots}
\usepackage{tikz}
\DeclareUnicodeCharacter{2212}{−}
\usepgfplotslibrary{groupplots,dateplot}
\usetikzlibrary{patterns,shapes.arrows}
\pgfplotsset{compat=newest}

%% file: custom/custom_commands.tex
\def\eg{\textit{e.g.}}
\def\ie{\textit{i.e.}}
\def\etal{\textit{et al.}}

\DeclareAcronym{mdl}{
  short = MDL ,
  long = minimum description length
}
\DeclareAcronym{mlp}{
  short = MLP ,
  long = multilayer perceptron
}
\DeclareAcronym{ood}{
  short = OOD ,
  long = out-of-distribution
}
\DeclareAcronym{id}{
  short = ID ,
  long = in-distribution
}
\DeclareAcronym{nn}{
  short = NN ,
  long = neural network
}
\DeclareAcronym{dl}{
  short = DL ,
  long = deep learning
}
\DeclareAcronym{bdl}{
  short = BDL ,
  long = Bayesian Deep Learning
}
\DeclareAcronym{gmm}{
  short = GMM ,
  long = Gaussian mixture model
}
\DeclareAcronym{vi}{
  short = VI ,
  long = Variational Inference
}
\DeclareAcronym{ml}{
  short = ML,
  long = machine learning
}
\DeclareAcronym{cnf}{
  short = CNF,
  long = conditional normalizing flow
}
\DeclareAcronym{rbf}{
  short = RBF,
  long = Radial Basis Function
}
\DeclareAcronym{inn}{
  short = INN,
  long = Invertible Neural Networks
}
\DeclareAcronym{pca}{
  short = PCA,
  long = principal component analysis
}
\DeclareAcronym{nfs}{
  short = NFs,
  long = Normalizing Flows
}
\DeclareAcronym{nf}{
  short = NF,
  long = Normalizing Flow
}
\DeclareAcronym{gp}{
  short = GP,
  long = Gaussian process,
  long-plural-form = Gaussian processes
}
\DeclareAcronym{dum}{
  short = DUM,
  long = Deterministic Uncertainty Method
}
\DeclareAcronym{mc}{
  short = MC,
  long = Monte-Carlo
}
\DeclareAcronym{raulc}{
  short = rAULC,
  long = Relative Area Under the Lift Curve
}
\DeclareAcronym{aulc}{
  short = AULC,
  long = Area Under the Lift Curve
}
\DeclareAcronym{auroc}{
  short = AUROC,
  long = Area Under the Receiver Operating Characteristic
}
\DeclareAcronym{aupr}{
  short = AUPR,
  long = Area Under the Precision-Recall curve
}
\DeclareAcronym{drn}{
  short = DRN,
  long = Dilated ResNet
}
\DeclareAcronym{mir}{
  short = MIR,
  long = Maximally Informative Representations
}
\DeclareAcronym{sn}{
  short = SN,
  long = Spectral Normalization
}
\DeclareAcronym{sngp}{
  short = SNGP,
  long = Spectral-normalized Neural Gaussian Process
}
\DeclareAcronym{duq}{
  short = DUQ,
  long = Deterministic Uncertainty Quantification
}
\DeclareAcronym{bnn}{
  short = BNN,
  long = Bayesian Neural Network
}
\DeclareAcronym{miou}{
  short = mIoU,
  long = mean Intersection over Union
}
\DeclareAcronym{ece}{
  short = ECE,
  long = Expected Calibration Error
}

\newcommand{\bfx}{\mathbf{x}}
\newcommand{\bfy}{\mathbf{y}}

%% file: custom/tikz_styles.tex
\newcommand{\scatterPlotWidthMainPaper}{0.26}
\newcommand{\scatterPlotHeightMainPaper}{0.22}
\newcommand{\scatterPlotWidth}{0.34}
\newcommand{\scatterPlotHeight}{0.25}
\newcommand{\colorbarWidth}{1.4}  % in mm
\tikzset{font={\fontsize{7pt}{12}\selectfont}}

\newenvironment{customlegend}[1][]{%
        \begingroup
        \csname pgfplots@init@cleared@structures\endcsname
        \pgfplotsset{#1}%
    }{%
        \csname pgfplots@createlegend\endcsname
        \endgroup
 }%
\def\addlegendimage{\csname pgfplots@addlegendimage\endcsname}

\tikzset{mark options={mark size=5, line width=3pt}}

\pgfplotsset{
    compat=1.15,
    scatter plots/.style={
        mark options={mark size=5, line width=3pt},
        width=0.1\textwidth,
        height=0.1\textwidth,
    },
    corrupted cifar plots/.style={
        y grid style={white!69.0196078431373!black},
        ytick style={color=black},
        width=0.21\linewidth,
        height=0.18\linewidth,
        legend style={font=\small},
        tick label style={font=\tiny},
        yticklabel style={rotate=90},
        tick align=outside,
        title style={yshift=-1.0ex},
        tick pos=left,
    },
    hyperparameters sensitivity plots/.style={
        yticklabel style={rotate=90},
        ytick style={color=black},
        tick label style={font=\tiny},
        title style={yshift=-1.5ex},
        x grid style={white!69.0196078431373!black},
        xlabel={rAULC},
        xtick style={color=black},
        y grid style={white!69.0196078431373!black},
        colorbar/width=\colorbarWidth mm,
        width=\scatterPlotWidthMainPaper\linewidth,
        height=\scatterPlotHeightMainPaper\linewidth,
        point meta max=8,
        point meta min=1,
        tick align=outside,
        tick pos=left,
    },
    corrupted cityscapes plots/.style={
        ytick style={color=black},
        width=0.35\linewidth,
        height=0.19\linewidth,
        title style={yshift=-1.5ex},
        xmin=-0.25, xmax=5.25,
        xtick style={color=black},
        y grid style={white!69.0196078431373!black},
        x grid style={white!69.0196078431373!black},
        legend cell align={left},
        legend style={fill opacity=0.8, draw opacity=1, text opacity=1, draw=white!80!black},
        tick align=outside,
        tick pos=left,
    },
    carla plots/.style={
        width=0.35\linewidth,
        height=0.19\linewidth,
    },
}

%% file: sections/0_abstract.tex
\begin{abstract}
  A set of novel approaches for estimating epistemic uncertainty in deep neural networks with a single forward pass has recently emerged as a valid alternative to Bayesian Neural Networks. On the premise of informative representations, these deterministic uncertainty methods (DUMs) achieve strong performance on detecting out-of-distribution (OOD) data while adding negligible computational costs at inference time. However, it remains unclear whether DUMs are well calibrated and can seamlessly scale to real-world applications - both prerequisites for their practical deployment.
  To this end, we first provide a taxonomy of DUMs, and evaluate their calibration under continuous distributional shifts.
  % their performance on OOD detection for image classification tasks. 
  %
  Then, we extend them to semantic segmentation. 
  We find that, while DUMs scale to realistic vision tasks and perform well on OOD detection, the practicality of current methods is undermined by poor calibration under distributional shifts.
\end{abstract}

%% file: sections/1_introduction.tex
Despite the dramatic enhancement of predictive performance of \ac{dl}, its adoption remains limited due to unpredictable failure on \ac{ood} samples~\cite{blanchard2011generalizing,muandet2013domain} and adversarial attacks~\cite{szegedy2013intriguing}.
Uncertainty estimation techniques aim at bridging this gap by providing accurate confidence levels on a model's output, allowing for a safe deployment of \acp{nn} in safety-critical tasks, \eg{} autonomous driving or medical applications.
% , and can also foster effective policies for exploration in reinforcement learning.

While \acp{bnn} represent the predominant holistic solution for quantifying uncertainty~\cite{hinton1993keeping,neal2012bayesian}, exactly modelling their full posterior is often intractable, and scalable versions usually require expensive variational approximations~\cite{kingma2015variational,gal2016dropout,zhang2018noisy,postels2019sampling,loquercio2020general}.
Moreover, it has recently been shown that true Bayes posterior can also lead to poor uncertainty~\cite{wenzel2020good}.
Thus, efficient approaches to uncertainty estimation largely remain an open problem, limiting the adoption within real-time applications under strict memory, time and safety requirements.

Recently, a promising line of work emerged for estimating epistemic uncertainty of a \ac{nn} with a single forward pass while treating its weights deterministically.
By regularizing the hidden representations of a model, these methods represent an efficient and scalable solution to epistemic uncertainty estimation and to the related \ac{ood} detection problem.
In contrast to \acp{bnn}, \acp{dum} quantify epistemic uncertainty using the distribution of latent representations~\cite{alemi2018uncertainty, wu2020simple, charpentier2020posterior, mukhoti2021deterministic, postels2020hidden, charpentier2021natural} or by replacing the final softmax layer with a distance-sensitive function~\cite{mandelbaum2017distance, van2020uncertainty, liu2020simple, van2021improving}.
Further, these methods have been applied to practical problems such as object detection~\cite{gasperini2021certainnet}.
While \ac{ood} detection is a prerequisite for a safe deployment of \ac{dl} in previously unseen scenarios, the calibration - \ie{} how well uncertainty correlates with model performance - of such methods under continuous distributional shifts is equally important.
Measuring the calibration of an epistemic uncertainty estimate on shifted data investigates whether it entails information about the predictive performance of the model. This an essential requirement for uncertainty and, unlike OOD detection, an evaluation that is not model-agnostic - i.e. one cannot perform well without taking the predictive model into account.
Nonetheless, previous work falls short of investigating calibration, and solely focuses on \ac{ood} detection~\cite{mandelbaum2017distance,van2020uncertainty,wu2020simple,van2021improving, mukhoti2021deterministic}. 
Further, \acp{dum} have thus far only been evaluated on toy datasets for binary classification, small-scale image classification tasks~\cite{van2020uncertainty, mukhoti2021deterministic} and toy prediction problems in natural language processing~\cite{liu2020simple}.
Despite claiming to solve practical issues of traditional uncertainty estimation approaches, the practicality of \acp{dum} remains to be assessed on more challenging tasks.

% highlight that DUMs are usually evaluated on ood detection, but for practical applications it's arguably more important to assess their calibration (i.e. whether uncertainty is a good indicator of errors)
%

%This work investigates the most crucial questions on the safety of \acp{dum} in practical applications and addresses their shortcomings. 
This work investigates  whether recently proposed \acp{dum} are a realistic alternative for practical uncertainty estimation.
In particular:
(i) we provide the first comprehensive taxonomy of \acp{dum};
(ii) we analyze the calibration of \acp{dum} under synthetic and realistic continuous distributional shifts;
(iii) we evaluate the 
% quality of \acp{dum} for \ac{ood} detection and 
sensitivity of \acp{dum} to their regularization strength;
(iv) we scale \acp{dum} to dense prediction tasks, \eg{} semantic segmentation.
Overall, we find that the practicality of many \acp{dum} is undermined by their poor calibration under both synthetic and realistic distributional shifts. Moreover, some techniques for regularizing hidden representations demonstrate only weak correlation with \ac{ood} detection and calibration performance. 

% In particular:
% (i) to the best of our knowledge, we are the first to analyze the calibration of \acp{dum} under continuous distributional shifts;%
% (ii) we evaluate the quality of \acp{dum} for \ac{ood} detection and their sensitivity to regularization strength;
% (iii) we scale promising deterministic uncertainty methods to dense prediction tasks, \eg{} semantic segmentation, and evaluate them on sequences collected in the CARLA simulator~\cite{Dosovitskiy17} under realistic continuous distributional shift;
% (iv) we show how the practicality of current \acp{dum} is undermined by their poor calibration under realistic distributional shifts.

\vspace{-3mm}

%% file: sections/2_related_work.tex
\textbf{Sources of uncertainty.} 
Uncertainty in a model's predictions can arise from two different sources~\cite{der2009aleatory, kendall2017uncertainties}.
While \textit{aleatoric} uncertainty encompasses the noise inherent in the data and is consequently irreducible~\cite{der2009aleatory}, \textit{epistemic} uncertainty quantifies the uncertainty associated with choosing the model parameters based on limited information, and vanishes - in principle - in the limit of infinite data. This work is concerned with estimating epistemic uncertainty.

\noindent\textbf{Properties of epistemic uncertainty.} This work distinguishes two properties of epistemic uncertainty - its performance on detecting \ac{ood} samples and its calibration (i.e. its correlation with model performance under distributional shifts). 
While the latter has been explored in the case of probabilistic approaches to uncertainty estimation~\cite{snoek2019can}, we are the first to investigate the behaviour of \acp{dum} in this scenario. 
Notably,~\cite{gustafsson2020evaluating} evaluates prominent scalable epistemic uncertainty estimates on semantic segmentation. 
However, they investigate calibration only on in-distribution data. Further, although~\cite{liu2020simple} evaluates the calibration of their approach, they do so exclusively on in-distribution data. 
Lastly, one line of work focuses on calibrating \acp{nn} to achieve good \ac{ood} detection~\cite{lee2018training, bates2021testing}. However, this tackles the model-agnostic task of \ac{ood} detection which is only one part of a good epistemic uncertainty estimate.

\noindent\textbf{\acp{bnn}}~\cite{neal2011mcmc,neal2012bayesian} represent a principled way of measuring uncertainty. 
However, their intractable posterior distribution requires approximate inference methods, such as Markov Chain Monte-Carlo~\cite{neal2011mcmc} or \ac{vi}~\cite{hinton1993keeping}. While these methods traditionally struggle with large datasets and architectures, a variety of scalable approaches - often based on \ac{vi} - have recently emerged. 

\noindent\textbf{Deep ensembles}, which typically consist of identical models trained from different initializations,
%The current state-of-the art in uncertainty estimation in deep learning is represented by ensemble models.
%
have been introduced to the deep learning community by Lakshminarayanan \etal~\cite{lakshminarayanan2017simple} and extended by~\cite{wen2020batchensemble,dusenberry2020efficient}. 
% Moreover, deep ensembles are related to \acp{bnn} when viewed through the lense of Bayesian model averaging~\cite{wilson2020bayesian}. 
While deep ensembles are widely regarded as a strong baseline for estimating epistemic uncertainty, they come with high computational as well as memory costs.

\textbf{Efficient approaches.}
%To reduce the computational overload and memory consumption of ensemble models and BNNs, several approaches to estimate uncertainty using just a single model have emerged.
%
%Some approaches exploit \textit{multiple forward passes} through a single model to simulate an ensemble of networks.
%
Recently, approaches based on stochastic regularization have been developed~\cite{kingma2015variational,gal2016dropout,zhang2018noisy,teye2018bayesian, osband2021epistemic}.
By keeping stochasticity at inference time, they estimate uncertainty using multiple forward passes.
Another line of work estimates the posterior distribution using the Laplace-approximation~\cite{ritter2018scalable, lee2020estimating, sharma2021sketching}.
Moreover, efficient ensemble methods were proposed producing predictions using a single model~\cite{rupprecht2017learning, dusenberry2020efficient, wen2020batchensemble, havasi2020training, rame2021mixmo}. 
Despite promising results on large-scale tasks and parameter reduction, these methods still require sampling through the model, which can render them impractical given limited compute.
To estimate uncertainty in real-time and resource-demanding tasks, recent work has focused on providing uncertainty estimates with a \textit{single forward pass}.
One line of work proposes a principled approach for variance propagation in \acp{nn}~\cite{postels2019sampling, haussmann2020sampling, loquercio2020general}. These approaches fundamentally differ from \acp{dum} due to their probabilistic treatment of the parameters.
Notably, another line of work proposes efficient approaches to estimate aleatoric uncertainty \cite{bishop1994mixture, oberdiek2018classification, oh2018modeling}.

Recently, \acp{dum} showed promising results on \ac{ood} detection.
By leveraging distances and densities in the feature space of a \ac{nn}, these methods provide confidence estimates while adding negligible computational cost.
Since they are united in their deterministic treatment of the weights, we term them Deterministic Uncertainty Methods (DUMs).
The next section provides a taxonomy of \acp{dum}.

%% file: sections/3_taxonomy.tex
% mandelbaum2017distance --> DCS (distance-based confidence score)
% alemi2018uncertainty --> VIB
% van2020uncertainty --> DUQ
% liu2020simple --> SNGP
% wu2020simple --> DUM
% van2021improving --> DUE
% mukhoti2021deterministic --> DDU
% postels2020hidden --> MIR
% nalisnick2019hybrid, ardizzone2020training --> invertible models
% TAXONOMY
% How to enforce Bi-Lipschitz regularization
% --> It is possible to reduce feature collapse by enforcing two constraints on the model: sensitivity and smoothness.
% Sensitivity implies that when the input changes the feature representation also changes: this means the model cannot simply collapse feature representations arbitrarily. 
%Smoothness means small changes in the input cannot cause massive shifts in the output. This appears to help optimization, and ensures the feature space accords with the implicit assumptions that for example RBF kernels make about the data.
%
%
% \input{tables/taxonomy}
%
% \input{tables/taxonomy_v1}
% %
% \input{tables/taxonomy_v2}
% %
% \input{tables/taxonomy_v3}
%
\input{tables/taxonomy_v4}
Since \acp{dum} thus far denote a novel and scattered phenomenon in literature, it is necessary to provide an overview of the research landscape. 
Note, that we provide more material on existing \ac{dum} trends identified in this taxonomy (\autoref{ssec:suppl_background_theory}) as well as the strength and weaknesses of each individual \ac{dum} (\autoref{ssec:suppl_method_oriented_perspective}).
Existing \acp{dum} mostly differentiate along two axis.
Firstly, \acp{dum} apply different regularization techniques to equip their representations with the ability to differentiate between \ac{id} and \ac{ood} data (\autoref{ssec:taxonomy_strategy}).
This is important because the primary goal of \acp{dum} is to quantify epistemic uncertainty while treating the weights of a \ac{nn} deterministically in order to avoid sampling at inference time. 
Since epistemic uncertainty is expected to increase on \ac{ood} data, the representations of a \ac{nn} need to be sensitive to the input distribution.
However, discriminative models suffer from the fundamental problem of feature collapse~\cite{van2020uncertainty,mukhoti2021deterministic} which has to be counteracted using appropriate regularization. 
%
% Thus, we firstly categorize \acp{dum} according to the regularization method used to counteract feature collapse (\autoref{ssec:taxonomy_strategy}).
%
Secondly, \acp{dum} use different methods to estimate uncertainty from such regularized representations (\autoref{ssec:taxonomy_density_models}).
%Finally, in \autoref{ssec:taxonomy_density_models} we detail how each method estimates uncertainty from the so-trained neural network.
%
%To facilitate the reader, we synthesize the proposed taxonomy in \autoref{table:taxonomy}.
\autoref{table:taxonomy} shows an overview of the resulting taxonomy.

%%%%%
\textbf{Feature Collapse.} \label{ssec:feature_collapse}
%
%Deterministic uncertainty methods typically rely on density estimation in the hidden space of a feature extractor to capture epistemic uncertainty.
%
%However, applying density models directly to the feature space does not work out of the box~\cite{mandelbaum2017distance}. 
%
Discriminative models can learn to discard a large part of their input information, as exploiting spurious correlations may lead to better performance on the training data distribution~\cite{peters2017elements, segu2020batch}. 
Such invariant representations learned may be blind to distributional shifts, resulting in a collapse of \ac{ood} embeddings to in-distribution features.
This problem is known as \textit{feature collapse}~\cite{van2020uncertainty}, and it makes \ac{ood} detection based on high-level representations impossible.
%
%Ideally, representations of \ac{ood} samples must be separable from in-distribution ones.
%
% 
% While this has proven to be effective on several toy datasets, it is unclear whether it can actually scale to more complicated tasks, \eg{} semantic segmentation.
%

\subsection{Regularization of Representations} \label{ssec:taxonomy_strategy}
We group \acp{dum} according to their approach to mitigating feature collapse.
Currently, there are two main paradigms - distance awareness and informative representations - which we discuss in \autoref{sssec:distance_awareness} and \autoref{sssec:informative_representations}.

\subsubsection{Distance Awareness} \label{sssec:distance_awareness}

Distance-aware representations avoid feature collapse by relating distances between latent representations to distances in the input space. 
Therefore, one constrains the bi-Lipschitz constant, as it enforces a lower and an upper bound to expansion and contraction performed by a model. 
A lower bound enforces that different inputs are mapped to distinct representations and, thus, provides a solution to feature collapse. 
The upper bound enforces smoothness, \ie{} small changes in the input do not result in large changes in the latent space.
%
% More formally, given any pair of inputs $x_1$ and $x_2$ the following lower and upper bounds must hold for the resulting activation of a feature extractor $f_\theta$ with parameters $\theta$: $c_1||x_1-x_2||_I \leq ||f_\theta(x_1)-f_\theta(x_2)||_F \leq c_2||x_1-x_2||_I$.
% %
% $c_1$ and $c_2$ denote respectively the lower and upper bound for the Lipschitz constant, and $||\cdot||_I$ and $||\cdot||_F$ are the chosen metrics in the input and feature space respectively.
%
%
% Mandelbaum \etal~\cite{mandelbaum2017distance} first attempted to enforce sensitive embeddings by proposing two alternative solutions - \ie{} adversarial examples augmentation and supervised distance-based contrastive learning - to increase the distance between pairs of training points with different labels.
%
While there exist other approaches, \eg{}~\cite{obukhov2021spectral}, recent proposals have primarily adopted two methods to impose the bi-Lipschitz constraint. 

The two-sided \textbf{Gradient Penalty} relates changes in the input to changes in feature space by directly constraining the gradient of the input~\cite{van2020uncertainty}. 
Note, that this leads to large computational overhead as it requires differentiation of the gradients of the input with respect to the \ac{nn}'s parameters. 
\textbf{\ac{sn}}~\cite{miyato2018spectral} is a less computationally-demanding alternative. 
\ac{sn} is applicable to residual layers and normalizes the weights $W$ of each layer using their spectral norm $sn(W)$ to constrain the bi-Lipschitz constant.
Various \acp{dum} - SNGP~\cite{liu2020simple}, DUE~\cite{van2021improving} and DDU~\cite{mukhoti2021deterministic} - rely on \ac{sn} to enforce distance-awareness of hidden representations.
More details on gradient penalty and \ac{sn} can be found in the supplement (\autoref{ssec:suppl_regularization_theory}).
%%%%%%%%%%%%%%%%%%

%
%%%%%%%%%%%%%%%%%%%%%%%%%

%
% For a detailed description of gradient penalty and spectral normalization, please refer to the Appendix.
%
Notably, the bi-Lipschitz constraint is defined with respect to a fix distance measure, which can be difficult to choose for high-dimensional data distributions.
For example, \ac{sn}~\cite{van2020uncertainty,liu2020simple,van2021improving} corresponds to the $L_2$ distance.
% which has been shown not to be particularly meaningful for image-based tasks~\cite{gouk2021regularisation}.
%
While \ac{sn} has empirically been found to perform well, it has been suggested~\cite{singla2019bounding} that popular \ac{sn} approximations behave sub-optimally, and their interaction with losses, architecture and optimization is yet to be fully understood~\cite{rosca2020case}. Principled approaches to providing exact singular values in convolutional layers~\cite{sedghi2018singular} result in prohibitive computational complexity. Further, \cite{smith2021can} provides an explanation for effectiveness of \ac{sn} in convolutional residual \acp{nn}. 
%
% \autoref{ssec:classification_results} analyses the impact of regularization strength on \ac{ood} detection.
%
%\autoref{sssec:hyperparameter_sensitivity} analyses the impact of regularization strength on calibration.

%
\subsubsection{Informative Representations} \label{sssec:informative_representations}
While distance-awareness achieves remarkable performance on OOD detection, it does not explicitly preserve sample-specific information.
%but only enforce distance awareness according to a predefined distance metric in the input space. 
%
Thus, depending on the underlying distance metric it may discard useful information about the input or act overly sensitive.
An alternative line of work avoids feature collapse by learning informative representations~\cite{alemi2018uncertainty,wu2020simple,postels2020hidden,nalisnick2019hybrid,ardizzone2018analyzing,ardizzone2020training}, thus forcing discriminative models to preserve information in its hidden representations beyond what is required to solve a task independent of the choice of an underlying distance metric.
Notably, while representations that are aware of distances in the input space are also informative, both categories remain fundamentally different in their approach to feature collapse.
While distance-awareness is based on the choice of a specific distance metric tying together input and latent space, informative representations incentivize a \ac{nn} to store more information about the input using an auxiliary task~\cite{postels2020hidden, wu2020simple} or forbid information loss by construction~\cite{ardizzone2018analyzing,nalisnick2019hybrid,ardizzone2020training}.
%enforce a constraint on the distribution of hidden representations.
%
There are currently four distinct approaches.

%To this end, Alemi \etal~\cite{alemi2018uncertainty} propose to maximize in variational autoencoders the mutual information between the desired output and its representation to preserve each class from collapsing onto another. 
%
%Optimal representation must also reduce the mutual information between an input and its representation to contain uninformative variability.
%
\textbf{Contrastive learning}~\cite{oord2018representation} has emerged as an approach for learning representations that are both informative and discriminative and provably maximize the mutual information with the data distribution.
This is utilized by Wu \etal~\cite{wu2020simple} and Winkens \etal~\cite{winkens2020contrastive}, who apply SimCLR~\cite{chen2020simple} to regularize hidden representations for a discriminative task by using a contrastive loss for pretraining and fine-tuning to force representations to discriminate between individual instances. 
% Subsequently, they estimate the distribution of hidden representations and use the log-likelihood of as an uncertianty proxy.

\textbf{Reconstruction regularization}~\cite{postels2020hidden} (MIR) instead forces the intermediate activations to fully represent the input. 
This is achieved by adding a decoder branch fed with the activations of a given layer to reconstruct the input.
%
% Consequently, the network's activations are not susceptible to feature collapse, since sample- and domain-specific features are retained in the hidden representations.
% 
% Compared to other approaches, MIR is in principle extendable to any architecture and avoids the strict constraints imposed by spectral normalization or gradient penalty.
%

\textbf{Entropy regularization. } PostNet~\cite{charpentier2020posterior} learns the class-conditional distribution of hidden representations end-to-end using a \ac{nf} parameterizing a Dirichlet distribution.
This allows them to enforce informative representations  by implicitly encouraging large entropy of the \ac{nf} during training. We refer to the supplement for details and further explanations (\autoref{ssec:suppl_regularization_informative_representations_theory}).

\textbf{\acp{inn}}~\cite{jacobsen2018revnet,ardizzone2018analyzing,nalisnick2019hybrid,ardizzone2020training}, built via a cascade of invertible layers, cannot discard information except at the final classification stage.
Consequently, the mutual information between input and hidden representation is maximized by construction.
Interestingly, Behrmann \etal~\cite{behrmann2018invertible} showed that a ResNet is invertible if its Lipschitz constant is lower than $1$, meaning that invertible ResNets both possess highly-informative representations and satisfy distance-awareness. However, note that this is not a necessary condition for invertibility, and thus information preservation.
%
% This property makes them an ideal candidate for \ac{ood} detection.
%
%
%%%%%%%%%%%%%%%%%%%%%%%%%%%%%
\subsection{Uncertainty Estimation} \label{ssec:taxonomy_density_models}
There are two directions regarding uncertainty estimation in \acp{dum} - generative and discriminative approaches. 
While generative approaches use the likelihood produced by an explicit generative model of the distribution of hidden representations as a proxy for uncertainty, discriminative methods directly use the predictions based on regularized representations to quantify uncertainty.

\textbf{Generative approaches} estimate the distribution of hidden representations post-training or end-to-end, and use the likelihood as an uncertainty proxy.
Wu \etal~\cite{wu2020simple} propose a method to estimate the distribution in the feature space, where the variance of the distribution is used as a confidence measure.
MIR~\cite{postels2020hidden},  DDU~\cite{mukhoti2021deterministic} and DCU~\cite{winkens2020contrastive} fit a class-conditional GMM to their regularized hidden representations and use the log-likelihood as an epistemic uncertainty proxy.
DEUP~\cite{jain2021deup} uses the log-likelihood of a normalizing flow in combination with an aleatoric uncertainty estimate to predict the generalization error.
A special instance of the generative approaches are \acp{inn} as they directly estimate the training data distribution. The likelihood of the input data is used as a proxy for uncertainty. While this idea is appealing, it can lead to training difficulties, imposes strong constraints on the underlying model and still remains susceptible to \ac{ood} data~\cite{nalisnick2018deep}.
PostNet~\cite{charpentier2020posterior} is a hybrid approach which estimates the distribution of hidden representations of each class using a separate \ac{nf} which is learned in an end-to-end fashion. Its log-likelihoods parameterize a Dirichlet distribution. We categorize PostNet as a generative approach since their epistemic uncertainty is the log-likelihood of the \ac{nf} associated with the predicted class.

\textbf{Discriminative approaches} use the predictive distribution to quantify uncertainty.
Mandelbaum \etal~\cite{mandelbaum2017distance} propose to use a Distance-based Confidence Score (DCS) learning a centroid for each class end-to-end.
Similarly, DUQ~\cite{van2020uncertainty} builds on Radial Basis Function (RBF) networks~\cite{lecun1998gradient} and proposes a novel centroid updating scheme. 
Both estimate uncertainty as the distance between the model output and the closest centroid.
DUMs adopting \ac{sn}~\cite{liu2020simple,van2021improving} (preserving $L_2$ distances) typically replace the softmax layer with \acp{gp} with RBF kernel, extending distance awareness to the output layer.
In particular, SNGP~\cite{liu2020simple} relies on a Laplace approximation of the \ac{gp} based on the random Fourier feature (RFF) expansion of the \ac{gp} posterior~\cite{rasmussen2003gaussian}.
DUE~\cite{van2021improving} uses the inducing point approximation~\cite{titsias2009variational,hensman2015scalable}, incorporating a large number of inducing points without overfitting~\cite{burt2019rates}.
%
%In practice, our experiments show that the RFF approximation allows for much more stable optimization, while still providing an excellent approximation of the posterior.
%
The uncertainty is derived as the Dempster-Shafer metric~\cite{liu2020simple}, resp. the softmax entropy~\cite{van2021improving}.
%

% While some of the uncertainty models are ideal for a specific latent regularization strategy, \eg{} \acp{gp} with RBF kernels benefit from spectral normalized networks~\cite{liu2020simple}, in principle all models can be applied to the feature space encoded with any of the previously introduced techniques.

%
%Preliminary approaches attempted to model densities in the feature space of discriminative models under Gaussian discriminant analysis~\cite{lee2018simple}. Since it does not address feature collapse, this method occasionally fails to separate in-distribution from OOD data. 
%

%

%Alemi \etal~\cite{alemi2018uncertainty} fit a Gaussian Mixture Model (GMM) on the variational information bottleneck (VIB). 
%

%the full posterior over parameter space, allowing to quantify uncertainty in a single forward pass thanks to the modelling of the density of the features~\cite{ardizzone2018analyzing,nalisnick2019hybrid,ardizzone2020training} typically done through normalizing flows.

%% file: tables/taxonomy_v4.tex
\begin{table*}[htbp]
\caption{Taxonomy of \acp{dum}. Methods are grouped according to their regularization (\textbf{Reg.}) of the hidden representations (rows), and their uncertainty estimation method (columns). For reference: DCS~\cite{mandelbaum2017distance}, DUQ~\cite{van2020uncertainty}, SNGP~\cite{liu2020simple}, DUE~\cite{van2021improving}, DDU~\cite{mukhoti2021deterministic}, DCU~\cite{wu2020simple, winkens2020contrastive}, MIR~\cite{postels2020hidden}, Invertible networks\cite{ardizzone2018analyzing,nalisnick2019hybrid, ardizzone2020training}, PostNet\cite{charpentier2020posterior}} \label{table:taxonomy}
  \centering
  \footnotesize
  \setlength{\aboverulesep}{0pt}
  \setlength{\belowrulesep}{0pt}
  \renewcommand{\arraystretch}{1.33}
  \setlength{\tabcolsep}{4pt}
  \begin{tabular}{|c|c|c|c|c|c|}
    \hline
    \multicolumn{2}{|c|}{\multirow{4}{*}{\textbf{\acp{dum}}}} & \multicolumn{4}{|c|}{\textbf{Uncertainty Estimation Method}} \\
    \cline{3-6}
    \multicolumn{2}{|c|}{} & \multicolumn{2}{|c|}{Discriminative} & \multicolumn{2}{|c|}{Generative} \\
    \cline{3-6}
    \multicolumn{2}{|c|}{} & Class centroid & Gaussian Processes & Gaussian Mixture Models & Normalizing Flows \\
    \cline{1-6}
    \parbox[t]{6mm}{\multirow{2}{*}{\rotatebox[origin=c]{90}{\textbf{Reg.}}}}  & \shortstack{Distance awareness}      & DCS, DUQ & SNGP, DUE& DDU &  - \\
    \cline{2-6}
    %  & \multirow{3}{*}{\shortstack{Informative \\ Representations}}   & & & VIB~\cite{alemi2018uncertainty} &  \multirow{3}{*}{\shortstack{Invertible\\Networks\\\cite{ardizzone2018analyzing,nalisnick2019hybrid, ardizzone2020training}}}              \\
     & Informative representations  & - & - & DCU, MIR &  Invertible networks, PostNet\\
    \hline
  \end{tabular}
% \vspace{-4mm}
\end{table*}

%% file: sections/4_experiments.tex
%
%The research effort on deterministic epistemic uncertainty modelling has so far focused on providing excellent confidence measure to detect \ac{ood} samples.
%
%However, the calibration of such epistemic uncertainty estimates is ambiguous close to the training distribution and is typically not assessed.
%
%Since discriminative models are often overconfident about the model's predictions~\cite{guo2017calibration,hein2019relu}, we argue that calibration of confidence measures is an indispensable property for practical use of such methods.

% Furthermore, while most of the proposed methods claim to be easily extendable to more complicated tasks and architectures, this has not been shown in practice. 
%
% In~\autoref{ssec:segmentation}, we successfully propose extended versions of some of these methods that scale to more challenging settings and outperform more popular and less efficient baselines. 
%
% Further, we evaluate how such techniques hold their promises on harder tasks, \eg semantic segmentation,  scaling nicely to the deeper architectures while providing inexpensive uncertainty estimates.
 %In this section, we introduce our setting for a complete and fair evaluation of \acp{dum}. 
 %
 We investigate whether a deterministic treatment of the weights of a \ac{nn} as proposed by \acp{dum} not only detects \ac{ood} well but also yields well calibrated epistemic uncertainty, and scales to realistic vision tasks.
 Therefore, our experiments are comprised of two parts. 
 Firstly, we evaluate \acp{dum} on image classification, where we measure their calibration under synthetic corruptions (\autoref{ssec:classification}) and sensitivity to their regularization strength. We also evaluate \acp{dum} on \ac{ood} detection in \autoref{sssec:suppl_classification_results}.
 Then, we extend \acp{dum} to a large-scale dense prediction task - semantic segmentation (\autoref{ssec:segmentation}) - where we evaluate their calibration on synthetic corruptions (\autoref{ssec:segmentation_synthetic}) based on Cityscapes as well as on more realistic distributional shifts (\autoref{ssec:segmentation_realistic}) based on data collected in the simulation environment CARLA~\cite{Dosovitskiy17}.
%
% In both settings, we analyze their \ac{ood} detection performance and calibration compared to traditional deep uncertainty estimation baselines.

\textbf{Baselines.} We compare \acp{dum} with two baselines for epistemic uncertainty - \ac{mc} dropout~\cite{gal2016dropout} and deep ensembles~\cite{lakshminarayanan2017simple}. Moreover, we report the softmax entropy of a vanilla \ac{nn} as a simple baseline. We refer to the supplement for details on uncertainty estimation in our baselines.
Note, that the softmax entropy is expected to yield suboptimal calibration under distributional shifts since it quantifies aleatoric uncertainty while adding no computational overhead. 

\textbf{Methods.} We evaluate DUQ~\cite{van2020uncertainty}, SNGP~\cite{liu2020simple} and DDU~\cite{mukhoti2021deterministic} as representatives of distance-awareness, since these cover both techniques - \ac{sn} and gradient penalty - and apply different techniques for uncertainty estimation. We exclude DUE~\cite{van2021improving} since it provides limited additional insights given SNGP. Moreover, we exclude DCS~\cite{mandelbaum2017distance} since it only leads to a marginal improvement in their own experiments and their contrastive loss only operates on class centroids and, thus, is not expected to lead to distance awareness within clusters.
Furthermore, we evaluate MIR~\cite{postels2020hidden}, DCU~\cite{winkens2020contrastive} and PostNet~\cite{charpentier2020posterior} as representatives of informative representations. However, we do not scale DCU and PostNet to semantic segmentation. DCU with its constrastive pretraining based on SimCLR~\cite{chen2020simple} is computationally too demanding due to large batch sizes. PostNet does not scale to semantic segmentation due to instabilities arising from the end-to-end training of the \ac{nf} for learning the distribution of hidden representations. They require a small hidden dimension ($\leq$10) which already leads to poor testset performance on CIFAR100.
Further, we do not evaluate methods based on invertible neural networks~\cite{nalisnick2019hybrid, ardizzone2020training} since they 1) enforce strict constraints on the underlying architecture (\eg{} fixed dimensionality of hidden representations) and often lead to training instabilities. 

\textbf{Calibration metrics.}
Typical calibration metrics are \ac{ece}~\cite{naeini2015obtaining} and Brier score~\cite{brier1950verification}. However, since most \acp{dum}, except SNGP, do not provide uncertainty in form of a probabilistic forecast, we cannot rely on measuring the calibration of probabilities. Thus, we will exploit another desired property of uncertainty to quantify calibration, namely the ability to distinguish correct from incorrect predictions. In fact, this property is a relaxation of calibrated probabilities, as it is independent of the absolute value of uncertainty estimates. It solely relies on the ability of an uncertainty estimate to sort predictions according to their correctness.
We assess the calibration of uncertainty estimates under distributional shifts using two metrics. Firstly, we report the \textit{\ac{auroc}} obtained when separating correct and incorrect predictions based on uncertainty. Moreover, we introduce a new metric, \textit{\ac{raulc}}, based on the \ac{aulc}~\cite{vuk2006roc}. The \ac{aulc} is obtained by ordering the predictions according to increasing uncertainty and plotting the performance of all samples with an uncertainty value smaller than a certain quantile of the uncertainty against the quantile itself.

Formally, given a set of uncertainty quantiles $q_i \in [0, 1]$, $i\in [1, ..., S]$, with some quantile step width $0<s<1$
and the function $F(q_i)$ which returns the accuracy of all samples with uncertainty $u < q_i$, the \ac{aulc} is defined as $AULC =  - 1 + \sum_{i\in [1, ..., S]} s \frac{F(q_i)}{F_R(q_i)}$. Here, $F_R(\cdot)$ refers to a baseline uncertainty estimate that corresponds to random guessing. We subtract 1 to shift the performance of the random baseline to zero.
Note, if an uncertainty estimate is anti-correlated with a models' performance, this score can also be negative. 
To alleviate a bias towards better performing models, we further compute the \ac{raulc} by dividing the \ac{aulc} by the \ac{aulc} of a hypothetical (optimal) uncertainty estimation that perfectly orders samples according to model performance. In classification we measure \ac{auroc} and \ac{raulc} on the image-level, in semantic segmentation on the pixel-level. In all experiments we set the quantile step width to $s=\frac{1}{N}$, where N is the number of predictions.

We compute \ac{auroc} and \ac{raulc} on continuous distributional shifts 1) across all severities of distributional shifts (including the clean testset) and 2) for each severity separately. We use the former method to establish a quantitative comparison among the methods and the latter to depict the calibration evolution qualitatively as a function of the distributional shift's severity.

\subsection{Image Classification}\label{ssec:classification}
\input{sections/4_1_image_classification}

\subsection{Semantic Segmentation}\label{ssec:segmentation}
\input{sections/4_2_semantic_segmentation}

%% file: sections/4_1_image_classification.tex
\textbf{Datasets.} 
We train \acp{dum} on CIFAR-10 and CIFAR-100~\cite{krizhevsky2014cifar} and evaluate on the corrupted versions of their test set CIFAR10/100-C~\cite{hendrycks2019robustness} (\autoref{ssec:classification}). These include 15 synthetic corruptions, each with 5 levels of severity.
Moreover, we explore how sensitive the calibration of \acp{dum} is to the choice of regularization strength on MNIST~\cite{lecun1998mnist} and FashionMNIST~\cite{xiao2017fashion}). 

\textbf{Models and optimization.} 
Each method shares the same backbone architecture and uses a method-specific prediction head.
When training on CIFAR-10/100, the backbone architecture is a ResNet-50~\cite{he2016deep} 
For the experiments regarding hyperparameter sensitivity on MNIST and Fashion-MNIST, we employ a \ac{mlp} as feature extractor with $3$ hidden layers of $100$ dimensions each and ReLU activation functions.
Each \ac{dum} has a hyperparameter for the regularization of its hidden representations. We choose the hyperparameter such that it minimizes the validation loss.
All results are averaged over $5$ independent runs. The standard deviation and optimization details can be found in the supplement where not present in the main paper. Moreover, we provide per-sample training and inference runtimes for each method in the supplement.

\subsubsection{Continuous Distributional Shifts}\label{ssec:classification}

\input{tables/cifar_10_100_c}

\autoref{tab:cifar_10_100_c} reports testset accuracy and calibration for the baselines and \acp{dum}. \ac{auroc} and \ac{raulc} are computed for each corruption across all severity levels, then averaged over all corruptions. Further, \autoref{fig:cifar_10_100_c_main_paper} depicts our metrics depending on the severity of corruptions. We generally observe that ensembles and \ac{mc} dropout demonstrate best performance in terms of calibration. Further, SNGP is the only \ac{dum} that consistently outperforms the softmax entropy. Overall we observe that \acp{dum} using the distribution of hidden representations for estimating epistemic uncertainty yield worse calibration. Among these we find that regularizing hidden representations by enforcing distance awareness (DDU) yields the worst calibration. The superior performance of DCU~\cite{winkens2020contrastive} in terms of testset accuracy originates from their extensive contrastive pretraining which includes extensive data augmentation and a prolonged training schedule. We note that DUQ~\cite{van2020uncertainty} did not converge on CIFAR100 due to training instabilities. These arise from maintaining the class centroids, which become very noisy for 100 classes with only 600 samples per class.

Moreover, we report \ac{ood} detection performance of models used in \autoref{tab:cifar_10_100_c} and \autoref{fig:cifar_10_100_c_main_paper} in the supplement. Interestingly, despite competitive performance on \ac{ood} detection \acp{dum} fall short in terms of uncertainty calibration compared to MC dropout and deep ensembles.

\begin{figure*}[!h]
% \centering
\input{figures/cifar_10_100_c_new/gaussian_cifar10}%
\input{figures/cifar_10_100_c_new/motion_blur_cifar10}%
\\
\input{figures/cifar_10_100_c_new/gaussian_cifar100}%
\hspace{1em} \input{figures/cifar_10_100_c_new/motion_blur_cifar100}%
\caption{Softmax entropy, ensembles~\cite{lakshminarayanan2017simple}, MC dropout~\cite{gal2016dropout}, DUQ~\cite{van2020uncertainty}, SNGP~\cite{liu2020simple}, MIR~\cite{postels2020hidden}, DDU~\cite{mukhoti2021deterministic} and DCU~\cite{winkens2020contrastive} on CIFAR10-C (upper row) and CIFAR100-C (lower row)~\cite{hendrycks2019robustness}. We show the accuracy, \ac{auroc} and \ac{raulc} on the corruptions gaussian\_noise (gn) and motion\_blur (mb) against the corruption severity. While all methods, except DCU, demonstrate a similar accuracy, \acp{dum} - in particular methods based on generative modeling of hidden representations - yield worse calibration. DUQ did not converge on CIFAR100. Other corruptions are included in the supplement.}
\label{fig:cifar_10_100_c_main_paper}
% \vspace{-2mm}
\end{figure*}

\subsubsection{Sensitivity to Hyperparameters} \label{sssec:hyperparameter_sensitivity}

We are interested in the impact of the regularization strength on the uncertainty calibration. Therefore, we train  DUQ~\cite{van2020uncertainty}, SNGP~\cite{liu2020simple}, MIR~\cite{postels2020hidden} and DDU~\cite{mukhoti2021deterministic} using various regularization strengths on MNIST and evaluate on continuously shifted data by rotating from 0 to 180 degrees in steps of 20 degrees. \autoref{fig:hyperparameter_sensitivity} depicts the test accuracy against the \ac{raulc} for various regularization strengths. Only for MIR we observe a clear, positive correlation between regularization strength and calibration. Moreover, \autoref{table:quantitative_correlation} reports the corresponding Pearson/Spearman correlation coefficients. The supplement depicts similar results on FashionMNIST~\cite{xiao2017fashion} as well as for \ac{ood} detection performance.

\begin{figure*}
% \centering
\input{figures/image_classification/regularization_strength/main_paper/duq_mnist}%
\input{figures/image_classification/regularization_strength/main_paper/sngp_mnist}%
\input{figures/image_classification/regularization_strength/main_paper/ddu_mnist}%
\input{figures/image_classification/regularization_strength/main_paper/mir_mnist}%
% \vspace{-5mm}
\caption{We analyze the sensitivity of DUQ~\cite{van2020uncertainty}, SNGP~\cite{liu2020simple}, MIR~\cite{postels2020hidden} and DDU~\cite{mukhoti2021deterministic} to their regularization strength. Therefore, we train models on MNIST and evaluate on continuously shifted data by rotating from 0 to 180 degrees in steps of 20 degrees. We plot the accuracy on the unperturbed testset against the \ac{raulc} computed using data from all levels of perturbation. Only for MIR we observe a clear, positive correlation between regularization strength and calibration. For DDU and SNGP, smaller regularization parameter denotes stronger regularization. }
\label{fig:hyperparameter_sensitivity}
% \vspace{-4mm}
\end{figure*}

\input{tables/hyperparameter_sensitivity_correlation}

%% file: tables/cifar_10_100_c.tex
\begin{table*}[!h]
\caption{We compare Softmax, MC Dropout \cite{gal2016dropout}, Deep Ensembles, SNGP, DDU, MIR, DUQ, DCU and PostNet on CIFAR10/100-C. We evaluate the accuracy (ACC) on the uncorrupted testset, \ac{auroc} and \ac{raulc}. Ensembles and MC dropout demonstrate better uncertainty calibration than most \acp{dum}. Only SNGP consistently outperforms the softmax entropy. DCU's superior performance is expected since it uses expensive contrastive pretraining. DUQ did not converge on CIFAR100-C due to training instabilities arising from dynamically updated cluster centroids.}\label{tab:cifar_10_100_c}
\centering
\begin{tabular}{c | c c c | c c c}
    \toprule
            \textbf{Method} & & \textbf{CIFAR10-C} & & &  \textbf{CIFAR100-C} & \\
            & ACC & AUROC & rAULC & ACC & AUROC & rAULC\\
            \midrule
            Softmax entropy & 0.882 & 0.782 & 0.708  & 0.610 & 0.762 & 0.596 \\ 
            MC Dropout~\cite{gal2016dropout} & 0.885 & \textbf{0.866} & 0.829 & 0.615 & 0.818 & \textbf{0.726} \\
            Ensemble~\cite{lakshminarayanan2017simple} & 0.910 & 0.85 0& \textbf{0.833} & 0.628 & \textbf{0.824} & 0.713 \\ 
            \midrule
            SNGP~\cite{liu2020simple} & 0.903 & 0.833 & 0.766 & 0.611 & 0.788 & 0.623 \\ 
            DDU~\cite{mukhoti2021deterministic} & 0.884 & 0.673 & 0.441 & 0.609 & 0.635 & 0.339 \\
            MIR~\cite{postels2020hidden} & 0.889 & 0.79 & 0.697 & 0.617 & 0.726 & 0.514 \\ 
            DUQ~\cite{van2020uncertainty} & 0.860 & 0.773 & 0.614 & - & - & - \\
            DCU~\cite{winkens2020contrastive} & \textbf{0.945} & 0.794 & 0.706 & \textbf{0.642} & 0.750 & 0.558 \\
            PostNet~\cite{charpentier2020posterior} & 0.882 & 0.784 & 0.676 & 0.520 & 0.743 & 0.603 \\
     \bottomrule
\end{tabular}
% \vspace{-4mm}
\end{table*}

%% file: tables/hyperparameter_sensitivity_correlation.tex
\begin{table}[!htb]
\caption{Quantitative correlation between rAULC and regularization strength using Pearson/Spearman's rank correlation coefficient. MIR~\cite{postels2020hidden} demonstrates the largest correlation. }\label{table:quantitative_correlation}
\footnotesize
\begin{center}
\setlength{\tabcolsep}{10pt}
\begin{tabular}{c | c | c | c | c}
        \toprule
        Metric & DUQ & DDU & SNGP & MIR \\
        \hline
        Pearson & -0.83 & -0.30 & 0.58 & 0.76 \\
        Spearman & -0.83 & -0.43 & 0.61 & 0.96 \\
        \bottomrule
\end{tabular}
\end{center}
\end{table}

%% file: sections/4_2_semantic_segmentation.tex
This section evaluates whether \acp{dum} seamlessly scale to realistic vision tasks and compares their behaviour under synthetic and realistic continuous distributional shifts with the softmax entropy, \ac{mc} dropout and ensembles. Therefore, we apply MIR~\cite{postels2020hidden}, SNGP~\cite{liu2020simple} and DDU~\cite{mukhoti2021deterministic} to semantic segmentation. Note that DUQ~\cite{van2020uncertainty} did not converge on this task.

We consider semantic segmentation as a multidimensional classification problem, where each pixel of the output mask represents an independent classification problem. 
Given an image $\bfx$ with $n$ pixels $\bfy = \{y_1, \cdots, y_n\}$, the predictive distribution factorizes according to
$p(\bfy \mid \bfx) = p(y_1 \mid \bfx) p(y_2 \mid \bfx) \cdots p(y_n \mid \bfx)$.
%
% \begin{equation}
%     p(\bfy \mid \bfx) = p(y_1 \mid \bfx) p(y_2 \mid \bfx) \cdots p(y_n \mid \bfx)
% \end{equation}
%
We evaluate the calibration of the pixel-level uncertainty in our experiments.

\input{tables/cityscapes_c_carla}

\textbf{Datasets.} 
We evaluate on synthetic distributional shifts using a corrupted version of Cityscapes~\cite{Cordts2016Cityscapes} (Cityscapes-C~\cite{michaelis2019benchmarking}) which contains the same corruptions as CIFAR10/100-C.
To further benchmark \acp{dum} in a realistically and continuously changing environment, we collect a synthetic dataset for semantic segmentation. 
We use the CARLA Simulator~\cite{Dosovitskiy17} and leverage the SHIFT dataset~\cite{shift2022} toolkit for rendering images and segmentation masks under controlled distributional shifts in a driving scenario. The classes definition is aligned with the CityScape dataset~\cite{Cordts2016Cityscapes}. Training data is collected from four towns in CARLA. We produce 32 sequences from each town. Vehicles and pedestrians are randomly generated for each sequence. Every sequence has 500 frames with a sampling rate of $10$ FPS. We uniformly sample a validation set. We introduce continuous distributional shifts by varying the time-of-the-day and weather conditions (visual examples and details on data collection are in the supplement). The time-of-the-day is parameterized by the sun's altitude angle, where $90^\circ$ means mid-day (training data) and the $0^\circ$ means dust/dawn. We produce samples with altitude angles from $90^\circ$ to $15^\circ$ by steps of $5^\circ$, and $15^\circ$ to $-5^\circ$, where the environment changes sharply, in $1^\circ$ steps. In order to continuously change the weather conditions, we increase the magnitude of the rain in four steps (see supplement for visual examples). We refer to this dataset as CARLA-C.

\textbf{Backbone.} 
We adopt \ac{drn}~\cite{Yu2016, Yu2017} as semantic segmentation backbone since it is based on residual connections allowing the use \ac{sn}.
Using dilated convolutions it improves spatial accuracy, achieving satisfactory results on CityScapes~\cite{Cordts2016Cityscapes}. 
We adopt the variant DRN-A-50.
%
% (ablation in supplement).
%
All results are averaged across 5 independent repetitions.

\textbf{SNGP.} 
\ac{drn} uses $1 \times 1$ convolutions at the last layer to map the latest feature map to the predicted segmentation mask.
This works under the assumption that all pixels in the output mask are i.i.d. random variables.
Following this intuition, we extend SNGP to semantic segmentation by fitting a $GP: \mathbb{R}^Z \rightarrow \mathbb{R}^C$  at pixel level that maps from the deep feature dimension $z$ to the number of classes $c$. By keeping the \ac{gp} kernel parameters shared across all pixels, we simulate a  $1 \times 1$ convolutional \ac{gp}, \ie{} $\sigma: (H_l \times W_l \times Z) \rightarrow (H_l \times W_l \times C)$, where $\sigma$ convolves the \ac{gp}, $H_l$ and $W_l$ are, respectively, feature map height and width at layer $l$, $Z$ is the number of latent features and $C$ is the number of output classes. For details about the \ac{gp} we refer to~\cite{liu2020simple} or the supplement.

\textbf{MIR and DDU} require fitting the distribution of hidden representations. We fit a \ac{gmm} with $20$ components (\ie{} number of classes) to each spatial location of the hidden representations using features extracted from the training data independently. This assumes that the distribution is translation invariant and factorizes along the spatial dimensions of the latent space. Pixel-level uncertainties are then computed using bi-cubic interpolation following a similar procedure as the one proposed in~\cite{blum2019fishyscapes}. We refer to the supplement for more details.

\subsubsection{Cityscapes Corrupted}\label{ssec:segmentation_synthetic}

We evaluate the softmax entropy, ensembles~\cite{lakshminarayanan2017simple}, MC dropout~\cite{gal2016dropout}, SNGP~\cite{liu2020simple}, MIR~\cite{postels2020hidden}, and DDU~\cite{mukhoti2021deterministic} on Cityscapes-C~\cite{michaelis2019benchmarking}. \autoref{tab:cityscapes_c_carla} depicts \ac{miou} and calibration performance in terms of \ac{auroc} and \ac{raulc}. Ensembles and \ac{mc} dropout yield the best calibration, while among \acp{dum} only SNGP consistently ourperforms the softmax entropy. A qualitative visualization of the \ac{auroc} and \ac{raulc} depending on the corruption strength is in the supplement (\autoref{fig:corruptions_cityscapes_c}, \autoref{fig:corruptions_cityscapes_c_2}, and \autoref{fig:corruptions_cityscapes_c_3}).

\subsubsection{Realistic Continuous Distributional Shifts}\label{ssec:segmentation_realistic}

Similarly, we evaluate the softmax entropy, ensembles~\cite{lakshminarayanan2017simple}, MC dropout~\cite{gal2016dropout}, SNGP~\cite{liu2020simple}, MIR~\cite{postels2020hidden}, and DDU~\cite{mukhoti2021deterministic} on CARLA-C. \autoref{tab:cityscapes_c_carla} depicts \ac{miou} and calibration performance in terms of \ac{auroc} and \ac{raulc}. Ensembles yield the best calibration. Among \acp{dum} only SNGP consistently ourperforms the softmax entropy which is in line with the results on image classification (\autoref{ssec:classification}). Further, we show the qualitative behaviour of \ac{auroc} and \ac{raulc} depending on the corruption strength in the supplement (\autoref{fig:corruptions_carla_c}).

%% file: tables/cityscapes_c_carla.tex
\begin{table*}[!h]
\caption{We compare semantic segmentation using Softmax, MC Dropout \cite{gal2016dropout}, Deep Ensembles, SNGP, DDU and MIR on Cityscapes-C and CARLA-C. We evaluate the \ac{miou} on the uncorrupted testset and \ac{auroc}/\ac{raulc} across all levels of corruption. Again, ensembles and MC dropout yield better calibrated uncertainty than most \acp{dum}. Most notably, only SNGP consistently outperforms the softmax entropy. \acp{dum} using an explicit generative model of hidden representations to estimate uncertainty perform particularly bad on realistic distributional shifts (CARLA-C).}\label{tab:cityscapes_c_carla}
\centering
\begin{tabular}{c | c c c | c c c}
    \toprule
            \textbf{Method} & & \textbf{Cityscapes-C} & & & \textbf{CARLA-C} & \\
             & mIoU & AUROC & rAULC & mIoU & AUROC & rAULC\\
            \midrule
            Softmax & 0.503 & 0.815 & 0.737 & 0.422 & 0.854 & 0.818 \\ 
            MC Dropout~\cite{gal2016dropout} & 0.506 & \textbf{0.846} & \textbf{0.785} & 0.410 & 0.843 & 0.730 \\  
            Ensemble~\cite{lakshminarayanan2017simple} & \textbf{0.525} & 0.835 & 0.751 & \textbf{0.428} & \textbf{0.863} & 0.812 \\ 
            \midrule
            SNGP~\cite{liu2020simple} & 0.519 & 0.833 & 0.759 & 0.424 & 0.853 & \textbf{0.813} \\
            DDU~\cite{mukhoti2021deterministic} & 0.505 & 0.731 & 0.542 & 0.408 & 0.467 & -0.038 \\ 
            MIR~\cite{postels2020hidden} & 0.504 & 0.729 & 0.564 & 0.412 & 0.744 & 0.619 \\
     \bottomrule
\end{tabular}
% \vspace{-2mm}
\end{table*}

%% file: sections/5_conclusion.tex
This work investigates the shortcomings of a recent trend in research on deterministic epistemic uncertainty estimation. To this end, we provide the first taxonomy of \acp{dum}. Moreover, we verify that \acp{dum} indeed scale to realistic vision tasks in terms of predictive performance. However we find that (i) the epistemic uncertainty of many \acp{dum} is not well calibrated under distributional shifts and (ii) the regularization strength in methods based on distance awareness does not correlate strongly with \ac{ood} detection and calibration.

\acp{dum} recently showed good \ac{ood} detection performance and are interesting for practical applications in need of efficient uncertainty quantification. We established that \acp{dum} mainly differ in their regularization technique for countering feature collapse and their approach to quantifying uncertainty (\autoref{sec:taxonomy}).
Regarding the calibration under continuous distributional shifts of \acp{dum}, we observe that such uncertainty estimates are considerably worse calibrated than scalable Bayesian methods.
This is the case for image classification (\autoref{ssec:classification}) as well as semantic segmentation (\autoref{ssec:segmentation}) and for synthetic as well as more realistic distributional shifts.
SNGP~\cite{liu2020simple} denotes the only \ac{dum} that consistently yields better calibrated uncertainties under continuous distributional shifts than the softmax entropy. Simultaneously, SNGP is the only \ac{dum} which derives their uncertainty from its predictive distribution.

In particular, we find that methods relying on the distribution of hidden representations to quantify uncertainty~\cite{winkens2020contrastive, postels2020hidden, mukhoti2021deterministic} are poorly calibrated. It is understandable that these methods are worse calibrated than SNGP since they do not take into account the predictive distribution. They assume that locations in the feature space entail information about the correctness of predictions. While this is arguably true, features also contain additional information that renders them sub-optimal for judging the correctness of predictions due to ambiguities. 
%
%We refer to the supplement for a further theoretical consideration of the limitations of \acp{dum}. 
%
Overall, this underlines the necessity to refrain from \acp{dum} that purely rely on distances or log-likelihoods in the feature space when well calibrated uncertainties are required. 

Interestingly, one may argue that \acp{dum} simply do not scale to realistic vision tasks and, for that reason, are not well calibration in \autoref{ssec:segmentation}. However, note that \acp{dum} in fact demonstrate strong predictive performance in such scenarios (see \autoref{tab:cityscapes_c_carla}) and have been shown to perform well on \ac{ood} detection on realistic vision datasets (\eg{} pixel-wise anomaly detection~\cite{blum2019fishyscapes}). Therefore, we can conclude that this is not a problem of scaling \acp{dum} to realistic scenarios but rather an inherent problem which these types of uncertainty estimates. 

Moreover, another desirable property of \acp{dum} would be that the strength of the feature space regularization correlates with the quality of the uncertainty - both in terms of calibration as well as \ac{ood} detection. 
Due to the original purpose of most \acp{dum}, this would be at least expected for \ac{ood} detection. 
%
% In particular, being distance awareness the aim of Lipschitz regularization, one might expect that \ac{ood} detection performance correlates well with regularization strength.
% for methods relying on gradient penalty or spectral normalization to regularize the feature space. 
%
However, we do not observe this for bi-Lipschitz regularization (\autoref{ssec:classification}). We hypothesize that this originates from the fact that these regularization techniques rely on an underlying distance metric - \ie{} $L_2$ distance. Such distance metrics are not meaningful in the case of high-dimensional data, \ie{} images.
%
% As expected, the lack of correlation also extends to calibration performance. 
% regularization parameters, highlighting how current proposals cannot be reliably deployed under distributional shift as a proxy for risk-informative decision-making in safety-critical applications.
We hope that our findings will foster future research on making these promising family of methods better calibrated and more broadly applicable.

%% file: sections/6_acknowledgement.tex
This work was partially supported by Google and by the Max Planck ETH Center for Learning Systems.

%% file: sections/7_supplement.tex
We here provide additional theoretical background, implementation and optimization details, additional results, comparisons and ablation studies.
In particular, 
%we outline an intuition of the poor calibration of \acp{dum} in \autoref{ssec:suppl_theoretical}, and 
we report additional details on the theoretical background necessary to understand \acp{dum}' design choices in \autoref{ssec:suppl_background_theory}. 
We describe techniques used for enforcing Lipschitz constraints \autoref{ssec:suppl_regularization_theory} and informative representations \autoref{ssec:suppl_regularization_informative_representations_theory} in more detail. Moreover, \autoref{ssec:suppl_uncertainty_theory} summarizes common choices of uncertainty estimation techniques for discriminative and generative \acp{dum}.

\autoref{sssec:suppl_classification_results}/\autoref{sssec:suppl_segmentation_results} show additional results for image classification/semantic segmentation.
We report optimization/implementation details in \autoref{ssec:suppl_training_details}/\autoref{ssec:suppl_implementation_details}. 
Furthermore, we describe the quantification of uncertainty for image classification \autoref{ssec:suppl_uncertainty_derivation_classification} and semantic segmentation \autoref{ssec:suppl_uncertainty_derivation_segmentation}.
Finally, we provide details on the data collection process in CARLA and examples from the sequences collected for semantic segmentation \autoref{ssec:suppl_segmentation_dataset}.

\subsection{\acp{dum} - Fundamentals} \label{ssec:suppl_background_theory}
Subsequently, we provide a more detailed introduction of the most common concepts applied by \acp{dum}. Furthermore, we discuss each \ac{dum} individually in light of the introduced taxonomy and highlight its strengths and weaknesses.
Initially (\autoref{ssec:suppl_regularization_theory}, \autoref{ssec:suppl_regularization_informative_representations_theory} and \autoref{ssec:suppl_uncertainty_theory}) maintain a modularized perspective of \acp{dum} by describing individual components including their advantages and disadvantages. Subsequently (\autoref{ssec:suppl_method_oriented_perspective}), we shed light on each \acp{dum} individually using insights from the modularized considerations.
% While providing the necessary insights to understand \acp{dum} design choices and functioning, we keep our description high level and refer the reader to the corresponding literature for a deeper understanding of the mathematical foundations.

In particular, We describe regularization techniques, including Lipschitz regularization (\autoref{ssec:suppl_regularization_theory}) for enforcing distance awareness and informative representations (\autoref{ssec:suppl_regularization_informative_representations_theory}). \autoref{ssec:suppl_uncertainty_theory} summarizes discriminative and generative approaches to uncertainty estimation in \acp{dum}. Finally, \autoref{ssec:suppl_method_oriented_perspective} discusses each individual method used in our empirical comparison.

\subsubsection{Regularization Techniques - Distance Awareness} \label{ssec:suppl_regularization_theory}

The fundamental idea of distance-aware hidden representations is to avoid feature collapse by enforcing distances between latent representations to mirror distances in the input space. 
This can be achieved by constraining the Lipschitz constant, as it enforces a lower and an upper bound to expansion and contraction performed by an underlying neural network. 
More formally, given any pair of inputs $x_1$ and $x_2$ the following lower and upper bounds must hold for the resulting activation of a feature extractor $f_\theta$ with parameters $\theta$: $c_1||x_1-x_2||_I \leq ||f_\theta(x_1)-f_\theta(x_2)||_F \leq c_2||x_1-x_2||_I$.
$c_1$ and $c_2$ denote respectively the lower and upper bound for the Lipschitz constant, and $||\cdot||_I$ and $||\cdot||_F$ are the chosen metrics in the input and feature space respectively.

Recent proposals have primarily adopted two methods to impose this constraint.

\textbf{Gradient Penalty.} 
First introduced to regularize the Lipschitz constant in GAN training~\cite{gulrajani2017improved}, a two-sided gradient penalty is used as an additional loss term to enforce sensitivity of the feature space to changes in the input by DUQ~\cite{van2020uncertainty}.
The gradient penalty is formulated as an additional loss term that regularises the Frobenius norm $||J||_F$ of the Jacobian $J$ of a \ac{nn} to enforce a bi-Lipschitz
constraint. Therefore, the training loss of a \ac{nn} is typically enhanced with the absolute difference between $||J||_F$ and some chosen positive constant.

Given a model $g$ and an input $x$, regularising the Frobenius norm $||J||_F$ of its Jacobian $J$ constraints its Lipschitz constant.
Therefore, the following two-sided gradient penalty is used:
$\lambda \left[ ||\nabla_x g(x)||_F-1\right]^2$,
where $\lambda$ is the regularization strength, $||\cdot||_2$ is the $L_2$ norm, the target bi-Lipschitz constant is 1. For more details, refer to~\cite{van2020uncertainty}.

\textbf{Spectral Normalization.} The two-sided gradient penalty described above requires backpropagating through the Jacobian of a \ac{nn} and is, thus, computationally demanding. A more efficient technique is \ac{sn}~\cite{miyato2018spectral}. For each layer $g: \mathbf{h}_{in} \rightarrow \mathbf{h}_{out}$, \ac{sn} normalizes the weights $W$ of each layer using their spectral norm $sn(W)$ to constrain the bi-Lipschitz constant. Thus, weight matrices are normalized according to: $W_{sn}=\frac{W}{c\cdot sn(W)}$.
This effectively constrains the layer's Lipschitz norm $||g||_{Lip}=sup_{\mathbf{h}}sn(\nabla g(\mathbf{h}))$, where $sn(A)$ is the spectral norm of the matrix A, which is equivalent to its largest singular value.
Consequently, \ac{sn} normalizes the spectral norm of the weights $W$ of each layer to satisfy the soft-Lipschitz constraint $sn(W)=c$ (hard- if the Lipschitz constant $c=1$): $W_{sn}=W/sn(W)$.
Note, that spectral normalization requires residual layers. 
We refer to~\cite{liu2020simple} for further details.

\textbf{Runtime.} Let $N$ denote the number of parameters of the underlying neural network and $B$ denote the batch size used during training of the underlying discriminative task. Gradient penalty leads to additional runtime/memory cost of $O(NB)$. This originates from backpropagation through the gradients of the input which essentially doubles the computation during backpropagation. Spectral normalization leads to additional runtime/memory  cost of $O(N)$ since its complexity equates to applying the affine layers of a model additionally on a single sample.

Overall, we summarize the advantages and disadvantages of each regularization technique enforcing distance aware representations.

\begin{itemize}
    \item \textbf{Distance awareness (general):} 
    \begin{itemize}
        \item \textbf{Advantages:} Can be used in combination with GPs and RBF kernels which both assume distance-aware inputs.
        \item \textbf{Disadvantages:} Assumes an underlying distance metric (e.g. $L_2$). This can be unsuitable/problematic for some data distributions (e.g. images). Does not correlate with \ac{ood} detection performance.
    \end{itemize}
    
    \item \textbf{Gradient Penalty:} 
    \begin{itemize}
        \item \textbf{Advantages:} Architecture-agnostic.
        \item \textbf{Disadvantages:} High computational and memory costs due to backpropagation through the input's gradients.
    \end{itemize}
    
    \item \textbf{Spectral Normalization:} 
    \begin{itemize}
        \item \textbf{Advantages:} Computationally more efficient compared to gradient penalty.
        \item \textbf{Disadvantages:} Not architecture-agnostic. Requires the use of residual layers.
    \end{itemize}
\end{itemize}

\subsubsection{Informative Representations} \label{ssec:suppl_regularization_informative_representations_theory}

Unlike approaches enforcing distance aware representations, informative representations do not rely on an underlying distance metric. These approaches rather aim at maximizing the mutual information between input data distribution and the distribution of hidden representations heuristically~\cite{postels2020hidden} or exactly~\cite{winkens2020contrastive, charpentier2020posterior}. Subsequently, we discuss the different approaches in greater detail.

\textbf{Constrastive learning.}
DCU~\cite{winkens2020contrastive} first pretrains its model using contrastive learning~\cite{chen2020simple}. Subsequently, they finetune on the actual classification task by training simultaneously on a classification and a contrastive learning objective. This approach provably encourages the model to increase the mutual information between the input distribution and the distribution of hidden representations \cite{oord2018representation}. A disadvantage of this approach is the large batch size required for training the contrastive objective. Furthermore, it heavily depends on the underlying data augmentations which need to be tailored to the discriminative task at hand (e.g. classification).

\textbf{Reconstruction regularization.}
The authors of MIR~\cite{postels2020hidden} try to heuristically increase the information content about the input in the hidden representations. Therefore, they require the model to be able to reconstruct its input from its hidden representations using a separate decoder module during training. The entire approach can also be viewed as a constrained autoencoder, where the constraint is the objective of the discriminative task at hand (e.g. classification). While this approach is only a heuristic, it is more agnostic of the underlying discriminative task. 

\textbf{Entropy regularization.}
PostNet~\cite{charpentier2020posterior} learns the distribution of hidden representations end-to-end during training of the discriminative model. They parameterize the distribution using one normalizing flow (radial flow) per class. While they do not explicitly mention the problem of feature collapse, their entropy regularization loss fulfills this purpose. In particular, they maximize the entropy of the predicted Dirichlet distribution $D(\alpha^{(i)})$ parameterized by $\alpha^{(i)} = (\alpha^{(i)}_1, \dots, \alpha^{(i)}_c)$ for a classification problem with $c$ classes. Each $\alpha^{(i)}_j$ is given by $\alpha^{(i)}_j = \beta^{prior} + N_c P(z^{(i)}|c, \phi)$. Here, $\beta^{prior}$ denotes a constant prior term shared across classes, $N_c$ denotes the number of occurrences of class $c$ in the training set, $z^{(i)}$ denotes the hidden representation of some input $x^{(i)}$ and $P(z^{(i)}|c, \phi)$ denotes the radial flow associated with class $c$ with parameters $\phi$. Importantly, $\beta^{prior}$ is set to 1 in their experiments which leads to $\alpha^{(i)}_j \geq 1 \forall j \in [1, \dots, c]$. PostNet then encourages large entropies of the Dirichlet distribution during training. The entropy is given by 
\begin{equation}
    H(D(\alpha^{(i)})) = \log(B(\alpha^{(i)})) + (\alpha^{(i)}_0 - c)\psi(\alpha^{(i)}_0) - \sum_{j=0}^c (\alpha^{(i)}_j - 1)\psi(\alpha^{(i)}_j)
\end{equation}
where $B$ is there beta-function, $\psi$ is the Digamma function and $\alpha^{(i)}_0 = \sum_{j=1}^c\alpha^{(i)}_j$. Importantly, this function has a global maximum for $\alpha^{(i)}_j = 1 \forall j \in [1, \dots c]$. Since $\beta^{prior}=1$, this term further encourages the normalizing flows to produce likelihoods close to zero. Thus, it encourages large values of the negative log-likelihoods and, consequently, entropies under each radial flow. 

\textbf{Runtime.} Let $N$ denote the number of parameters of the underlying neural network and $B$ denote the batch size used during training of the underlying discriminative task. Contrastive learning leads to additional runtime/memory cost of $O(N)$. This originates from the fact that it requires large batch sizes and thus likely increases the batch size compared to the original discriminative task. Reconstruction regularization~\cite{postels2020hidden} and entropy regularization~\cite{charpentier2020posterior} lead to additional runtime/memory cost of $O(B)$, since the size of the decoder model in reconstruction regularization ~\cite{postels2020hidden}, and resp. the normalizing flows in entropy regularization~\cite{charpentier2020posterior}, are in principal independent of the size of the original model.

Overall, we summarize the advantages and disadvantages of each regularization technique enforcing informative representations.

\begin{itemize}

    \item \textbf{Informative representations (general):} 
    \begin{itemize}
        \item \textbf{Advantages:} Does not rely on an underlying distance metric.
        \item \textbf{Disadvantages:} When paired with generative modeling of hidden representations, strong regularization is expected to show similar pathologies as explicit generative models trained directly on the data distribution \cite{nalisnick2018deep}. Moreover, it cannot be paired with RBF kernels and GPs approximations based on RBF kernels, since they work under the assumption that also the feature extractor is a distance-preserving function. 
    \end{itemize}
    
    \item \textbf{Contrastive Learning:} 
    \begin{itemize}
        \item \textbf{Advantages:} Is shown to simultaneously boost predictive performance \cite{chen2020simple, winkens2020contrastive} - particularly on classification. Architecture-agnostic. Provably maximizes mutual-information between input distribution and distribution of hidden representations \cite{oord2018representation}.
        \item \textbf{Disadvantages:} High computational and memory costs due to the necessity of large batch sizes. Contrastive learning needs to be tailored (e.g. data augmentations) to the underlying discriminative task.
    \end{itemize}
    
    \item \textbf{Reconstruction regularization:} 
    \begin{itemize}
        \item \textbf{Advantages:} Architecture-agnostic.
        \item \textbf{Disadvantages:} Only heuristically maximizes mutual information between input distribution and distribution of hidden representations.
    \end{itemize}
    
    \item \textbf{Entropy regularization:} 
    \begin{itemize}
        \item \textbf{Advantages:} Assuming a deterministic neural network, enforcing large entropy in the latent space equates maximizing mutual-information between input distribution and distribution of hidden representations.
        \item \textbf{Disadvantages:} Not architecture-agnostic, since it requires Batch Normalization prior to entropy regularized hidden representations for training stabilization. Further, requires low-dimensional hidden representations.
    \end{itemize}
\end{itemize}

\subsubsection{Uncertainty Estimation} \label{ssec:suppl_uncertainty_theory}
Training a feature extractor under the regularization constraints imposed by distance awareness (\autoref{sssec:distance_awareness}, \autoref{ssec:suppl_regularization_theory}) or representation informativeness (\autoref{sssec:informative_representations}) allows to leverage intermediate representations to quantify uncertainty over network's predictions.
Extending \autoref{ssec:taxonomy_density_models}, we here distinguish between \textit{generative} and \textit{discriminative} approaches to uncertainty quantification in \acp{dum} and provide a detailed categorization of such techniques.

\textbf{Generative approaches.}
Given a model trained under some above-discussed regularization constraint, generative approaches estimate the distribution of hidden representations by fitting density models on the regularized feature space, and use the likelihood as uncertainty metric to detect \ac{ood} samples.
MIR~\cite{postels2020hidden}, DDU~\cite{mukhoti2021deterministic} and DCU~\cite{winkens2020contrastive} learn the density of hidden representations post-training based on the features observed on the training data. In contrast, PostNet~\cite{charpentier2020posterior} learns the density model end-to-end with the underlying discriminative model.

% DCU~\cite{winkens2020contrastive,wu2020simple} introduces a distribution contrastive learning approach to train a deep neural network to act as a distribution encoder. The objective is designed to maximise the likelihood of the positive distribution while pushing away a set of negative distributions. The log-likelihood with respect to the refined positive distribution is later used to estimate the epistemic uncertainty.

Predominantly, class-conditional \acp{gmm} are fitted on the regularized intermediate feature space to estimate its distribution, as done in MIR~\cite{postels2020hidden}, DDU~\cite{mukhoti2021deterministic} and DCU~\cite{winkens2020contrastive} - i.e. one multi-variate gaussian per class, to the hidden representations post training. Subsequently, log-likelihood~\cite{postels2020hidden} or the log-likelihood of the mixture component associated with the predicted class~\cite{mukhoti2021deterministic, winkens2020contrastive} is used as a proxy for epistemic uncertainty.

On the other hand, PostNet~\cite{charpentier2020posterior} learns the class-conditional distribution of hidden representations end-to-end using normalizing flows (in particular radial flows). They learn one normalizing flow per class. In their work the class-conditional distribution is used to parameterize a Dirichlet distribution. Following PostNet's notation, the parameters $\alpha^{(i)}$ of the Dirichlet distribution associated with a particular sample $x^{(i)}$ are given by $\alpha^{(i)}_c=\beta^{prior} + \beta^{i}$ with $\beta^{i} = N_c P(z^{(i)}|c, \phi)$. Here, $c$ denotes the class, $\beta^{prior}$ a constant prior term shared across all classes, $N_c $ the number of samples observed in class $c$ and $P(z^{(i)}|c, \phi)$ ($\phi$ are the parameters of the normalizing flow) is the probability of observing the hidden representation $z^{(i)}$ given the normalizing flow associated with class $c$. Ultimately, epistemic uncertainty is then quantified as the maximum alpha among all classes. Thus, the epistemic uncertainty is directly derived from the likelihood of the normalizing flow associated with the predicted class of the \ac{nn}, which mostly corresponds to the normalizing flow with the maximum likelihood assuming a balanced class distribution. We refer to \cite{charpentier2020posterior} for a more detailed treatment.

Empirically, we find generative approaches to show worse calibration. The underlying assumption of generative approaches to uncertainty estimation is that locations in feature space entail information about the correctness of predictions. While this is arguably true, features also contain additional information that which can render them suboptimal for judging the correctness of predictions due to ambiguities.
% Given that density models are fit on the distribution of features encoded from the training domain, the learned density model will be representative of the distribution regularized feature space conditioned on the input training distribution.
% %
% Shifted domains would instead be encoded to different latent distributions, which a density model fit on the in-domain feature distribution is not expected to represent. 
% %
% As a result, computing the log-likelihood with respect to the training distribution is not meaningful for samples extracted from a shifted distribution.
% %
% We thus deem generative approaches inadequate to provide calibrated uncertainties under distributional shift, as also highlighted by the quantitative and qualitative analysis on their calibration conducted in our study (\autoref{tab:cifar_10_100_c}, \autoref{tab:cifar_10_100_c_main_paper}, \autoref{tab:cityscapes_c_carla}, \autoref{fig:carla_weather}, \autoref{fig:carla_time}).

\textbf{Discriminative}
While generative approaches use the likelihood produced by an explicit generative model fit to the distribution of regularized hidden representations to quantify uncertainty, discriminative methods directly rely on the predictions based on regularized representations.

Centroid-based techniques use distances between points in the latent space to parameterize predictions. Centroids are defined with respect to the distribution of the feature space generated by the training set. 
Mandelbaum \etal~\cite{mandelbaum2017distance} propose to use a Distance-based Confidence Score (DCS) to estimate local density at a point as the Euclidean distance in the embedded space between the point and its k nearest neighbors in the training set.
Similarly, DUQ~\cite{van2020uncertainty} builds on Radial Basis Function (RBF) networks~\cite{lecun1998gradient}, which requires the preservation of input distances in the output space which is achieved using the gradient penalty. The class-specific centroids used in the RBF kernel are maintained as a running mean of the features observed for each class. 
%
%Based on the resulting regularized representations, the authors propose a novel centroid updating scheme with respect to which a distance measure is compute as a proxy for uncertainty. 
%
%Analogously to generative models, relying on confidence proxies based on the training distribution is useful for \ac{ood} detection purposes, but not meaningful to assess the trustworthiness of a model prediction under distributional shift, thus resulting in poor calibration.

Other methods are based on the idea that, since \acp{gp} with RBF kernels are distance preserving functions~\cite{liu2020simple}, they can be combined with regularization techniques that enforce distance awareness of the feature extractor~\cite{liu2020simple,van2021improving} to obtain an end-to-end distance-aware model.
The uncertainty can then be computed at the network's output level as the Dempster-Shafer metric~\cite{liu2020simple} or the softmax entropy~\cite{van2021improving}.
Based on this intuitive idea, SNGP~\cite{liu2020simple} and DUE~\cite{van2021improving} simply rely on different approximations of the \ac{gp}, adopting respectively the Laplace approximation based on the random Fourier feature (RFF) expansion of the \ac{gp} posterior~\cite{rasmussen2003gaussian} and the inducing point approximation~\cite{titsias2009variational,hensman2015scalable}. Another minor difference lies in the spectral normalization algorithm, with DUE providing a SN implementation also for batch normalization layers.
While both methods rely on spectral-normalized feature extractors, they could in principle be applied together with any distance-preserving regularization technique. For example, the \acp{gp} could be placed on top of a feature extractor trained with gradient penalty~\cite{van2020uncertainty} to regularize the bi-Lipschtiz constant.
%
%By not relying on any density proxy based on the in-domain training distribution, \ac{gp}-based techniques have shown the best calibration performance among all \acp{dum}, since the provided uncertainty estimate is computed with respect to the output predictions directly and does suffer from the distributional shift affecting learned density models.

\subsubsection{A Method-oriented Perspective on Individual \acp{dum}}\label{ssec:suppl_method_oriented_perspective}

\input{tables/dums_comparison_supplement}

Here, we discuss each \acp{dum} in our empirical comparison individually. Furthermore, \autoref{table:dums_comparison_supplement} provides a comparison of \acp{dum} used in our empirical evaluation regarding properties which are interesting for practitioners. We consider four characteristics: Architecture/task constraints, well calibrated uncertainties and computational/memory overhead. We consider the computational/memory overhead of a method not minimal when it scales with at least $O(N)$ where $N$ denotes the number of parameters of the underlying neural network. This is the case for contrastive learning in DCU~\cite{winkens2020contrastive} due to large batch sizes, the gradient penalty in DUQ~\cite{van2020uncertainty} due to backpropagation through the gradients of a neural network's input and spectral normalization in SNGP~\cite{liu2020simple} and DDU~\cite{mukhoti2021deterministic}.

\textbf{SNGP}~\cite{liu2020simple} uses spectral normalization for regularizing hidden representations (see \autoref{ssec:suppl_regularization_theory}). While this denotes an efficient approach to enforcing distance-aware representations, it renders them dependent on an underlying distance metric in the input space. Moreover, they require the underlying model to be composed of residual layers. Furthermore, we find that enforcing distance awareness, does not directly correlate with \ac{ood} detection performance. SNGP estimates uncertainty by replacing the softmax layer with a \ac{gp} based on the RBF kernel. In particular, they use a Laplace approximation of the \ac{gp}. They estimate epistemic uncertainty using the Dempster-Shafer metric~\cite{liu2020simple}. We find that SNGP yields reasonable uncertainty calibration.

\textbf{DUQ}~\cite{van2020uncertainty} prevents feature collapse by enforcing distance-aware hidden representations. Distance-awareness is enforced using the gradient penalty. While the latter allows DUQ to be model agnostic, it dramatically increases the computational/memory cost at training time. Moreover, following the general drawbacks of enforcing distance-aware representations it depends on an underlying distance metric in the input space. DUQ is only applicable to classification and replaces the softmax output layer using an RBF kernel which compares observed representations to centroids where each class in the classification problem is associated with one centroid. These centroids are updated using a running mean during the training of the model. This renders DUQ sensitive to instabilities at training time in case the mean updates are noisy. For example, the latter case can arise when the number of classes in the classification problem becomes large.

\textbf{DDU}~\cite{mukhoti2021deterministic} uses spectral normalization for regularizing hidden representations (see \autoref{ssec:suppl_regularization_theory}). While this denotes an efficient approach to enforcing distance-aware representations, it renders them dependent on an underlying distance metric in the input space. Moreover, they require the underlying model to be composed of residual layers. In order to estimate uncertainty, DDU estimates the distribution of hidden representations of the penultimate layer using a class-conditional \ac{gmm}, i.e. they train one multivariate Gaussian per class. Then, epistemic uncertainty is approximated as the negative log-likelihood of the mixture components with the highest probability. This approach to uncertainty estimation allows DDU to be applied across different tasks. However, due to the generative approach to uncertainty estimation it yields poorly calibrated uncertainty.

\textbf{DCU}~\cite{winkens2020contrastive} enforces informative representations at training time to counter feature collapse. To this end DCU uses constrastive learning (see \autoref{ssec:suppl_regularization_informative_representations_theory}) as a regularization objective. While this approach has theoretical guarantees for maximizing the information content in the hidden representations and is architecture-agnostic, in practically requires very large batch size to generate hard negative samples at training time~\cite{chen2020simple}. Moreover, while the contrastive learning objective boosts performance on classification, it is not directly transferable to other tasks than classification since those may require a different set of data augmentations which are essential to the success of contrastive learning~\cite{chen2020simple}. In order to estimate uncertainty, DCU also estimates the distribution of hidden representations of the penultimate layer using a class-conditional \ac{gmm}, i.e. they train one multivariate Gaussian per class. Then, epistemic uncertainty is approximated as the negative log-likelihood of the mixture components with the highest highest probability. This approach to uncertainty estimation allows DCU to be applied across different tasks. However, due to the generative approach to uncertainty estimation it yields poorly calibrated uncertainty.

\textbf{MIR}~\cite{postels2020hidden} regularizes hidden representations using reconstruction regularization. While this approach only heuristically increases the information content in the hidden representations, it is efficient and architecture- and task-agnostic. In order to estimate uncertainty, MIR also estimates the distribution of hidden representations of the penultimate layer using a class-conditional \ac{gmm}, i.e. they train one multivariate Gaussian per class. Then, epistemic uncertainty is approximated as the negative marginal log-likelihood. This approach to uncertainty estimation allows DCU to be applied across different tasks. However, due to the generative approach to uncertainty estimation it yields poorly calibrated uncertainty.

\textbf{PostNet}~\cite{charpentier2020posterior} learns the distribution of hidden representations end-to-end which allows them to regularize its entropy directly (see \autoref{ssec:suppl_regularization_informative_representations_theory}). While this is theoretically guaranteed to lead to high information content in the hidden representations for deterministic models, it leads to some difficulties during training. To ensure stability during training the hidden representations are require to be low dimension. Furthermore, it is required to apply a Batch Normalization layer directly prior to hidden representations which distribution is estimated. While PostNet's approach can be generalized to other predictive distributions, they focus on Dirichlet distributions and thus only classification. PostNet estimates the distribution of hidden representations using one radial flow per class. Uncertainty is estimated using the likelihood of the radial flow with the maximum likelihood. Precisely, they multiply the likelihood with the elements observed for a particular class in the training set which is usually constant in their/our experiments and add one to the result. In accordance with other approaches that use generative modeling of hidden representations for uncertainty estimation, we observe poor calibration of epistemic uncertainty.

%%% Commenting this out due to ICLR final decision
% \subsection{Official Implementation} \label{ssec:code}
% An official implementation of this work can be found, completely anonymized, at the following link:
% %
% \url{https://1drv.ms/u/s!AmvoAvndzeKvarxW2fCTi_4W9-o?e=KjaQNE}

\subsection{Intuition for Poor Calibration} \label{ssec:suppl_theoretical}
Given a neural network NN trained on a dataset $D=(X, Y)$ with $X = \{x_i\}_{i \in |D|}$ and $Y = \{y_i\}_{i \in |D|}$, some DUMs~\cite{winkens2020contrastive, postels2020hidden, mukhoti2021deterministic} require modeling of the intermediate activations $p(z|X, Y)$  of the NN to derive estimates of the predictive uncertainty. For this reason, different types of regularization over the feature space are applied with the aim of making the latent distribution representative of the input one. Since this only captures the data distribution through the lense of a fixed set of model parameters, we argue that a key ingredient is missing to account for the total variability over latent distribution.

% We here provide a complete mathematical modeling of the distribution over the latent space as seen through the eyes of a neural network, in an attempt to find flaws in the current modeling of latent distributions considered by DUMs.

%, we must , which can be typically approximated by a surrogate distribution $q(\theta)$, e.g., Bernoullian as in MC Dropout. We can thus write

Taking into account the distribution $p(\theta|X,Y)$ over networks' parameters $\theta$, which is approximated by a surrogate distribution $q(\theta)$, we obtain for the distribution $p(z|X, Y)$:

\begin{equation}
\begin{aligned}
p(z|X,Y) &= \int_x p(z|x,X,Y) p(x|X,Y) dx \\
         &= \int_x \left( \int_\theta p(z|x,\theta) p(\theta|X,Y)d\theta\right) p(x|X,Y) dx \\
         &= \int_x \left( \int_\theta p(z|x,\theta) q(\theta)d\theta\right) p(x|X,Y) dx \\
         &= \int_\theta \left( \int_x p(z|x,\theta) p(x|X,Y) dx \right) q(\theta) d\theta
\end{aligned}
\end{equation}

% We observe that DUMs do not typically take into account the posterior $p(\theta|X,Y)\approx q(\theta)$ over the model's parameters $\theta$.
% % , thus completely ignoring the outer integral over the network's weights.

DUMs typically assume that network weights are fixed, i.e. they are distributed according to a Dirac delta function:

\begin{equation}
\begin{aligned}
p(\theta|X,Y) \approx q(\theta) = \delta(\theta) =  \begin{cases} +\infty, & \theta=\hat{\theta} \\ 0, & \theta \ne \hat{\theta} \end{cases}
\end{aligned}
\end{equation}

subject to $\int_{-\infty}^\infty \delta(x) \, dx = 1$. Then,

\begin{equation}
\begin{aligned}
p(z|X,Y) &= \int_\theta \left( \int_x p(z|x,\theta) p(x|X,Y) dx \right) q(\theta) d\theta \\
         &= \int_x \left( p(z|x,\hat{\theta} \right) p(x|X,Y) dx
\end{aligned}
\end{equation}

Intuitively, DUMs only approximate the inner integral over the distribution of inputs $x$, since only the distribution over the input space is taken into account and weights $\hat{\theta}$ are fixed. While this justifies the good performance of DUMs on OOD detection, failing to model the weights distribution may not account for the total variability over latent distributions, thus underestimating the output predictive uncertainty and making it a bad hint for networks' expected error. 
% As shown in experimental results, when datasets are large enough to explain away part of the epistemic uncertainty that is related to the target task, model weights converge to a point distribution and DUMs can perform well (e.g. MNIST and CIFAR10). Instead, when the task is more complex and possibly requires more annotations to estimate precise point estimate model weights (e.g. CIFAR100 and semantic segmentation), DUMs fail to account for the contribution to the feature space variability due to lack of knowledge over network weights. This possibly explains the inadequate calibration of current deterministic uncertainty methods in complex scenarios.

Incorporating the lack of knowledge over the network's weights is a promising direction to make DUMs well calibrated.
%%% Commenting this out due to ICLR final decision

%
\subsection{Additional Results} \label{ssec:suppl_additional_results}
\subsubsection{Additional Visualization of Quantitative Results on Continuous Distributional Shifts}
Fig.~\ref{fig:suppl_acc_vs_raulc} provides additional visualization of the test accuracy versus calibration performance for the methods compared in Tab.~\ref{tab:cifar_10_100_c} and~\ref{tab:cityscapes_c_carla}. 
\subsubsection{Image Classification} \label{sssec:suppl_classification_results}
\textbf{OOD Detection.}
\autoref{table:ood_performance_appendix} shows quantitative results on detecting \ac{ood} data for \acp{dum}, \ac{mc} dropout and deep ensembles trained on CIFAR10/100.
We observe that \acp{dum} are able to outperform MC dropout and deep ensembles on \ac{ood} detection.
\input{tables/acc_vs_raulc}
\input{tables/image_classification_ood_cifar10_100}

\textbf{Sensitivity to regularization strength.}
\input{figures/image_classification/supplementary_regularization_strength/mnist_regularization_strength}
\input{figures/image_classification/supplementary_regularization_strength/fashion_mnist_regularization_strength}
We provide additional ablation studies on the sensitivity to regularization strength for different methods on MNIST (\autoref{fig:suppl_mnist_sensitivity_to_regularization_strength}) and FashionMNIST (\autoref{fig:suppl_fashion_mnist_sensitivity_to_regularization_strength})
% , CIFAR10 (\autoref{fig:suppl_cifar_sensitivity_to_regularization_strength}), and SVHN (\autoref{fig:suppl_svhn_sensitivity_to_regularization_strength}).
%
These results confirm the findings of the main manuscript, \ie{} that only MIR and Dropout are sensible to regularization strength, while \acp{dum} based on Lipschitz regularization are not influenced by the regularization strength. 
% Note that for MIR we observe that regularization strength and calibration/ood detection are anti-correlated. We hypothesize that this originates from generative models inability to distinguish these datasets~\cite{nalisnick2018deep}. Thus, storing more information in the latent representations about the input leads to worse \ac{ood} detection performance.

\textbf{Corruption Severity Analysis.}
We show the classification accuracy, \ac{auroc} and \ac{raulc} for each method on the CIFAR10-C (\autoref{fig:corruptions_cifar_10_c}, \autoref{fig:corruptions_cifar_10_c_2}, \autoref{fig:corruptions_cifar_10_c_3}) and CIFAR100-C (\autoref{fig:corruptions_cifar_100_c}, \autoref{fig:corruptions_cifar_100_c_2}, \autoref{fig:corruptions_cifar_100_c_3}) datasets under different types of corruptions. Each type of corruption is applied at 5 increasing levels of severity.  
For ease of visualization, we split the 15 different types of corruptions applied on the CIFAR10-C and CIFAR100-C datasets into 3 different figures each. Each figure shows 5 different types of corruptions.

While all methods demonstrate a similar predictive performance pattern, \acp{dum} - in particular methods based on generative modeling of hidden representations - yield worse calibration across corruption severities.

\textbf{Training/Inference Runtime Comparison.}
We report the per sample training and inference runtime in \autoref{table:runtime_appendix}. The runtimes were  measured on a single V100 using CIFAR10 and a ResNet50 backbone.
\input{tables/runtime}

\subsubsection{Semantic Segmentation}  \label{sssec:suppl_segmentation_results}
% \textbf{Weather conditions.} Additionally, we evaluate the calibration performance of DUMs under changing weather conditions. Fig.~\ref{fig:carla_weather} of Sec.~\ref{ssec:suppl_segmentation_dataset} demonstrates examples of this distributional shift. The quantitative evaluation results are summarized in Table~\ref{table:calibration_segmentation_weather}. Similar to the trend in the Table~\ref{table:calibration_segmentation}, all methods yield decent performance on pixel accuracy and mIoU, and the DUMs obtain significant improvements over the Softmax baseline. However, we again observe that \ac{mc} dropout outperforms \acp{dum} on continuous distributional shifts. 
% \begin{table}[!h]
%     \centering
%     \scalebox{0.9}{
%     \input{tables/segmentation2}
%     }
%     \caption{Calibration results on semantic segmentation for weather conditions.}
%     \label{table:calibration_segmentation_weather}
% \end{table}
 
\textbf{Examples of segmentation and uncertainty masks.}
\input{figures/supplementary_segmentation_uncertainty/minimal_shift_uncertainty}
\input{figures/supplementary_segmentation_uncertainty/maximal_shift_uncertainty}
We show qualitative examples of predicted masks, error masks and uncertainty masks for Softmax, \ac{mc} dropout, SNGP and MIR on semantic segmentation. 
\autoref{fig:suppl_segmentation_minimal} illustrates examples under \textit{minimal} distributional shift (\ie{} Azimuth angle of the sun $=85\degree$ and \autoref{fig:suppl_segmentation_maximal} under \textit{maximal} distributional shift (\ie{} Azimuth angle of the sun $=-5\degree$.
We show the input image (Input), the segmentation ground truth (GT), the predicted segmentation mask (Prediction), the error mask (Error) and the uncertainty mask (Uncertainty).
The error mask is computed as a boolean mask with True values when a pixel is predicted wrongly (yellow) and False (blue) when the prediction is instead correct. 
The uncertainty mask is preprocessed to facilitate visualization. In particular, we first compute mean $\mu$ and standard deviation $\sigma$ of per-pixel uncertainties over each uncertainty mask. Then, the uncertainty mask is clipped between $[\mu-2\sigma, \mu+2\sigma]$. Finally, the uncertainty mask is normalized between $0$ and $1$ before being visualized.

While the softmax entropy provides decent uncertainty estimates under minimal distributional shift, it tends to be overconfident under severe distributional shift.
In particular, Softmax models are only uncertain close to object borders, but they are confident about large portions of the image that are instead predicted wrongly.
This can be observed in \autoref{fig:suppl_segmentation_maximal}, where all models tend to predict the entire sky wrongly (assigned to `building' class), but the Softmax model is the most confident about its predictions of the sky being correct.
\acp{dum} and \ac{mc} dropout do a better job at recognizing wrong predictions under sever domain shift by outputting higher uncertainty values.

\textbf{Qualitative Behaviour on Continuous Distributional Shifts}
We show the \ac{miou}, \ac{auroc} and \ac{raulc} for each method on the Carla (\autoref{fig:corruptions_carla_c}) and Cityscapes (\autoref{fig:corruptions_cityscapes_c}, \autoref{fig:corruptions_cityscapes_c_2}, \autoref{fig:corruptions_cityscapes_c_3}) datasets under different levels of corruptions.
For ease of visualization, we split the 15 different types of corruptions applied on the Cityscapes dataset into 3 different figures, showing 5 types of corruptions each.

While all methods demonstrate a similar \ac{miou} pattern, \acp{dum} - in particular methods based on generative modeling of hidden representations - yield worse calibration across corruption severities.

\subsection{Training Details}
\label{ssec:suppl_training_details}
We provide training and optimization details for all evaluated methods.
All methods using spectral normalization use $1$ power iteration.
Hyperparameters were chosen to minimize the validation loss.

\subsubsection{Image Classification - MNIST/FashionMNIST}
All methods trained on MNIST/FashionMNIST used a MLP as backbone with $3$ hidden layers of $100$ dimensions each and ReLU activation functions.
We used a batch size of $128$ samples and trained for $200$ epochs.
No data augmentation is performed.

\textbf{Softmax and Deep ensembles.}
We used for the single softmax model the Adam optimizer with learning rate $0.003$, and $L_2$ weight regularization $0.0001$. When using ensembles, $10$ models are trained from different random initializations.

\textbf{\ac{mc} dropout.}
We used for all baselines the Adam optimizer with learning rate $0.003$, dropout rate $0.4$ and $L_2$ weight regularization $0.0001$.

% \textbf{DUE}
% We trained DUE with the SGD optimizer with learning rate $0.01$, $L_2$ weight regularization $0.0005$, and a multi-step learning rate decay policy with decay rate $0.2$ and decay steps at the epochs $60$, $120$, $160$.
%
We found the optimal \ac{sn} coefficient to be $7$, with the GP approximation using $10$ (number of classes) inducing points initialized using k-means over $10000$ samples.

\textbf{DUQ}
We trained DUQ with the SGD optimizer with learning rate $0.01$, $L_2$ weight regularization $0.0001$, and a multi-step learning rate decay policy with decay rate $0.3$ and decay steps at the epochs $10$, $20$.
Lengthscale for the RBF kernel is $0.1$ and optimal gradient penalty loss weight is $0$ where we searched along the grid $[0.0, 0.0000001, 0.0000003, 0.000001, 0.000003, 0.00001$, $0.00003, 0.0001, 0.0003, 0.001, 0.005, 0.01, 0.025, 0.05$, $0.075, 0.1, 0.2, 0.5]$.

\textbf{DDU.}
We trained DDU with the Adam optimizer with learning rate $0.001$, $L_2$ weight regularization $0.0001$, and a multi-step learning rate decay policy with decay rate $0.2$ and decay steps at the epochs $100$, $200$, $300$.
We found the optimal \ac{sn} coefficient to be $6$ searching along the grid $[1, 2, 3, 4, 5, 6, 7, 8, 10, 12, 15]$. The GMM is fitted by estimating the empirical mean and covariance matrix of the representations on the training data associated with each class.

\textbf{SNGP.}
We trained SNGP with the SGD optimizer with learning rate $0.05$, $L_2$ weight regularization $0.0003$, and a multi-step learning rate decay policy with decay rate $0.2$ and decay steps at the epochs $60$, $120$, $160$.
We found the optimal \ac{sn} coefficient to be $6$ searching along the grid $[1, 2, 3, 4, 5, 6, 7, 8, 10, 12, 15]$., with the GP approximation using $10$ hidden dimensions, lengthscale $2$ and mean field factor $30$.

\textbf{MIR.} 
We trained MIR with the Adam optimizer with learning rate $0.001$, and $L_2$ weight regularization $0.0001$.
We found the optimal reconstruction loss weight to be $1$ after searching along the grid $[0.0, 0.1, 0.5, 1.0, 2.0, 5.0, 10.0, 20.0, 50.0, 100.0]$.

\subsubsection{Image Classification - CIFAR10/SVHN}
When training on CIFAR-10/SVHN, we use a ResNet-18~\cite{he2016deep} as backbone. 
The dimensionality of the last feature space encoded with the ResNet backbone is $100$ for all methods.
We used a batch size of $128$ samples and trained for $400$ epochs.
The training set is augmented with common data augmentation techniques. We apply random horizontal flips, random brightness augmentation with maximum delta $0.2$ and random contrast adjustment with multiplier lower bound $0.8$ and upper bound $1.2$.

\textbf{Softmax and Deep ensembles.}
We used for the single softmax model the Adam optimizer with learning rate $0.003$, $L_2$ weight regularization $0.0001$, and a multi-step learning rate decay policy with decay rate $0.2$ and decay steps at the epochs $250$, $300$, $400$. When using ensembles, $10$ models are trained from different random initializations.

\textbf{\ac{mc} dropout.}
We used for all baselines the Adam optimizer with learning rate $0.003$, dropout rate $0.3$, $L_2$ weight regularization $0.0001$, and a multi-step learning rate decay policy with decay rate $0.2$ and decay steps at the epochs $250$, $300$, $400$.

\textbf{DUE}
We trained DUE with the SGD optimizer with learning rate $0.01$, $L_2$ weight regularization $0.0005$, dropout rate $0.1$, and a multi-step learning rate decay policy with decay rate $0.2$ and decay steps at the epochs $100$, $200$, $300$.
We found the optimal \ac{sn} coefficient to be $7$ for SVHN and $9$ for CIFAR-10, with the GP approximation using $10$ (number of classes) inducing points initialized using k-means over $10000$ samples.

\textbf{DUQ}
We trained DUE with the SGD optimizer with learning rate $0.01$, $L_2$ weight regularization $0.0001$, dropout rate $0.1$, and a multi-step learning rate decay policy with decay rate $0.3$ and decay steps at the epochs $200$, $250$. $300$.
Lengthscale for the RBF kernel is $0.1$ and optimal gradient penalty loss weight is $0$

\textbf{DDU.}
We trained DDU with the Adam optimizer with learning rate $0.001$, $L_2$ weight regularization $0.0001$, dropout rate $0.3$, and a multi-step learning rate decay policy with decay rate $0.2$ and decay steps at the epochs $80$, $120$, $180$.
We found the optimal \ac{sn} coefficient to be $7$. The GMM fit on top of the pretrained feature extractor is trained for $100$ epochs and is fit with $64$ batches.

\textbf{SNGP.}
We trained SNGP with the SGD optimizer with learning rate $0.05$, $L_2$ weight regularization $0.0004$, dropout rate $0.1$, and a multi-step learning rate decay policy with decay rate $0.2$ and decay steps at the epochs $100$, $200$, $300$.
We found the optimal \ac{sn} coefficient to be $7$, with the GP approximation using $10$ hidden dimensions, lengthscale $2$ and mean field factor $30$.

\textbf{MIR.} 
We trained MIR with the Adam optimizer with learning rate $0.003$, $L_2$ weight regularization $0.0001$, dropout rate $0.1$, and a multi-step learning rate decay policy with decay rate $0.2$ and decay steps at the epochs $150$, $200$, $250$, $300$.
We found the optimal reconstruction loss weight to be $1$.

\subsubsection{Semantic Segmentation.}
When training on semantic segmentation, we use a \ac{drn}~\cite{Yu2016, Yu2017} (DRN-A-50) as backbone.
We used a batch size of $4$ samples and trained for $200$ epochs.
Images are rescaled to size $400 \times 640$
The training set is augmented with common data augmentation techniques.
All training samples are augmented with random cropping with factor $0.8$. We apply random horizontal flips, random brightness augmentation with maximum delta $0.2$ and random contrast adjustment with multiplier lower bound $0.8$ and upper bound $1.2$.

\textbf{Softmax.}
We used for the single softmax model the Adam optimizer with learning rate $0.0004$, $L_2$ weight regularization $0.0001$, and a multi-step learning rate decay policy with decay rate $0.3$ and decay steps at the epochs $30$, $60$, $90$, $120$. 

\textbf{\ac{mc} dropout.}
We used for all baselines the Adam optimizer with learning rate $0.0004$, dropout rate $0.4$, $L_2$ weight regularization $0.0001$, and a multi-step learning rate decay policy with decay rate $0.3$ and decay steps at the epochs $30$, $60$, $90$, $120$.

\textbf{SNGP.}
We trained SNGP with the SGD optimizer with learning rate $0.0002$, $L_2$ weight regularization $0.0003$, dropout rate $0.1$, and a multi-step learning rate decay policy with decay rate $0.2$ and decay steps at the epochs $20$, $40$, $60$, $80$, $100$.
We found the optimal \ac{sn} coefficient to be $6$, with the GP approximation using $128$ hidden dimensions, lengthscale $2$ and mean field factor $25$.

\textbf{MIR.}
We trained MIR with the Adam optimizer with learning rate $0.0002$, $L_2$ weight regularization $0.0001$, dropout rate $0.1$, and a multi-step learning rate decay policy with decay rate $0.3$ and decay steps at the epochs $30$, $60$, $90$, $120$.
We found the optimal reconstruction loss weight to be $1$.

\subsection{Implementation Details.} \label{ssec:suppl_implementation_details}
All methods were re-implemented in Tensorflow 2.0.
We payed attention to all the details reported in each paper and we run all experiments for each method multiple times to account for stochasticity, \ie{} $5$ times for classification and $3$ times for segmentation.
When an implementation was publicly available, we relied on it. 
This is the case for DUQ (\url{https://github.com/y0ast/deterministic-uncertainty-quantification}), SNGP (\url{https://github.com/google/uncertainty-baselines/blob/master/baselines/imagenet/sngp.py}) and DUE (\url{https://github.com/y0ast/DUE}).

\textbf{SNGP.} We follow the publicly available implementation of SNGP, which, compared to the implementation described in the original paper, proposes to further reduce the computational overhead of the GP approximation by replacing the Monte-Carlo averaging with the mean-field approximation~\cite{daunizeau2017semi}.
This is especially relevant in large-scale tasks like semantic segmentation, were it is important to reduce the computational overload.

\subsubsection{Image Classification}
\textbf{DUE.} Note that only DUE uses a \ac{sn} approximation also for the batch normalization layer. 
All other methods only restrict the Lipschitz constant of convolutional and fully connected layers.

\textbf{MIR} only differs from regular softmax models in its decoder module used for the reconstruction regularization loss~\cite{postels2020hidden}. When training MLP architectures the decoder is comprised of two fully-connected layer. The first has a ReLU activation function and 200 output neruons. The second has a linear activation function and its output dimensionality equals that of the models' input data. When training convolutional neural networks the decoder is comprised of four blocks of transpose convolutions, batch normalization layers and ReLU activation functions that gradually upscale the hidden representations to the dimensionality of the input data. These four blocks are followed by a 1x1 convolution with linear activation function.

\subsubsection{Semantic Segmentation}

\textbf{MIR.} Similar to image classification, MIR only differs from regular segmentation models in its decoder module used for the reconstruction regularization loss~\cite{postels2020hidden}. The decoder module is comprised of a single point-wise feed forward layer that maps the hidden representations $\textbf{z}\in \mathbb{R}^{W_z\times H_z \times C_z}$ to $\textbf{z}\in \mathbb{R}^{W_z\times H_z \times 3}$. Subsequently, the result is bilinearly upsampled to the image resolution on which we compute the reconstruction loss.

\subsubsection{Uncertainty Derivation.} \label{ssec:suppl_uncertainty_derivation_classification} 
We provide details on the estimation of uncertainty for the baseline methods.
For details on the uncertainty derivation in DUMs, please refer to \autoref{sec:taxonomy} of the main paper or to the original paper of each amalysed method.

\textbf{Softmax.} In case of the softmax baseline we estimate uncertainty using the entropy of the predictive distribution parameterized by the neural network. Given an input $\textbf{x}$ the entropy $H$ is given by $H(\textbf{y}|\textbf{x}) = \sum -p(\textbf{y}|\textbf{x}) \log (p(\textbf{y}|\textbf{x}))$ where $p(\textbf{y}|\textbf{x})$ are the softmax probabilities.

\textbf{\ac{mc} dropout and deep ensembles.} We following~\cite{gal2017deep} and compute epistemic uncertainty as the conditional mutual information between the weights $\textbf{w}$ and the predictions $\hat{y}$ given the input $\textbf{x}$. Given an input $\textbf{x}$ and a set of weights $\textbf{w}$ we observe the predictive distribution $p(\hat{y}|x, w)$. Then epistemic uncertainty $u_{ep}$ is calculated by approximating the mutual information conditioned on the input $\textbf{x}$:
\begin{align*}
    u_{ep} &= I\left( \hat{y}, w | x \right) \\ 
    &= H(\hat{y}| x) -  H(\hat{y}| w, x) \\
    &= E_{y\sim p(\hat{y}|x)} \left[ -\log(p(\hat{y}|x)) \right] - u_{al}
\end{align*}
where $u_{al}$ denotes the aleatoric uncertainty. Here, $p(\hat{y}|x) = \int \mathrm{d}w p(w) p(\hat{y}|x, w)$ is evaluated using a finite set of samples/ensemble members.

\subsubsection{Uncertainty Derivation for Semantic Segmentation.} \label{ssec:suppl_uncertainty_derivation_segmentation}
We derive uncertainty estimates for each method for semantic segmentation.
We average pixel-level uncertainties under the assumption that all pixels are represented by i.i.d. variables.

\textbf{Uncertainty.}
In our experiments on continuous distributional shifts we want to estimate pixel-level uncertainty for the output map. 
% Therefore, we propose to approximate the uncertainty of the predicted segmentation masks as the average of all pixel-level uncertainties, i.e.,
% %
% \begin{equation}
% \begin{aligned}
%     H(\bfy \mid \bfx) & = \mathbb{E}_{\bfx} \left[ -\int_{\bfy} p(\bfy \mid \bfx) \log p(\bfy \mid \bfx) \mathrm{d} \bfy \right] \\
%     & = \mathbb{E}_{\bfx} \left[ -\int_{\bfy} p(\bfy \mid \bfx) \log p(y_1 \mid \bfx) \mathrm{d} \bfy \right] + \cdots + \mathbb{E}_{\bfx} \left[ -\int_{\bfy} p(\bfy \mid \bfx) \log p(y_n \mid \bfx) \mathrm{d} \bfy \right] \\
%     & \approx \frac{1}{n} \sum_{i=1}^n H(y_i \mid \bfx)
% \end{aligned}
% \end{equation}

\textbf{MIR} estimates epistemic uncertainty using the likelihood of hidden representations $\textbf{z}\in \mathbb{R}^{W_z\times H_z \times C_z}$. Since $\textbf{z}$ is high-dimensional in our experiments, we assume that it factorizes along $W_z$ and $H_z$ and is translation invariant. Formally, $p(\textbf{z}) = \prod_i^{W_z}\prod_{j}^{H_z}p_{\theta}(\textbf{z}_{ij})$ where $\textbf{z}_{ij} \in \mathbb{R}^{W_z\times H_z}$ and $\theta$ is shared across $W_z$ and $H_z$. 

We parameterize $p_{\theta}$ with a GMM with $n=10$ components where each component has a full covariance matrix. We fit the GMM on $100000$ hidden representations ($\textbf{z}_{ij} \in \mathbb{R}^{C_z}$) randomly picked from the training dataset post-training. Since $C_z = 1024$ is still high-dimensional, we first apply PCA to reduce its dimensionality to $32$. 

In the dilated resnet architecture used for semantic segmentation the latent representation $\textbf{z}$ is passed through a point-wise feedforward layer $f: \mathbb{R}^{W_z\times H_z \times C_z} \mapsto \mathbb{R}^{W_z\times H_z \times 3}$ and, subsequently, bilinearly upsampled to image resolution ($\mathbb{R}^{W\times H \times K}$) where K is the number of classes. We could estimate the global, \ie image-level, uncertainty of an input, by providing the negative log-likelihood of the factorizing distribution. However, in order to also obtain pixel-wise uncertainties using MIR, we first compute the negative log-likelihood (\ie epistemic uncertainty) associated with each latent representation $\textbf{z}_{ij}$. Then, we bilinearly upsample the negative log-likelihoods and use the result as proxy for pixel-wise epistemic uncertainty. If we wanted to obtain a global, \ie image-level, uncertainty we could average pixel-level uncertainties.

\subsection{Dataset} \label{ssec:suppl_segmentation_dataset}
To benchmark our model on data with realistically and continuously changing environment, we collect a synthetic dataset for semantic segmentation. 
We use the CARLA Simulator~\cite{Dosovitskiy17} for rendering the images and segmentation masks. The classes definition is aligned with the CityScape dataset~\cite{Cordts2016Cityscapes}. In order to obtain a fair comparison, all the OOD data are sampled with the same trajectory and the environmental objects, except for the time-of-the-day or weather parameters. 

\paragraph{In-domain data} The data is collected from 4 towns in CARLA. We produce 32 sequences from each town. The distribution of the vehicles and pedestrians are randomly generated for each sequence. Every sequence has has 500 frames with a sampling rate of $10$ FPS. From them we randomly sample the training and validation set.

\paragraph{Out-of-domain data} Here, we consider the time-of-the-day and the rain strength as the  parameters for the continuous changing environment. In practice, these two parameters have major influence for autonomous driving tasks. 

\begin{figure}[htbp]
    \def\figwidth{3.4cm}
    \centering
    
    \subfloat[Midday ($90^\circ$)]{\includegraphics[width=\figwidth]{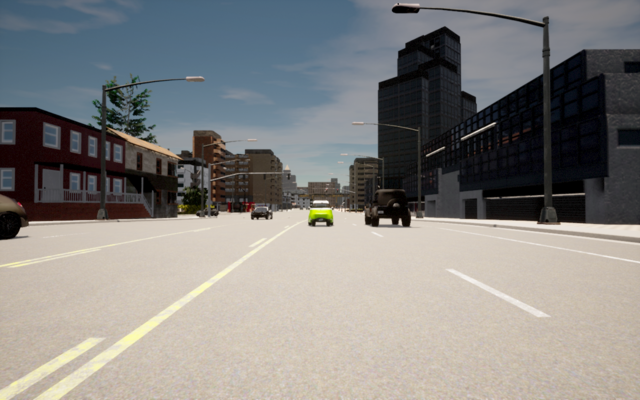}} \ %
    \subfloat[Afternoon ($45^\circ$)]{\includegraphics[width=\figwidth]{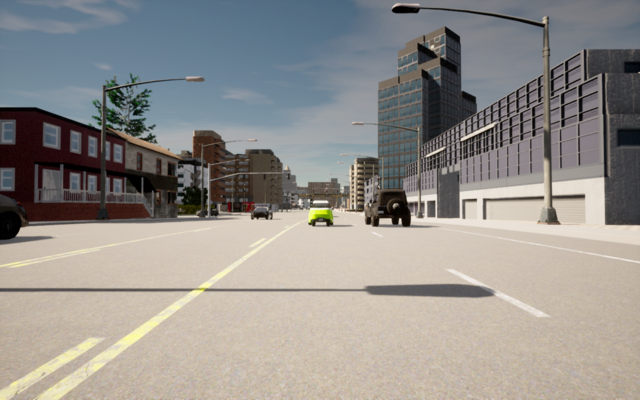}} \ %
    \subfloat[Afternoon ($15^\circ$)]{\includegraphics[width=\figwidth]{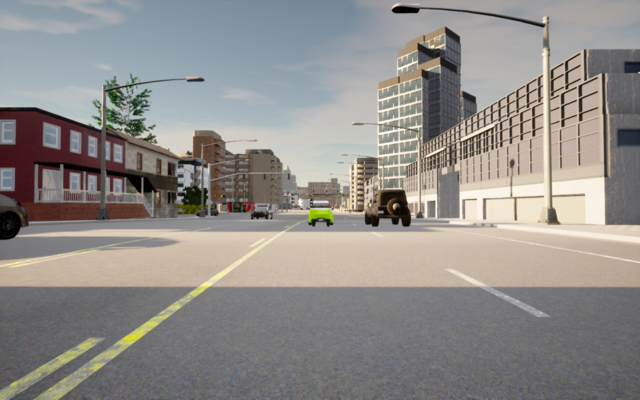}} \ %
    \subfloat[Evening ($10^\circ$)]{\includegraphics[width=\figwidth]{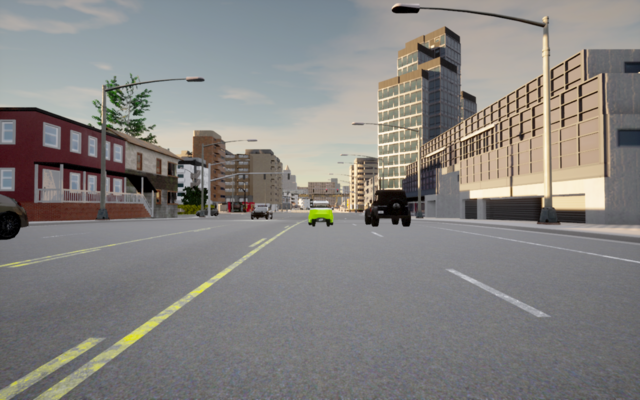}} \\%
    \subfloat[Dusk ($2^\circ$)]{\includegraphics[width=\figwidth]{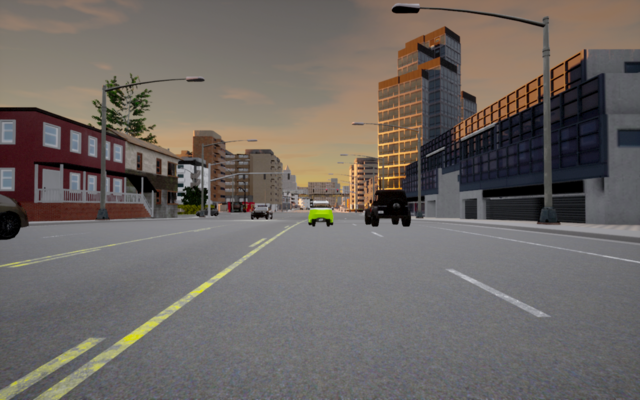}} \ %
    \subfloat[Sunset ($0^\circ$)]{\includegraphics[width=\figwidth]{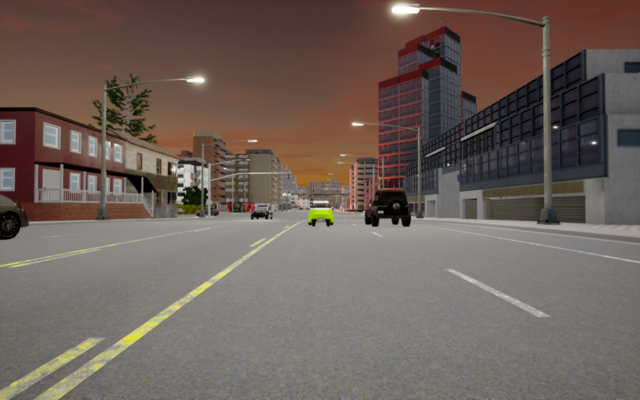}} \ %
    \subfloat[Sunset ($-2^\circ$)]{\includegraphics[width=\figwidth]{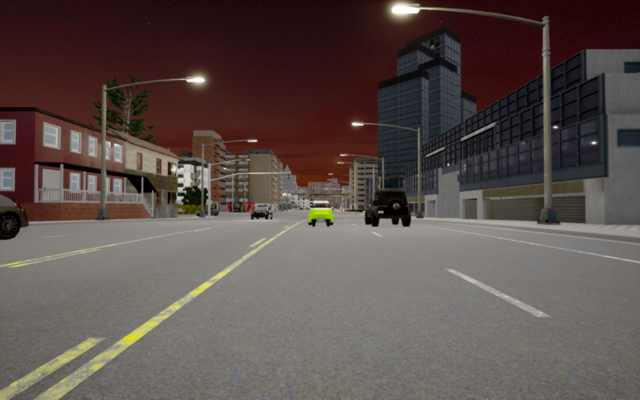}} \ %
    \subfloat[Night ($\leq -5 ^\circ$)]{\includegraphics[width=\figwidth]{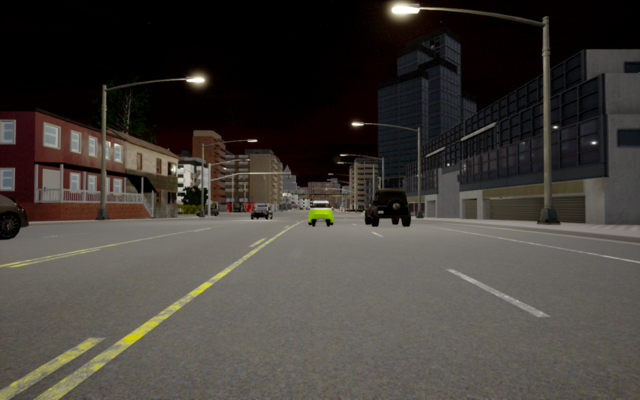}} \\ %
    \subfloat[Clear]{\includegraphics[width=\figwidth]{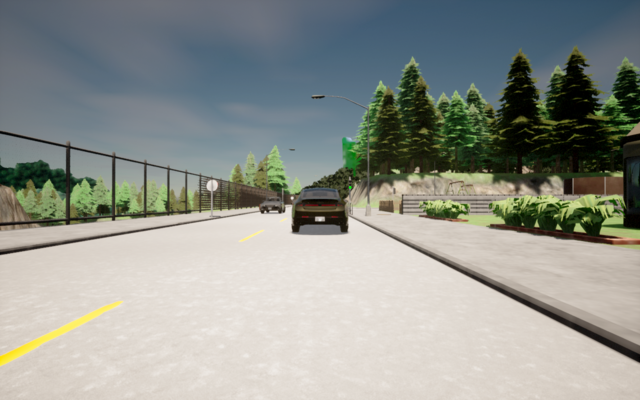}} \ %
    \subfloat[Cloudy]{\includegraphics[width=\figwidth]{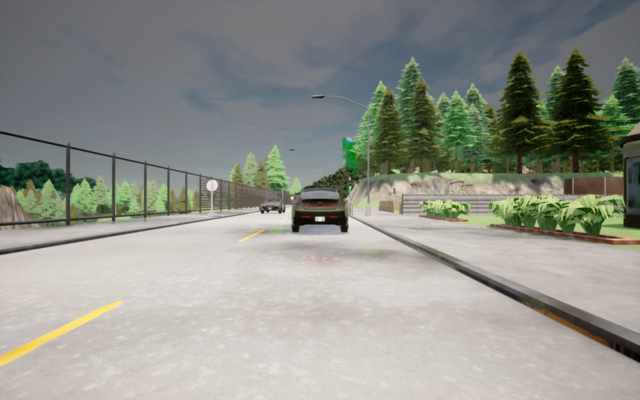}} \ %
    \subfloat[Small rain]{\includegraphics[width=\figwidth]{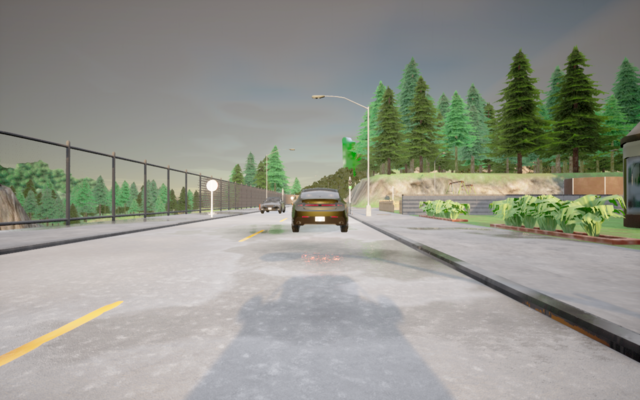}} \ %
    \subfloat[Heavy rain]{\includegraphics[width=\figwidth]{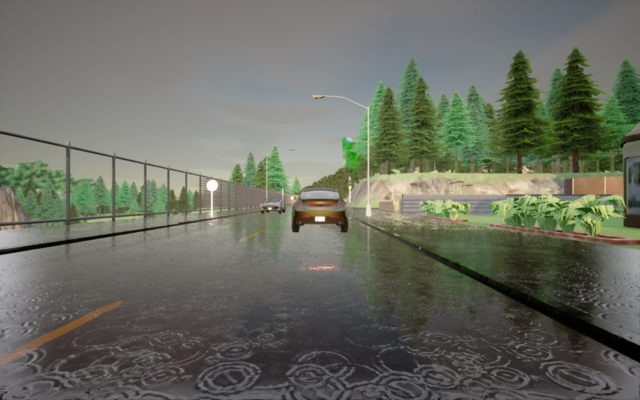}}
    
    \caption{Changing of the time-of-the-day and the weather}
    \label{fig:carla_time}
\end{figure}

 The change of the time-of-the-day is illustrated in Fig.~\ref{fig:carla_time} (first and second row). The time-of-the-day is parametrized by the Sun's altitude angle, where $90^\circ$ means the mid-day and the $0^\circ$ means the dusk or dawn. Here, we produce samples with the altitude angle changes from $90^\circ$ to $15^\circ$ by step of $5^\circ$, and $15^\circ$ to $-5^\circ$ by step of $1^\circ$ where the environment changes shapely. From these examples, we can confirm that the change of time-of-the-day leads to the major change in the lightness, color and visibility of the sky, roads and the buildings nearby.  
The effect of rain strength is demonstrated in Fig.~\ref{fig:carla_time} (bottom). Here the cloudiness and, ground wetness and ground reflection are the main changing parameters. 

% \begin{figure}[htbp]
%     \def\figwidth{3.4cm}
%     \centering
    
%     \caption{Changing of the weather}
%     \label{fig:carla_weather}
% \end{figure}

\input{figures/supplementary_corrupted/cifar10_c}
\input{figures/supplementary_corrupted/cifar100_c}
\input{figures/supplementary_corrupted/cityscapes_c}
\input{figures/supplementary_corrupted/carla_c}

%% file: tables/dums_comparison_supplement.tex
\begin{table}[htbp]
  \centering
  \footnotesize
  \setlength{\aboverulesep}{0pt}
  \setlength{\belowrulesep}{0pt}
  \renewcommand{\arraystretch}{1.33}
  \setlength{\tabcolsep}{4pt}
  \begin{tabular}{|c|c|c|c|c|}
    \hline 
    Method & \shortstack{Architecture\\ constraints} & Task constraints & Well-calibrated & Minimal computational/memory cost \\
    \hline
    SNGP\cite{liu2020simple} & residual layers & classification & \ding{51} & \ding{55} \\
    DUQ\cite{van2020uncertainty} & - & classification & \ding{55} & \ding{55} \\
    DDU\cite{mukhoti2021deterministic} & residual layers & - & \ding{55} & \ding{55} \\
    DCU\cite{winkens2020contrastive} & - & classification & \ding{55} & \ding{55} \\
    MIR\cite{postels2020hidden} & - & - & \ding{55} & \ding{51} \\
    PostNet\cite{charpentier2020posterior} & \shortstack{batch\\normalization} & classification & \ding{55} & \ding{51} \\
    \hline
  \end{tabular}
  \vspace{0.1em}
   \caption{Qualitative comparison of different \acp{dum} used in our empirical comparison. We are interested in four characteristics that are interested from a practical perspective: Architecture/task constraints, well calibrated uncertainties and computational/memory overhead. We consider the computational/memory overhead of a method not minimal when it scales with at least $O(N)$ where $N$ denotes the number of parameters of the underlying neural network. This is the case for contrastive learning in DCU~\cite{winkens2020contrastive} due to large batch sizes, the gradient penalty in DUQ~\cite{van2020uncertainty} due to backpropagation through the gradients of a neural network's input and spectral normalization in SNGP~\cite{liu2020simple} and DDU~\cite{mukhoti2021deterministic}.} \label{table:dums_comparison_supplement}
\end{table}

%% file: tables/acc_vs_raulc.tex
\begin{figure}[t]
\setlength{\tabcolsep}{-2pt}
\includegraphics[width=0.22\linewidth]{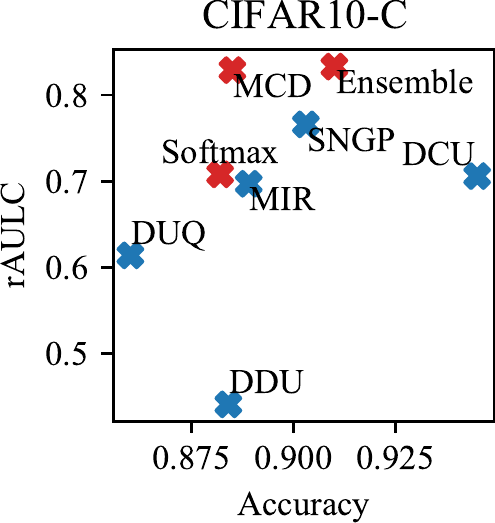} \hfill
\includegraphics[width=0.22\linewidth]{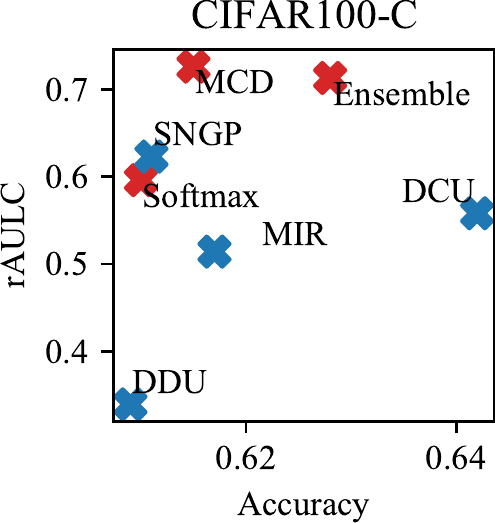} \hfill
\includegraphics[width=0.22\linewidth]{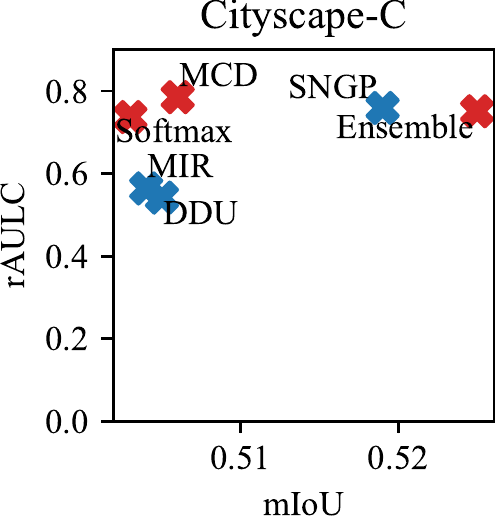} \hfill
\includegraphics[width=0.22\linewidth]{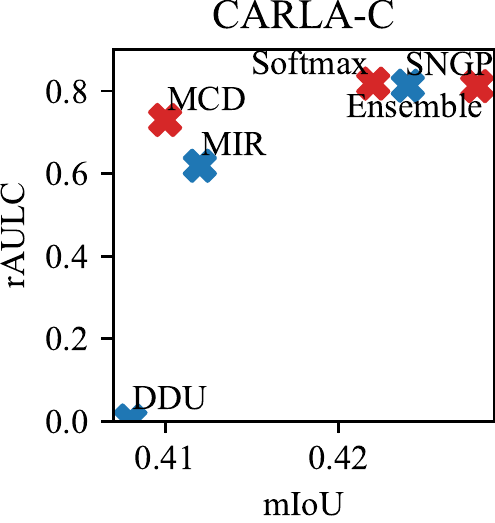}
\vspace{-0.5em}
\caption{Scatter plots for Accuracy/mIoU versus \ac{raulc} on 4 testsets, based on Tab.~\ref{tab:cifar_10_100_c} and~\ref{tab:cityscapes_c_carla}. The baselines (red) usually occupy the top part of the figure, confirming their effectiveness of uncertainty calibration. Among the DUMs (blue), SNGP and MIR are closer to the region of baselines than others. For realistic shifts (CARLA-C), a significant drop of uncertainty calibration performance can be found for DDU. (``MCD''=``MC Dropout'')}\label{fig:suppl_acc_vs_raulc}
\end{figure}

%% file: tables/image_classification_ood_cifar10_100.tex
\begin{table*}[!htb]
\caption{\ac{ood} detection performance when training on CIFAR10/100 and testing on various other datasets. We report \ac{auroc} averaged across 5 independent trainings.}\label{table:ood_performance_appendix}
\footnotesize
\begin{center}
\setlength{\tabcolsep}{10pt}
\begin{tabular}{c | l | c c c }
\toprule
 & OOD Data & STL10 & SVHN & CIFAR100 \\
  \midrule
 \multirow{7}{5mm}{\rotatebox{90}{\scriptsize {CIFAR10}}}
 & MC Dropout~\cite{gal2016dropout} & 0.686 $\pm$ 0.004 & 0.885 $\pm$ 0.002 & 0.82 $\pm$ 0.003 \\ 
 & Ensemble~\cite{lakshminarayanan2017simple} & \textbf{0.875 $\pm$ 0.001} & 0.937 $\pm$ 0.009 & 0.758 $\pm$ 0.003 \\ 
 & DUQ~\cite{van2020uncertainty} & 0.633 $\pm$ 0.008 & 0.843 $\pm$ 0.016 & 0.766 $\pm$ 0.003 \\ 
 & SNGP~\cite{liu2020simple} & 0.726 $\pm$ 0.007 & 0.925 $\pm$ 0.02 & 0.861 $\pm$ 0.004 \\ 
 & MIR~\cite{postels2020hidden} & 0.752 $\pm$ 0.015 & 0.916 $\pm$ 0.025 & 0.840 $\pm$ 0.007 \\ 
 & DDU~\cite{mukhoti2021deterministic} & 0.737 $\pm$ 0.018 & 0.663 $\pm$ 0.073 & 0.638 $\pm$ 0.004 \\ 
 & DCU~\cite{winkens2020contrastive} & 0.725 $\pm$ 0.027 & \textbf{0.992 $\pm$ 0.014} & \textbf{0.921 $\pm$ 0.014} \\ 
 \bottomrule
\end{tabular}

\begin{tabular}{c | l | c c c }
\toprule
 & OOD Data & STL10 & SVHN & CIFAR10 \\
  \midrule
 \multirow{7}{5mm}{\rotatebox{90}{\scriptsize {CIFAR100}}}
 & MC Dropout~\cite{gal2016dropout} & 0.772 $\pm$ 0.004 & 0.846 $\pm$ 0.01 & 0.735 $\pm$ 0.002 \\ 
 & Ensemble~\cite{lakshminarayanan2017simple} & \textbf{0.801 $\pm$ 0.014} & 0.741 $\pm$ 0.003 & 0.756 $\pm$ 0.007 \\ 
 & DUQ~\cite{van2020uncertainty} & - & - & -\\ 
 & SNGP~\cite{liu2020simple} & 0.744 $\pm$ 0.02 & 0.795 $\pm$ 0.112 & 0.686 $\pm$ 0.007 \\ 
 & MIR~\cite{postels2020hidden} & 0.789 $\pm$ 0.025 & 0.809 $\pm$ 0.031 & 0.663 $\pm$ 0.004 \\ 
 & DDU~\cite{mukhoti2021deterministic} & 0.698 $\pm$ 0.021 & 0.809 $\pm$ 0.056 & 0.\textbf{764 $\pm$ 0.019} \\ 
 & DCU~\cite{winkens2020contrastive} & 0.798 $\pm$ 0.019 & \textbf{0.978 $\pm$ 0.005} & 0.755 $\pm$ 0.024 \\ 
 \bottomrule
\end{tabular}
\end{center}
\end{table*}

%% file: figures/image_classification/supplementary_regularization_strength/mnist_regularization_strength.tex
\begin{figure}[t]
\setlength{\tabcolsep}{-2pt}
\begin{tabular}{ccc}
% \vspace{-4mm}
\input{figures/image_classification/regularization_strength/dropout_mnist_AULC_vs_ACC} &
\input{figures/image_classification/regularization_strength/dropout_mnist_fashion_mnist_AUROC_vs_ACC} &
\input{figures/image_classification/regularization_strength/dropout_mnist_omniglot_AUROC_vs_ACC} 
\\ \vspace{-1mm}
\input{figures/image_classification/regularization_strength/duq_mnist_AULC_vs_ACC} & 
\input{figures/image_classification/regularization_strength/duq_mnist_fashion_mnist_AUROC_vs_ACC} &
\input{figures/image_classification/regularization_strength/duq_mnist_omniglot_AUROC_vs_ACC} 
\\ \vspace{-1mm}
\input{figures/image_classification/regularization_strength/sngp_mnist_AULC_vs_ACC} & 
\input{figures/image_classification/regularization_strength/sngp_mnist_fashion_mnist_AUROC_vs_ACC} & 
\input{figures/image_classification/regularization_strength/sngp_mnist_omniglot_AUROC_vs_ACC}
\\ \vspace{-1mm}
\input{figures/image_classification/regularization_strength/ddu_mnist_AULC_vs_ACC} & 
\input{figures/image_classification/regularization_strength/ddu_mnist_fashion_mnist_AUROC_vs_ACC} & 
\input{figures/image_classification/regularization_strength/ddu_mnist_omniglot_AUROC_vs_ACC}
\\ \vspace{-1mm}
\input{figures/image_classification/regularization_strength/mir_mnist_AULC_vs_ACC} & 
\input{figures/image_classification/regularization_strength/mir_mnist_fashion_mnist_AUROC_vs_ACC} &
\input{figures/image_classification/regularization_strength/mir_mnist_omniglot_AUROC_vs_ACC} 
\\ \vspace{-1mm}
\end{tabular}
\vspace{-0.5em}
\caption{Trained on MNIST. Vertical axis: Test accuracy. Horizontal axis: \ac{raulc} (left), \ac{auroc} against FashionMNIST (center) and Omniglot (right) for Dropout (1st row), DUQ (2nd row), SNGP (3rd row), DDU (4th row) and MIR (5th row) using different regularization strength. For SNGP a larger hyperparameter corresponds to less regularization. For Dropout and MIR we observe a correlation between regularization strength and performance.}\label{fig:suppl_mnist_sensitivity_to_regularization_strength}
\vspace{-1em}
\end{figure}

%% file: figures/image_classification/supplementary_regularization_strength/fashion_mnist_regularization_strength.tex
\begin{figure}[t]
\setlength{\tabcolsep}{-2pt}
\begin{tabular}{ccc}
\input{figures/image_classification/regularization_strength/dropout_fashion_mnist_AULC_vs_ACC} &
\input{figures/image_classification/regularization_strength/dropout_fashion_mnist_mnist_AUROC_vs_ACC} &
\input{figures/image_classification/regularization_strength/dropout_fashion_mnist_omniglot_AUROC_vs_ACC} 
\\
\input{figures/image_classification/regularization_strength/duq_fashion_mnist_AULC_vs_ACC} & 
\input{figures/image_classification/regularization_strength/duq_fashion_mnist_mnist_AUROC_vs_ACC} &
\input{figures/image_classification/regularization_strength/duq_fashion_mnist_omniglot_AUROC_vs_ACC} 
\\
\input{figures/image_classification/regularization_strength/sngp_fashion_mnist_AULC_vs_ACC} & 
\input{figures/image_classification/regularization_strength/sngp_fashion_mnist_mnist_AUROC_vs_ACC} & 
\input{figures/image_classification/regularization_strength/sngp_fashion_mnist_omniglot_AUROC_vs_ACC}
\\
\input{figures/image_classification/regularization_strength/mir_fashion_mnist_AULC_vs_ACC} & 
\input{figures/image_classification/regularization_strength/mir_fashion_mnist_mnist_AUROC_vs_ACC} &
\input{figures/image_classification/regularization_strength/mir_fashion_mnist_omniglot_AUROC_vs_ACC} 
\\
\end{tabular}
\caption{Trained on FashionMNIST. Vertical axis: Test accuracy. Horizontal axis: \ac{raulc} (left), \ac{auroc} against MNIST (center) and Omniglot (right) for Dropout (1st row), DUQ (2nd row), SNGP (3rd row) and MIR (4th row) using different regularization strength. For SNGP a larger hyperparameter corresponds to less regularization. For Dropout and MIR we observe a correlation between regularization strength and performance.}\label{fig:suppl_fashion_mnist_sensitivity_to_regularization_strength}
\vspace{-1em}
\end{figure}

%% file: tables/runtime.tex
\begin{table*}[!htb]
\caption{Per sample runtime during training and inference on CIFAR10. The runtimes of MC dropout were obtained using 10 samples.}\label{table:runtime_appendix}
\footnotesize
\begin{center}
\setlength{\tabcolsep}{10pt}
\begin{tabular}{c | c | c }
\toprule
 Method & Training Runtime [ms] & Inference Runtime [ms] \\
  \midrule
 Softmax & 1.14 & 0.36  \\ 
 MC Dropout~\cite{gal2016dropout} & 1.13 & 5.17 \\ 
 DUQ~\cite{van2020uncertainty} & 3.68 & 0.35 \\ 
 SNGP~\cite{liu2020simple} & 2.47 & 0.44 \\ 
 DUE~\cite{mukhoti2021deterministic} & 2.26 & 0.49 \\
 MIR~\cite{postels2020hidden} & 1.34 & 0.55 \\ 
 DDU~\cite{mukhoti2021deterministic} & 2.26 & 0.49 \\ 
 \bottomrule
\end{tabular}
\end{center}
\end{table*}

%% file: figures/supplementary_segmentation_uncertainty/minimal_shift_uncertainty.tex
\begin{figure}[t]
\centering
\setlength{\tabcolsep}{3pt}
\setlength{\aboverulesep}{0pt}
\setlength{\belowrulesep}{0pt}
% \renewcommand{\arraystretch}{3}
% \resizebox{1.1\linewidth}{!}{
\begin{tabular}{cccccc}
% \vspace{-4mm}
% \hspace{-3mm}
Input & GT & & Prediction & Error & Uncertainty \\
\hline
\addlinespace[4pt]
% \\
% \\[\dimexpr-\normalbaselineskip+2pt] &
\multirow{4}{*}[-5em]{\includegraphics[width=2.5cm,trim={2.1cm 2.15cm 1.65cm 2.5cm},clip]{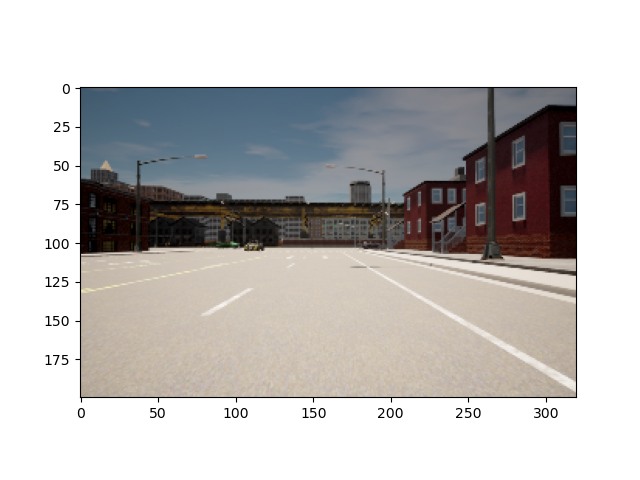}} & 
\multirow{4}{*}[-5em]{\includegraphics[width=2.5cm,trim={2.1cm 2.15cm 1.65cm 2.5cm},clip,trim={2.1cm 2.15cm 1.65cm 2.5cm},clip]{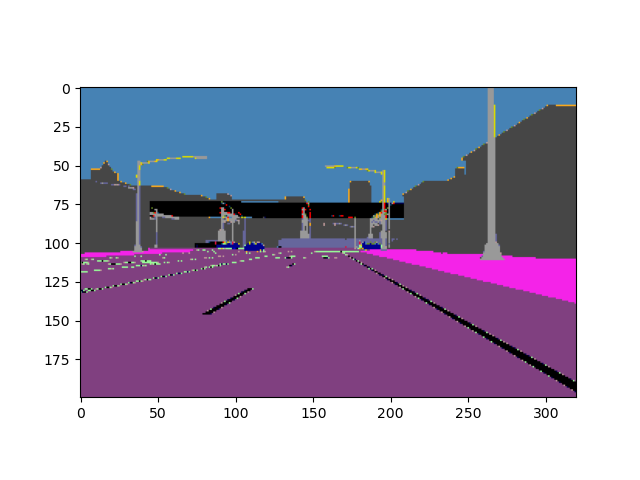}} & 
\rotatebox[origin=l]{90}{Softmax} &
\includegraphics[width=2.5cm,trim={2.1cm 2.15cm 1.65cm 2.5cm},clip]{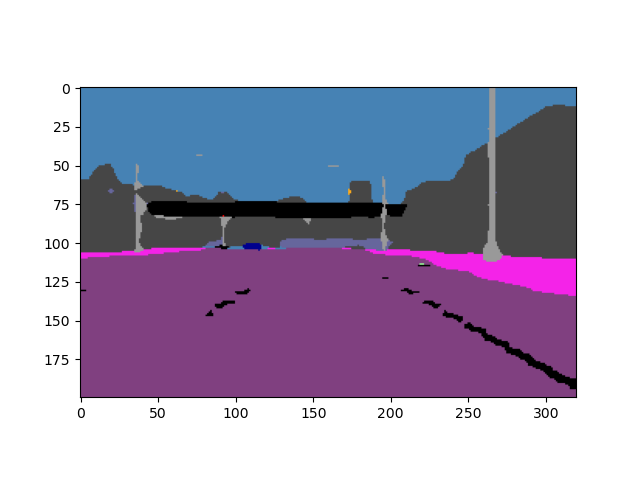} &
\includegraphics[width=2.5cm,trim={2.1cm 2.15cm 1.65cm 2.5cm},clip]{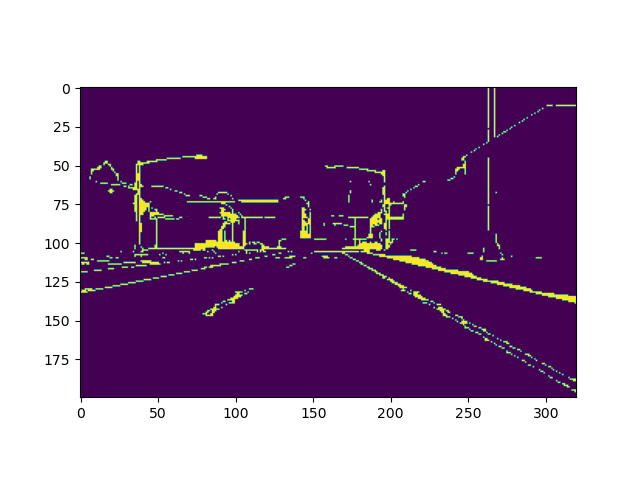} &
\includegraphics[width=2.5cm,trim={2.1cm 2.15cm 1.65cm 2.5cm},clip]{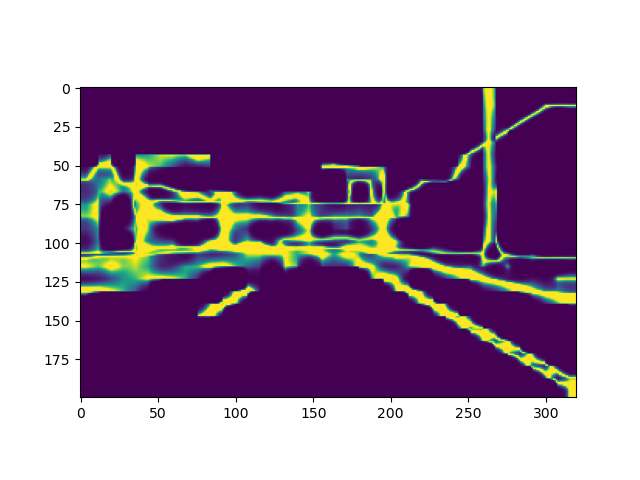} 
\\ 
& & \rotatebox[origin=l]{90}{Dropout} &
\includegraphics[width=2.5cm,trim={2.1cm 2.15cm 1.65cm 2.5cm},clip]{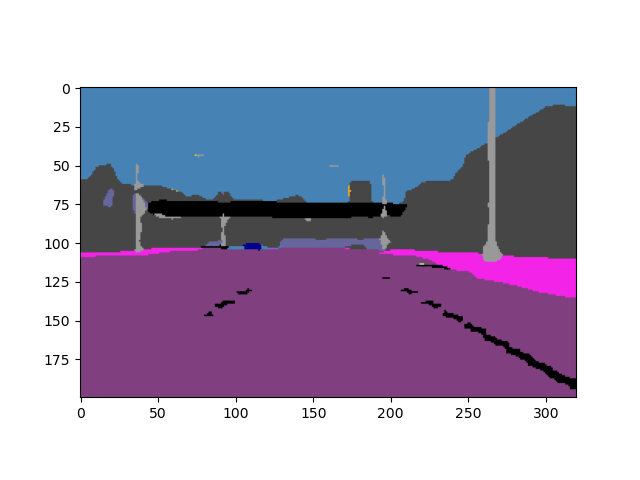} &
\includegraphics[width=2.5cm,trim={2.1cm 2.15cm 1.65cm 2.5cm},clip]{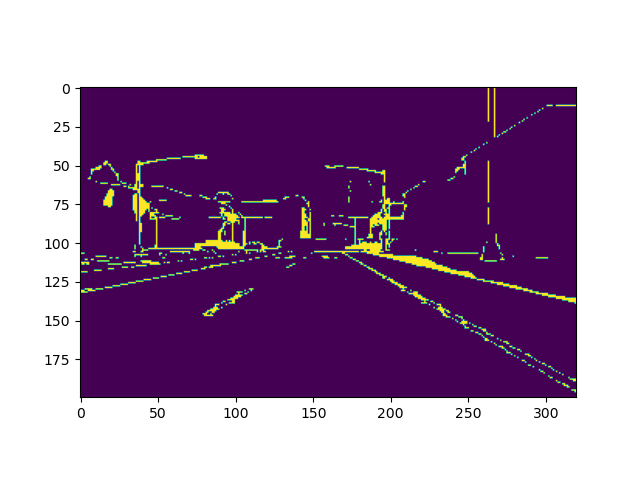} &
\includegraphics[width=2.5cm,trim={2.1cm 2.15cm 1.65cm 2.5cm},clip]{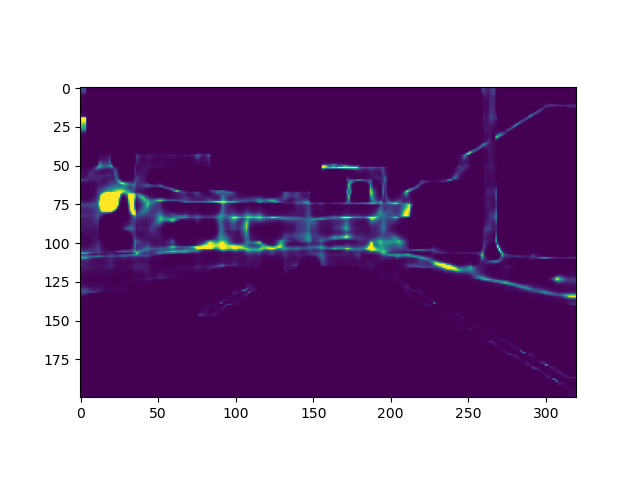} 
\\
& & \rotatebox[origin=l]{90}{SNGP} &
\includegraphics[width=2.5cm,trim={2.1cm 2.15cm 1.65cm 2.5cm},clip]{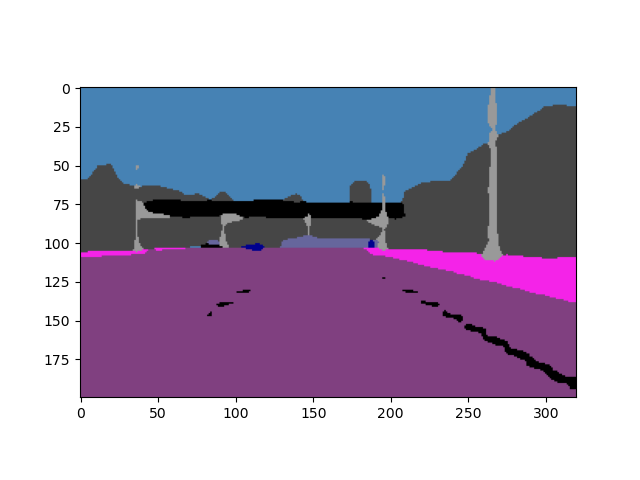} &
\includegraphics[width=2.5cm,trim={2.1cm 2.15cm 1.65cm 2.5cm},clip]{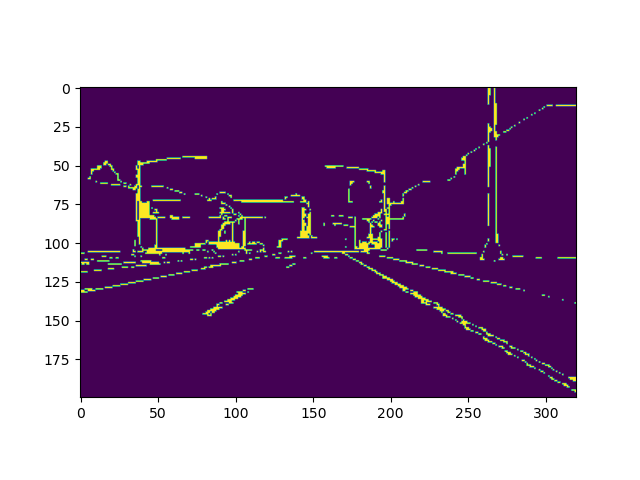} &
\includegraphics[width=2.5cm,trim={2.1cm 2.15cm 1.65cm 2.5cm},clip]{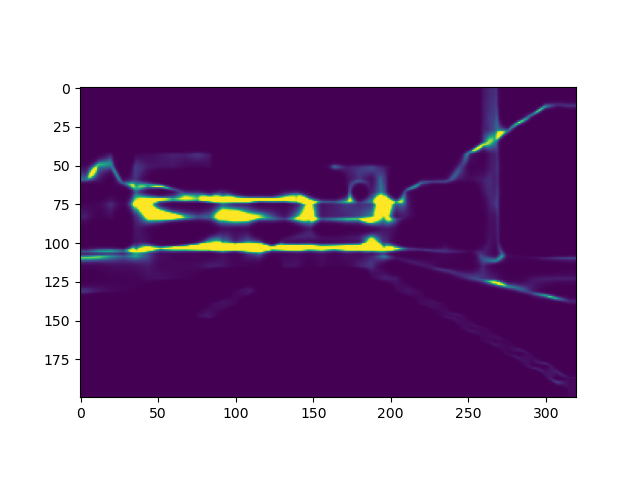} 
\\
& & \rotatebox[origin=l]{90}{MIR} &
\includegraphics[width=2.5cm,trim={2.1cm 2.15cm 1.65cm 2.5cm},clip]{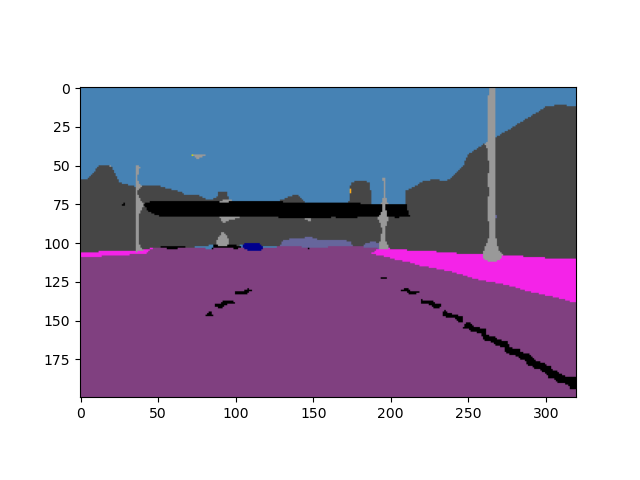} &
\includegraphics[width=2.5cm,trim={2.1cm 2.15cm 1.65cm 2.5cm},clip]{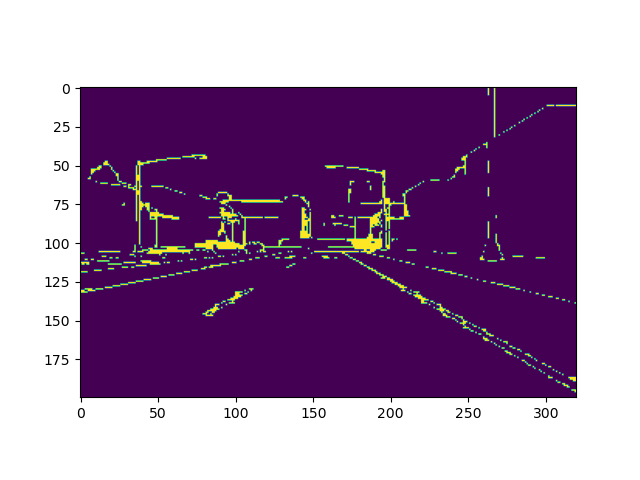} &
\includegraphics[width=2.5cm,trim={2.1cm 2.15cm 1.65cm 2.5cm},clip]{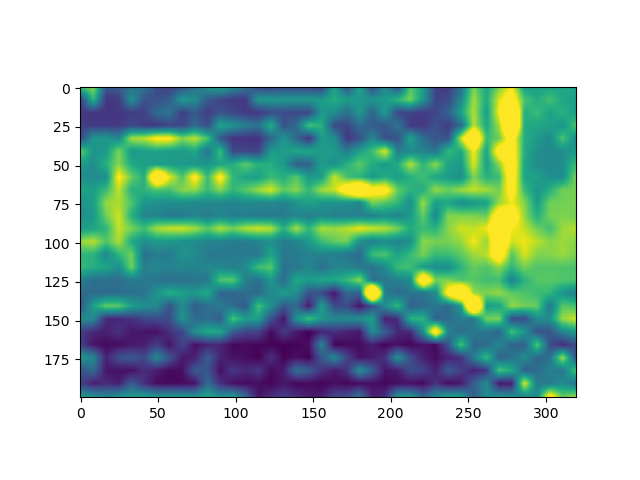} 
\\
\hline\addlinespace[4pt]
\multirow{4}{*}[-5em]{\includegraphics[width=2.5cm,trim={2.1cm 2.15cm 1.65cm 2.5cm},clip]{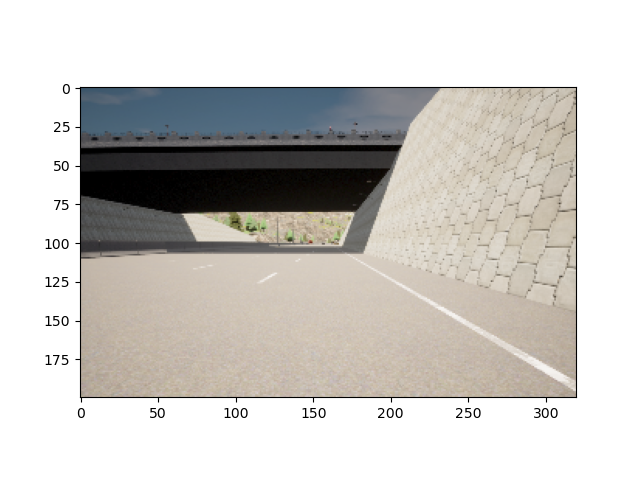}} & 
\multirow{4}{*}[-5em]{\includegraphics[width=2.5cm,trim={2.1cm 2.15cm 1.65cm 2.5cm},clip,trim={2.1cm 2.15cm 1.65cm 2.5cm},clip]{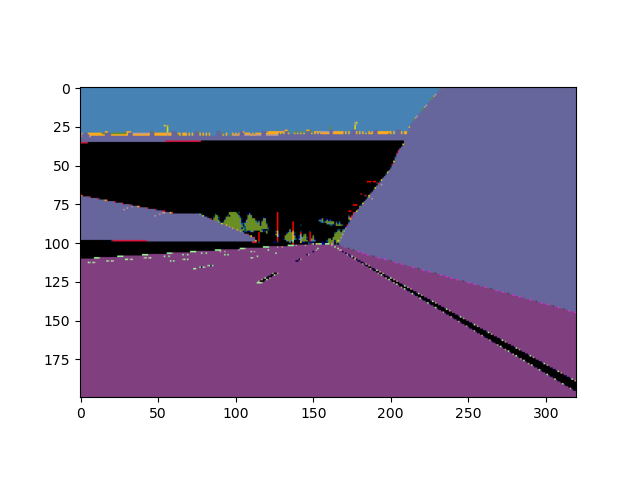}} & 
\rotatebox[origin=l]{90}{Softmax} &
\includegraphics[width=2.5cm,trim={2.1cm 2.15cm 1.65cm 2.5cm},clip]{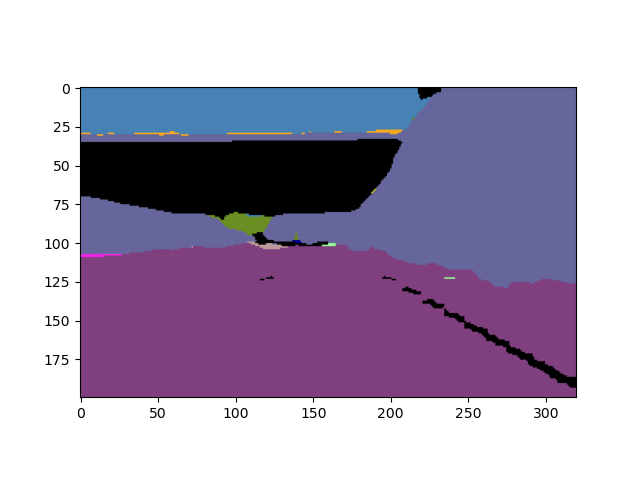} &
\includegraphics[width=2.5cm,trim={2.1cm 2.15cm 1.65cm 2.5cm},clip]{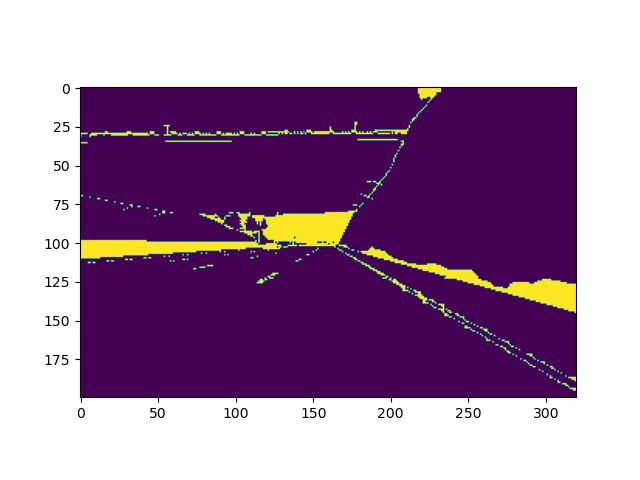} &
\includegraphics[width=2.5cm,trim={2.1cm 2.15cm 1.65cm 2.5cm},clip]{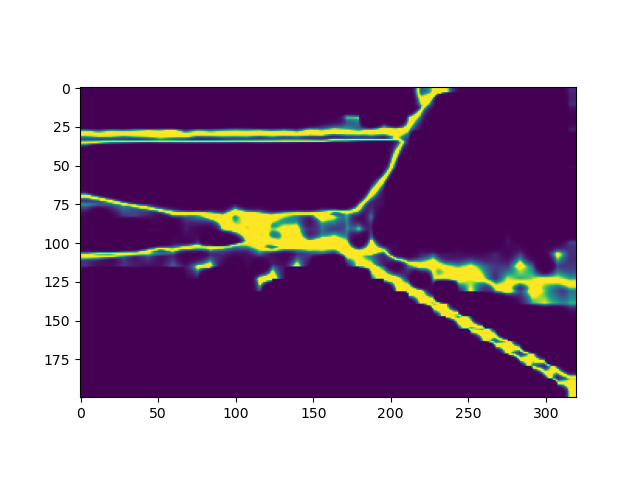} 
\\
& & \rotatebox[origin=l]{90}{Dropout} &
\includegraphics[width=2.5cm,trim={2.1cm 2.15cm 1.65cm 2.5cm},clip]{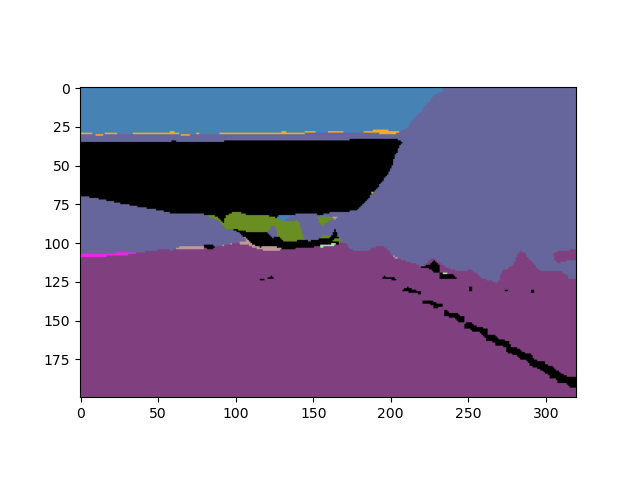} &
\includegraphics[width=2.5cm,trim={2.1cm 2.15cm 1.65cm 2.5cm},clip]{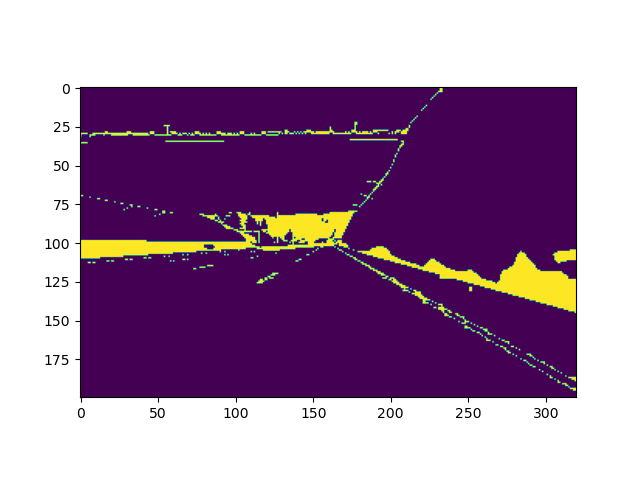} &
\includegraphics[width=2.5cm,trim={2.1cm 2.15cm 1.65cm 2.5cm},clip]{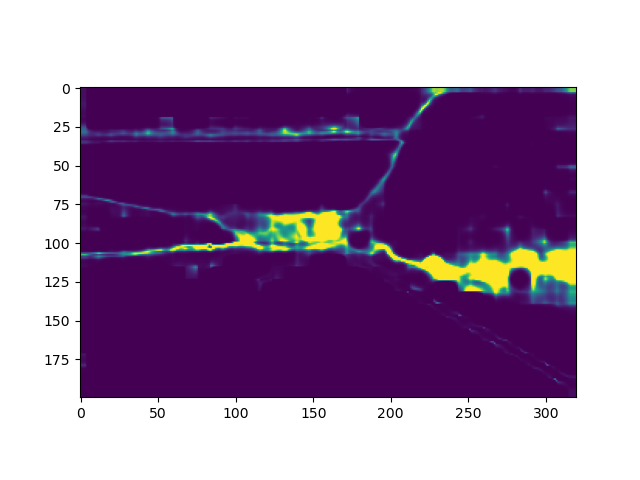} 
\\
& & \rotatebox[origin=l]{90}{SNGP} &
\includegraphics[width=2.5cm,trim={2.1cm 2.15cm 1.65cm 2.5cm},clip]{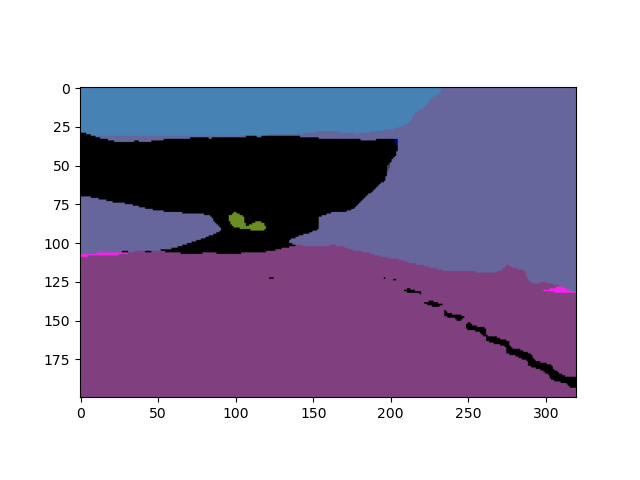} &
\includegraphics[width=2.5cm,trim={2.1cm 2.15cm 1.65cm 2.5cm},clip]{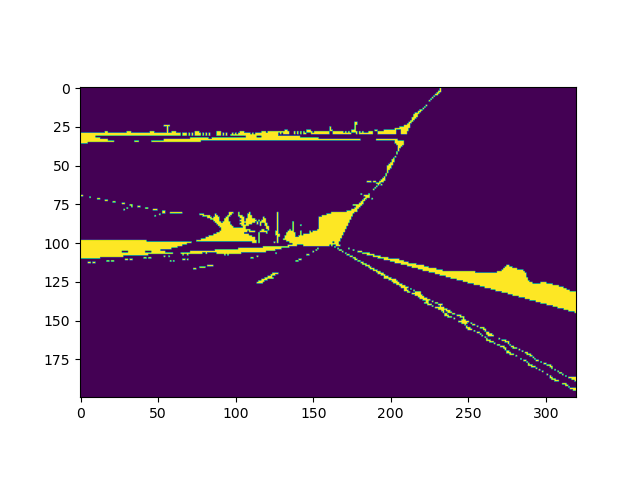} &
\includegraphics[width=2.5cm,trim={2.1cm 2.15cm 1.65cm 2.5cm},clip]{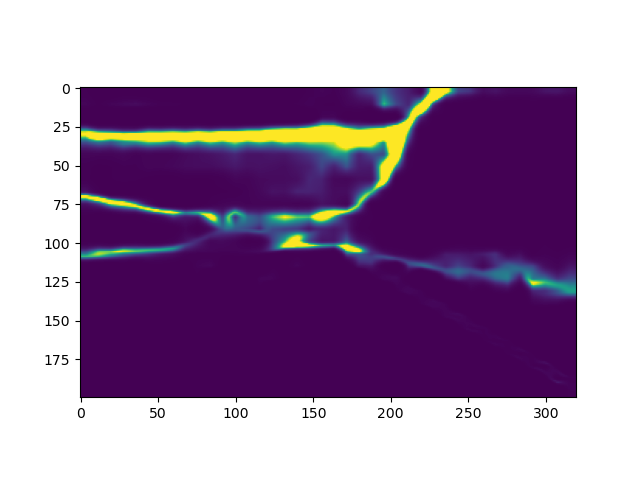} 
\\
& & \rotatebox[origin=l]{90}{MIR} &
\includegraphics[width=2.5cm,trim={2.1cm 2.15cm 1.65cm 2.5cm},clip]{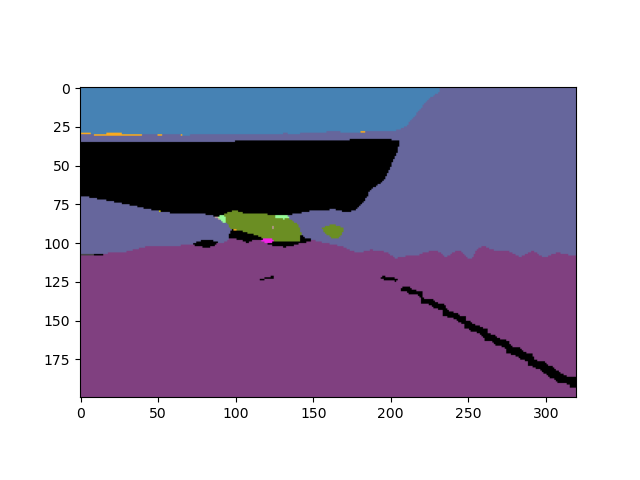} &
\includegraphics[width=2.5cm,trim={2.1cm 2.15cm 1.65cm 2.5cm},clip]{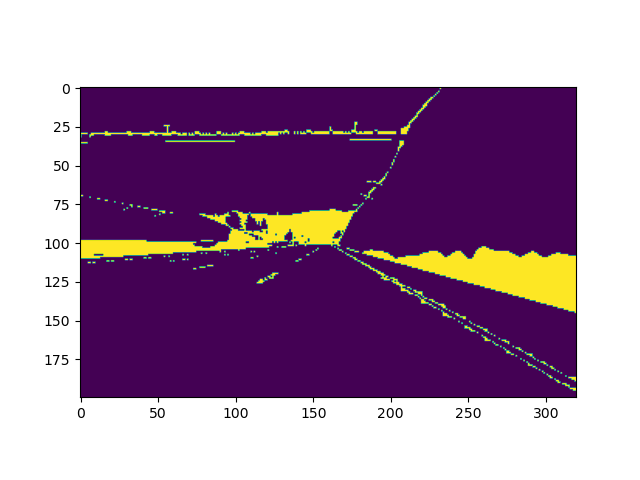} &
\includegraphics[width=2.5cm,trim={2.1cm 2.15cm 1.65cm 2.5cm},clip]{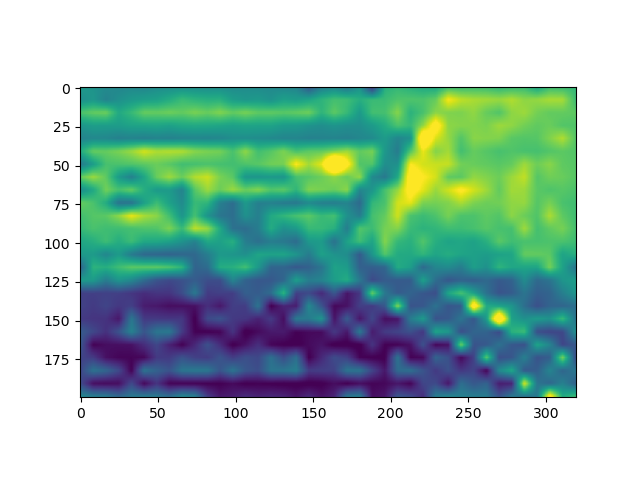} 
\\
\end{tabular}
% }
\vspace{-0.5em}
\caption{Qualitative comparison of uncertainty from Softmax, \ac{mc} Dropout, SNGP and MIR under minimal time-of-the-day distribution shift (\ie{} Azimuth angle of the sun $=85\degree$). We show the input image (Input) and the ground truth mask (GT), and we report for each method the predicted segmentation mask (Prediction), the error mask (Error) and the uncertainty mask (Uncertainty).}\label{fig:suppl_segmentation_minimal}
\vspace{-1em}
\end{figure}

%% file: figures/supplementary_segmentation_uncertainty/maximal_shift_uncertainty.tex
\begin{figure}[t]
\centering
\setlength{\tabcolsep}{3pt}
\setlength{\aboverulesep}{0pt}
\setlength{\belowrulesep}{0pt}
% \renewcommand{\arraystretch}{3}
% \resizebox{1.1\linewidth}{!}{
\begin{tabular}{cccccc}
% \vspace{-4mm}
% \hspace{-3mm}
Input & GT & & Prediction & Error & Uncertainty \\
\hline
\addlinespace[4pt]
% \\
% \\[\dimexpr-\normalbaselineskip+2pt] &
\multirow{4}{*}[-5em]{\includegraphics[width=2.5cm,trim={2.1cm 2.15cm 1.65cm 2.5cm},clip]{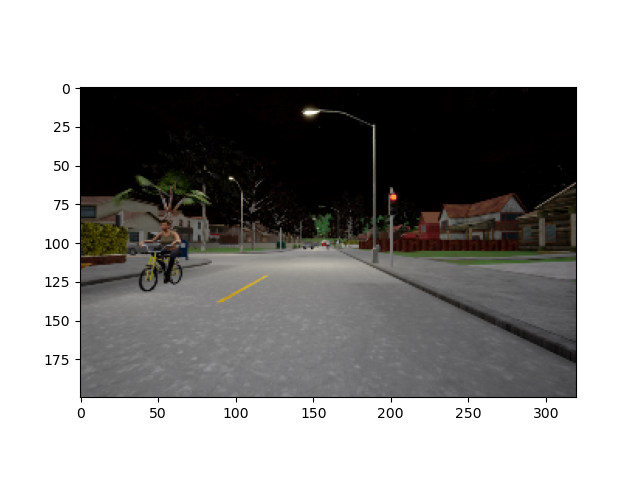}} & 
\multirow{4}{*}[-5em]{\includegraphics[width=2.5cm,trim={2.1cm 2.15cm 1.65cm 2.5cm},clip,trim={2.1cm 2.15cm 1.65cm 2.5cm},clip]{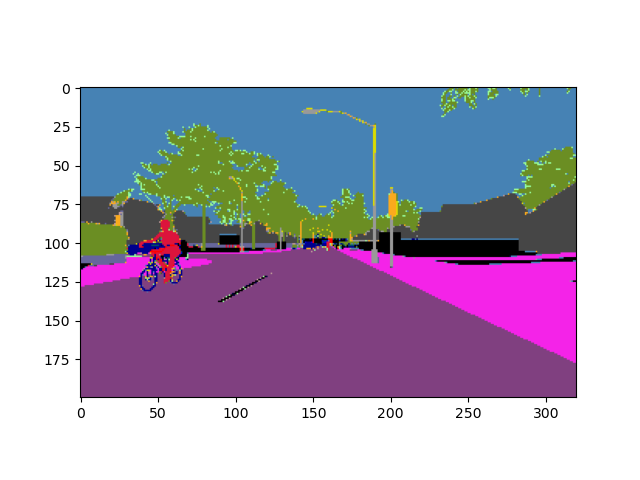}} & 
\rotatebox[origin=l]{90}{Softmax} &
\includegraphics[width=2.5cm,trim={2.1cm 2.15cm 1.65cm 2.5cm},clip]{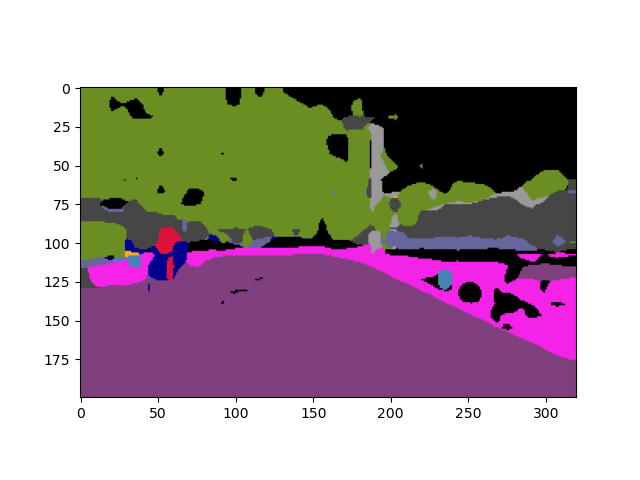} &
\includegraphics[width=2.5cm,trim={2.1cm 2.15cm 1.65cm 2.5cm},clip]{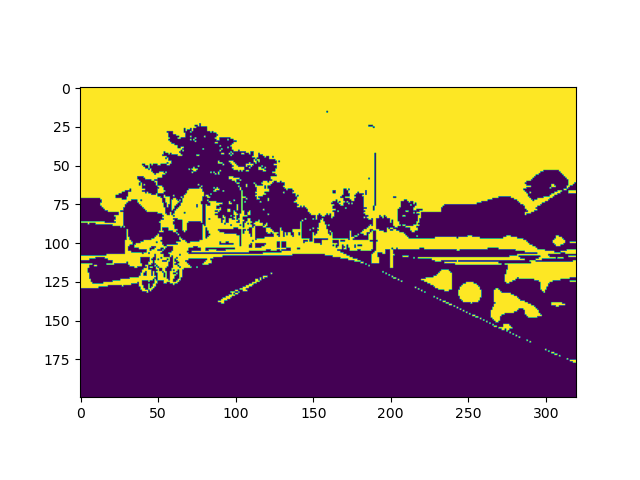} &
\includegraphics[width=2.5cm,trim={2.1cm 2.15cm 1.65cm 2.5cm},clip]{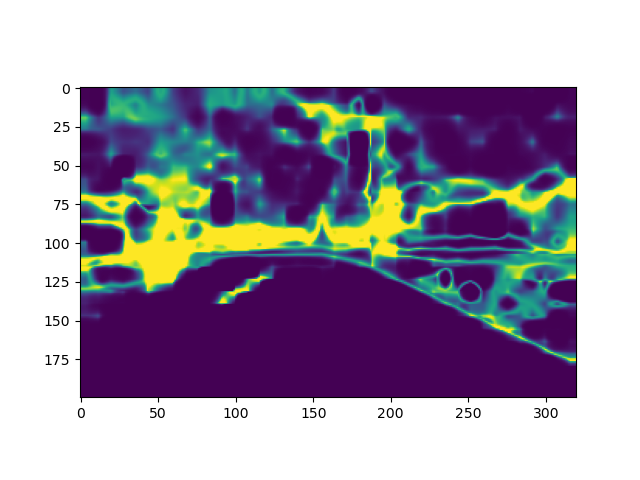} 
\\ 
& & \rotatebox[origin=l]{90}{Dropout} &
\includegraphics[width=2.5cm,trim={2.1cm 2.15cm 1.65cm 2.5cm},clip]{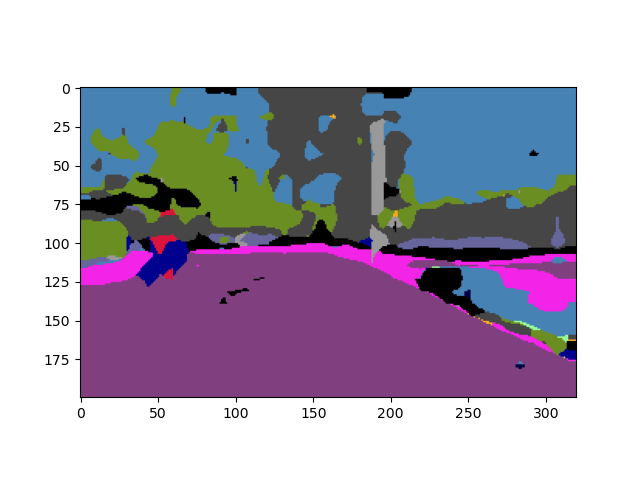} &
\includegraphics[width=2.5cm,trim={2.1cm 2.15cm 1.65cm 2.5cm},clip]{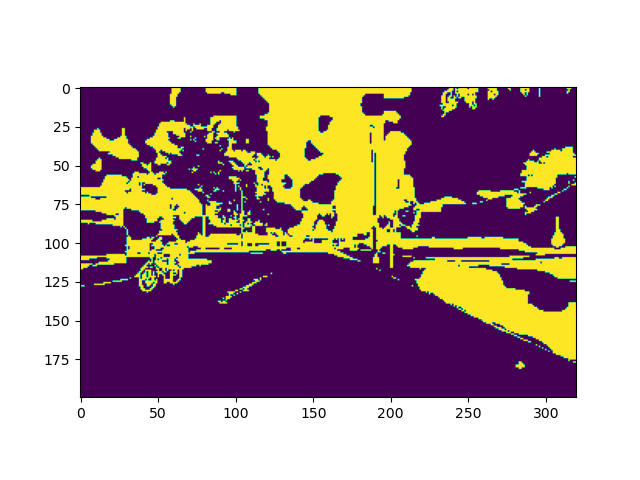} &
\includegraphics[width=2.5cm,trim={2.1cm 2.15cm 1.65cm 2.5cm},clip]{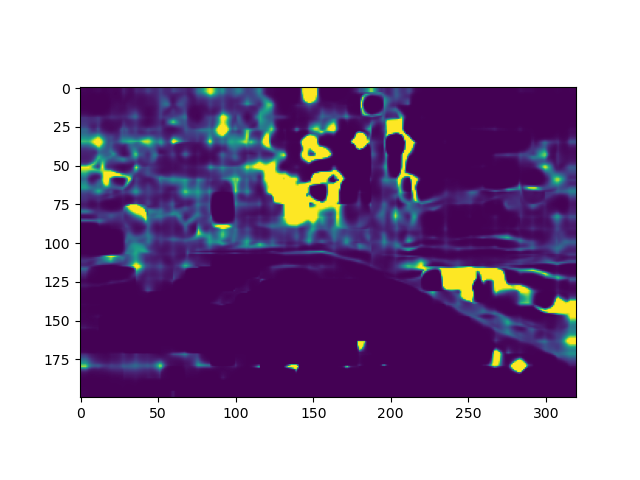} 
\\
& & \rotatebox[origin=l]{90}{SNGP} &
\includegraphics[width=2.5cm,trim={2.1cm 2.15cm 1.65cm 2.5cm},clip]{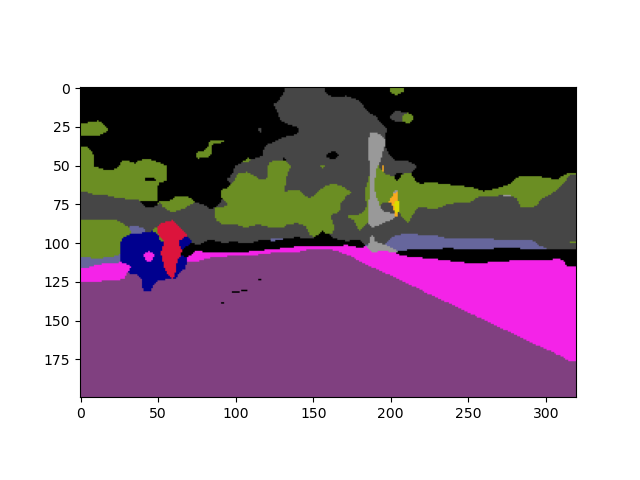} &
\includegraphics[width=2.5cm,trim={2.1cm 2.15cm 1.65cm 2.5cm},clip]{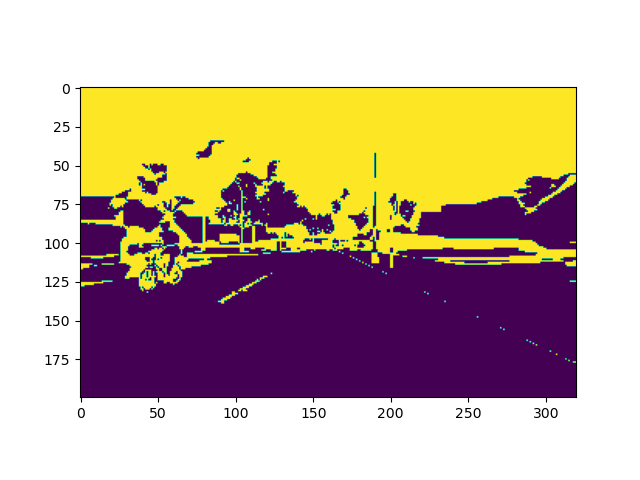} &
\includegraphics[width=2.5cm,trim={2.1cm 2.15cm 1.65cm 2.5cm},clip]{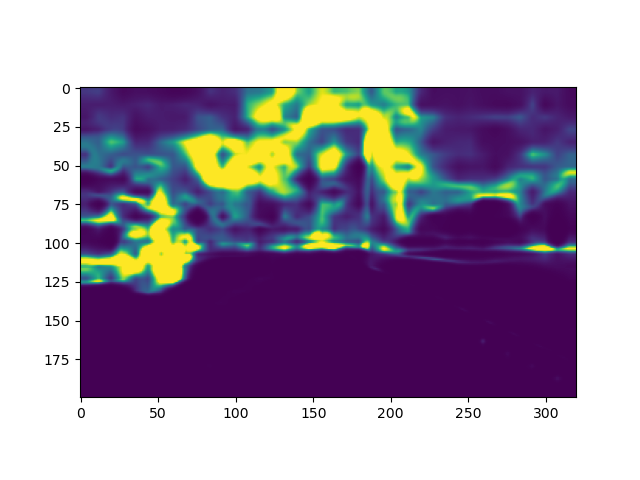} 
\\
& & \rotatebox[origin=l]{90}{MIR} &
\includegraphics[width=2.5cm,trim={2.1cm 2.15cm 1.65cm 2.5cm},clip]{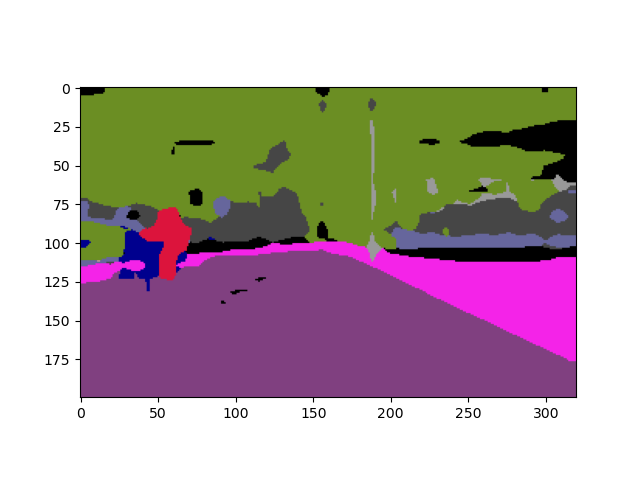} &
\includegraphics[width=2.5cm,trim={2.1cm 2.15cm 1.65cm 2.5cm},clip]{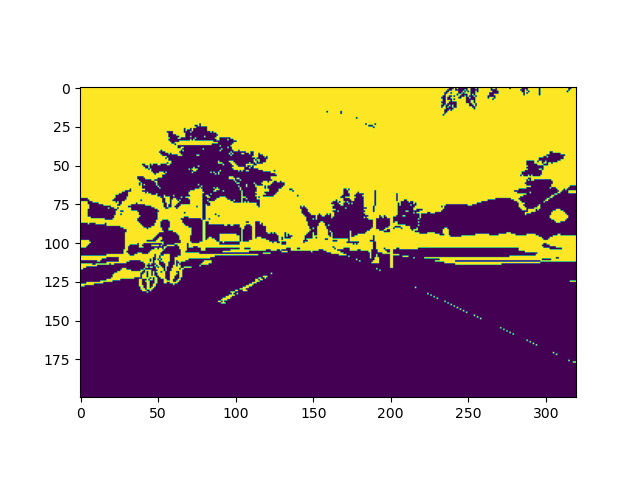} &
\includegraphics[width=2.5cm,trim={2.1cm 2.15cm 1.65cm 2.5cm},clip]{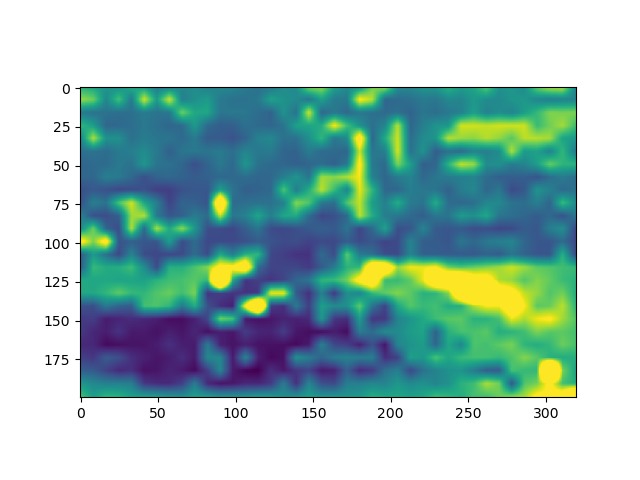} 
\\
\hline\addlinespace[4pt]
\multirow{4}{*}[-5em]{\includegraphics[width=2.5cm,trim={2.1cm 2.15cm 1.65cm 2.5cm},clip]{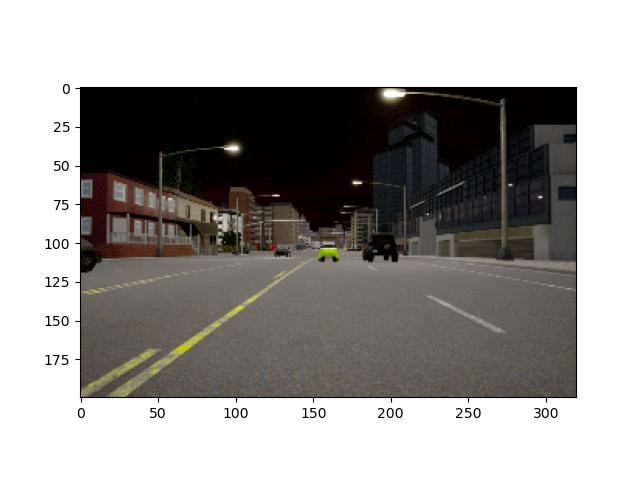}} & 
\multirow{4}{*}[-5em]{\includegraphics[width=2.5cm,trim={2.1cm 2.15cm 1.65cm 2.5cm},clip,trim={2.1cm 2.15cm 1.65cm 2.5cm},clip]{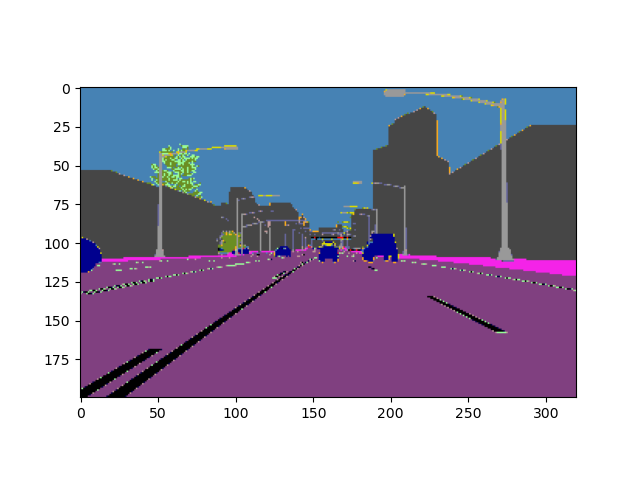}} & 
\rotatebox[origin=l]{90}{Softmax} &
\includegraphics[width=2.5cm,trim={2.1cm 2.15cm 1.65cm 2.5cm},clip]{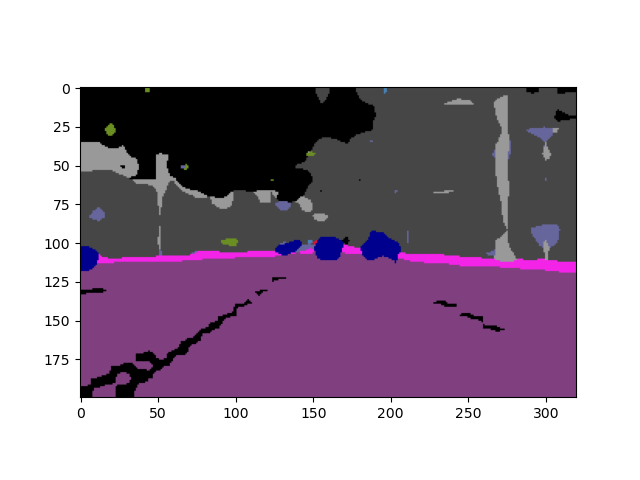} &
\includegraphics[width=2.5cm,trim={2.1cm 2.15cm 1.65cm 2.5cm},clip]{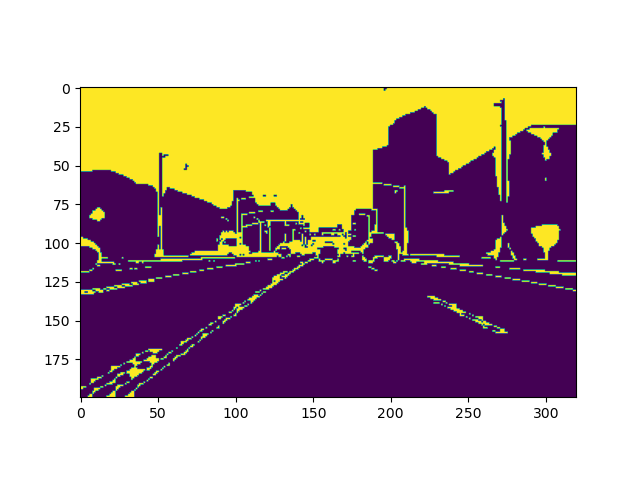} &
\includegraphics[width=2.5cm,trim={2.1cm 2.15cm 1.65cm 2.5cm},clip]{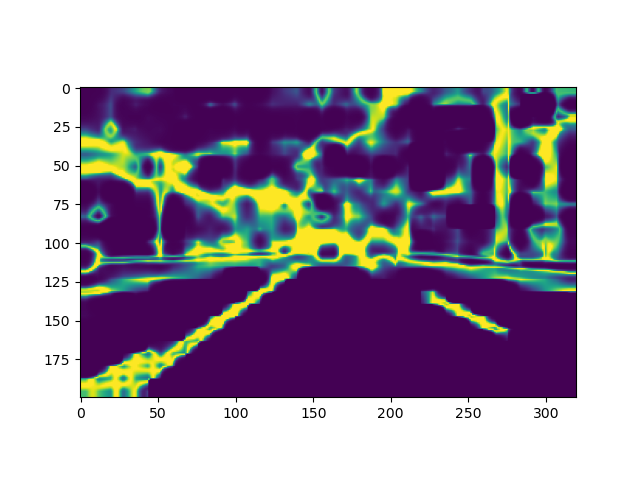} 
\\
& & \rotatebox[origin=l]{90}{Dropout} &
\includegraphics[width=2.5cm,trim={2.1cm 2.15cm 1.65cm 2.5cm},clip]{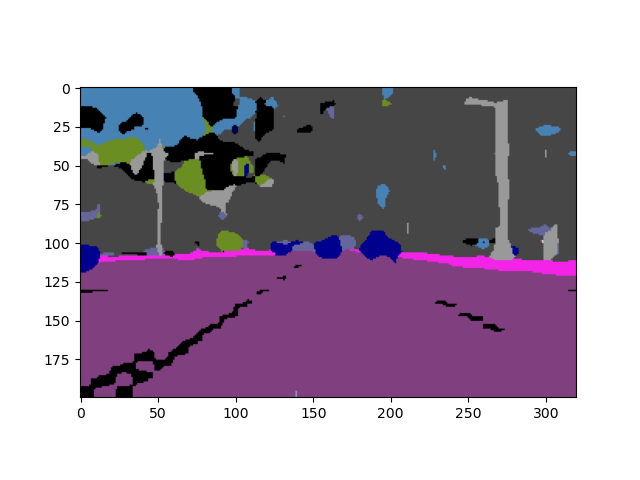} &
\includegraphics[width=2.5cm,trim={2.1cm 2.15cm 1.65cm 2.5cm},clip]{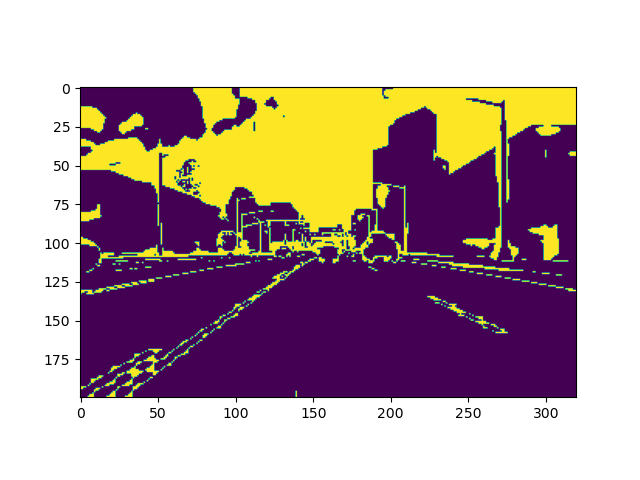} &
\includegraphics[width=2.5cm,trim={2.1cm 2.15cm 1.65cm 2.5cm},clip]{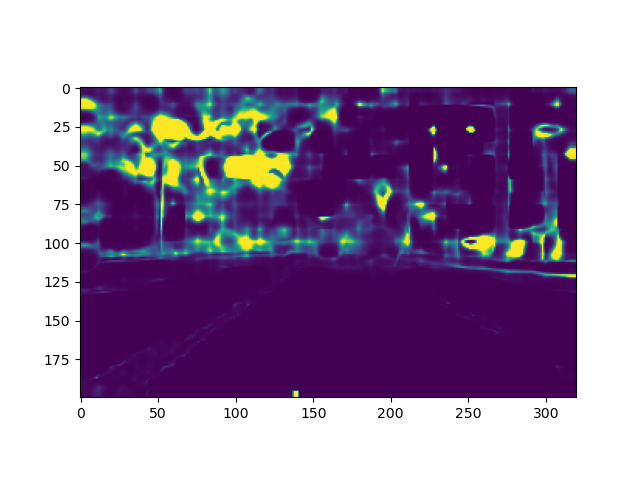} 
\\
& & \rotatebox[origin=l]{90}{SNGP} &
\includegraphics[width=2.5cm,trim={2.1cm 2.15cm 1.65cm 2.5cm},clip]{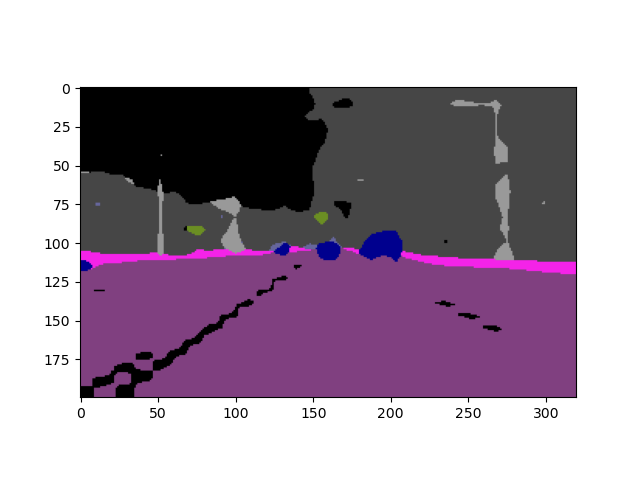} &
\includegraphics[width=2.5cm,trim={2.1cm 2.15cm 1.65cm 2.5cm},clip]{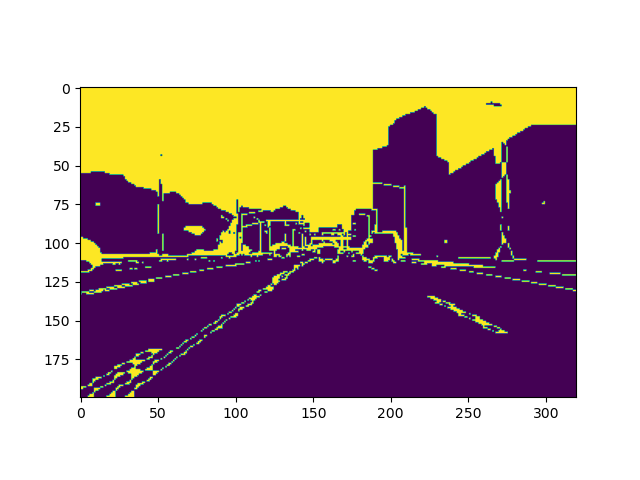} &
\includegraphics[width=2.5cm,trim={2.1cm 2.15cm 1.65cm 2.5cm},clip]{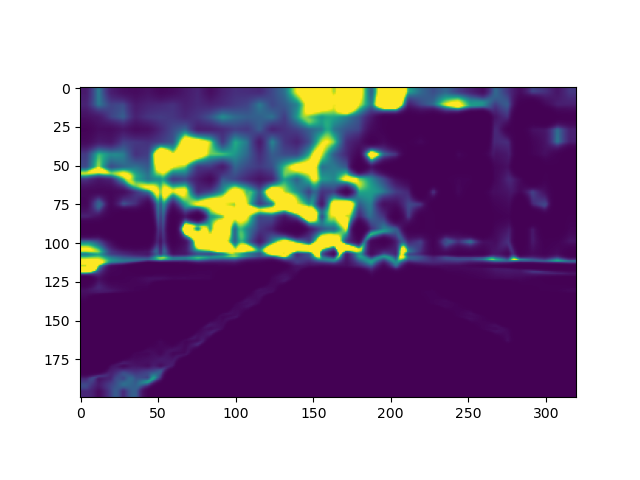} 
\\
& & \rotatebox[origin=l]{90}{MIR} &
\includegraphics[width=2.5cm,trim={2.1cm 2.15cm 1.65cm 2.5cm},clip]{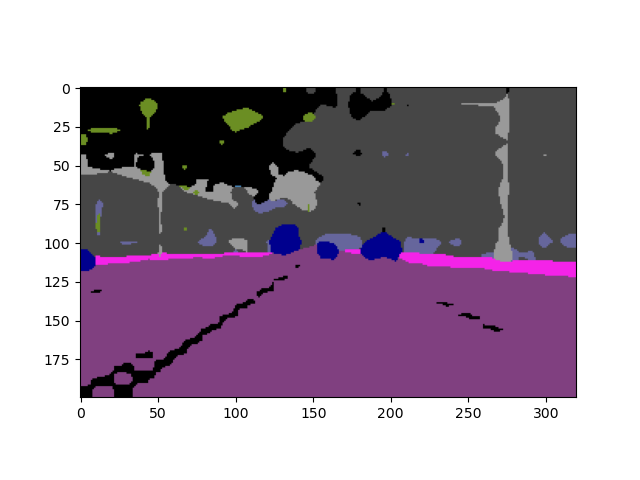} &
\includegraphics[width=2.5cm,trim={2.1cm 2.15cm 1.65cm 2.5cm},clip]{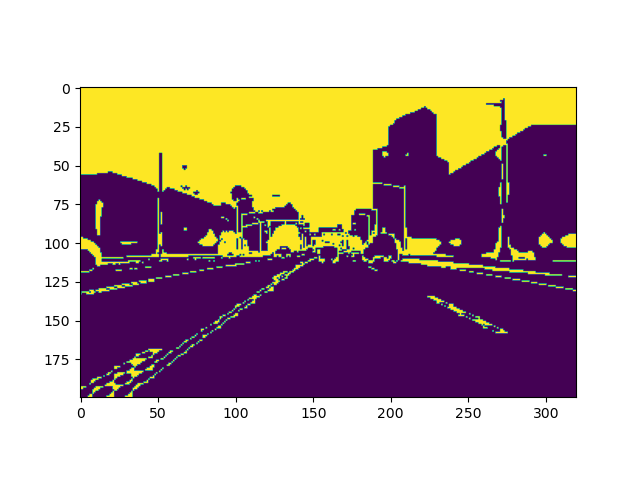} &
\includegraphics[width=2.5cm,trim={2.1cm 2.15cm 1.65cm 2.5cm},clip]{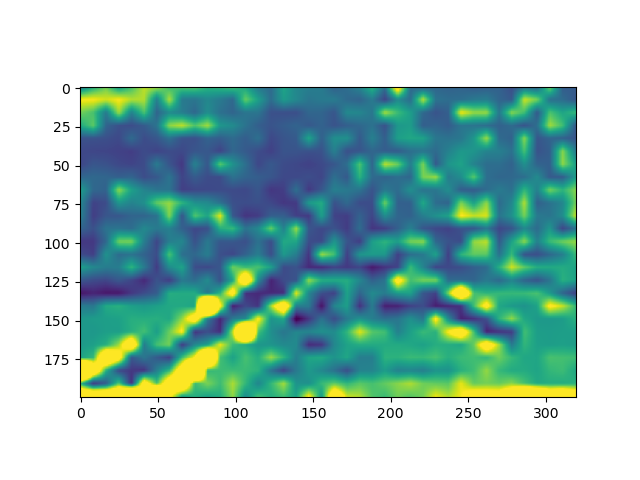} 
\\
\end{tabular}
% }
\vspace{-0.5em}
\caption{Qualitative comparison of uncertainty from Softmax, \ac{mc} Dropout, SNGP and MIR under maximal time-of-the-day distribution shift (\ie{} Azimuth angle of the sun $=-5\degree$). We show the input image (Input) and the ground truth mask (GT), and we report for each method the predicted segmentation mask (Prediction), the error mask (Error) and the uncertainty mask (Uncertainty).}\label{fig:suppl_segmentation_maximal}
\vspace{-1em}
\end{figure}

%% file: figures/supplementary_corrupted/cifar10_c.tex
\begin{figure*}[!h]
% \centering
\input{figures/cifar10_c/brightness} \\
\input{figures/cifar10_c/contrast} \\
\input{figures/cifar10_c/defocus_blur} \\
\input{figures/cifar10_c/elastic} \\
\input{figures/cifar10_c/fog} \\
\caption{We here compare the performance of \acp{dum} and of the baselines under different corruption types and severities applied on the CIFAR10-C dataset. We show the accuracy, \ac{auroc} and \ac{raulc} (vertical axis) for each method depending on the corruption severity (horizontal axis) of the following corruption types (listed from top to bottom): brightness, contrast, defocus blur, elastic, fog.}
\label{fig:corruptions_cifar_10_c}
\vspace{-2mm}
\end{figure*}

\begin{figure*}[!h]
% \centering
\input{figures/cifar10_c/frost} \\
\input{figures/cifar10_c/frosted_glass_blur} \\
\input{figures/cifar10_c/gaussian_noise} \\
\input{figures/cifar10_c/impulse_noise} \\
\input{figures/cifar10_c/jpeg_compression} \\
\caption{We here compare the performance of \acp{dum} and of the baselines under different corruption types and severities applied on the CIFAR10-C dataset. We show the accuracy, \ac{auroc} and \ac{raulc} (vertical axis) for each method depending on the corruption severity (horizontal axis) of the following corruption types (listed from top to bottom): frost, frosted glass blur, gaussian noise, impulse noise, jpeg compression.}
\label{fig:corruptions_cifar_10_c_2}
\vspace{-2mm}
\end{figure*}

\begin{figure*}[!h]
% \centering
\input{figures/cifar10_c/motion_blur} \\
\input{figures/cifar10_c/pixelate} \\
\input{figures/cifar10_c/shot_noise} \\
\input{figures/cifar10_c/snow} \\
\input{figures/cifar10_c/zoom_blur} \\
\caption{We here compare the performance of \acp{dum} and of the baselines under different corruption types and severities applied on the CIFAR10-C dataset. We show the accuracy, \ac{auroc} and \ac{raulc} (vertical axis) for each method depending on the corruption severity (horizontal axis) of the following corruption types (listed from top to bottom): motion blur, pixelate, shot noise, snow, zoom blur.}
\label{fig:corruptions_cifar_10_c_3}
\vspace{-2mm}
\end{figure*}

%% file: figures/supplementary_corrupted/cifar100_c.tex
\begin{figure*}[!h]
% \centering
\input{figures/cifar100_c/brightness} \\
\input{figures/cifar100_c/contrast} \\
\input{figures/cifar100_c/defocus_blur} \\
\input{figures/cifar100_c/elastic} \\
\input{figures/cifar100_c/fog} \\
\caption{We here compare the performance of \acp{dum} and of the baselines under different corruption types and severities applied on the CIFAR100-C dataset. We show the accuracy, \ac{auroc} and \ac{raulc} (vertical axis) for each method depending on the corruption severity (horizontal axis) of the following corruption types (listed from top to bottom): brightness, contrast, defocus blur, elastic, fog.}
\label{fig:corruptions_cifar_100_c}
\vspace{-2mm}
\end{figure*}

\begin{figure*}[!h]
% \centering
\input{figures/cifar100_c/frost} \\
\input{figures/cifar100_c/frosted_glass_blur} \\
\input{figures/cifar100_c/gaussian_noise} \\
\input{figures/cifar100_c/impulse_noise} \\
\input{figures/cifar100_c/jpeg_compression} \\
\caption{We here compare the performance of \acp{dum} and of the baselines under different corruption types and severities applied on the CIFAR100-C dataset. We show the accuracy, \ac{auroc} and \ac{raulc} (vertical axis) for each method depending on the corruption severity (horizontal axis) of the following corruption types (listed from top to bottom): frost, frosted glass blur, gaussian noise, impulse noise, jpeg compression.}
\label{fig:corruptions_cifar_100_c_2}
\vspace{-2mm}
\end{figure*}

\begin{figure*}[!h]
% \centering
\input{figures/cifar100_c/motion_blur} \\
\input{figures/cifar100_c/pixelate} \\
\input{figures/cifar100_c/shot_noise} \\
\input{figures/cifar100_c/snow} \\
\input{figures/cifar100_c/zoom_blur} \\
\caption{We here compare the performance of \acp{dum} and of the baselines under different corruption types and severities applied on the CIFAR100-C dataset. We show the accuracy, \ac{auroc} and \ac{raulc} (vertical axis) for each method depending on the corruption severity (horizontal axis) of the following corruption types (listed from top to bottom): motion blur, pixelate, shot noise, snow, zoom blur.}
\label{fig:corruptions_cifar_100_c_3}
\vspace{-2mm}
\end{figure*}

%% file: figures/supplementary_corrupted/cityscapes_c.tex
\begin{figure*}[!h]
% \centering
\input{figures/cityscapes_c/brightness} \\
\input{figures/cityscapes_c/contrast} \\
\input{figures/cityscapes_c/defocus_blur} \\
\input{figures/cityscapes_c/elastic} \\
\input{figures/cityscapes_c/fog} \\
\caption{We here compare the performance of \acp{dum} and of the baselines under different corruption types and severities applied on the CIFAR100-C dataset. We show the \ac{miou}, \ac{auroc} and \ac{raulc} (vertical axis) for each method depending on the corruption severity (horizontal axis) of the following corruption types (listed from top to bottom): brightness, contrast, defocus blur, elastic, fog.}
\label{fig:corruptions_cityscapes_c}
\vspace{-2mm}
\end{figure*}

\begin{figure*}[!h]
% \centering
\input{figures/cityscapes_c/frost} \\
\input{figures/cityscapes_c/frosted_glass_blur} \\
\input{figures/cityscapes_c/gaussian_noise} \\
\input{figures/cityscapes_c/impulse_noise} \\
\input{figures/cityscapes_c/jpeg_compression} \\
\caption{We here compare the performance of \acp{dum} and of the baselines under different corruption types and severities applied on the CIFAR100-C dataset. We show the \ac{miou}, \ac{auroc} and \ac{raulc} (vertical axis) for each method depending on the corruption severity (horizontal axis) of the following corruption types (listed from top to bottom): frost, frosted glass blur, gaussian noise, impulse noise, jpeg compression.}
\label{fig:corruptions_cityscapes_c_2}
\vspace{-2mm}
\end{figure*}

\begin{figure*}[!h]
% \centering
\input{figures/cityscapes_c/motion_blur} \\
\input{figures/cityscapes_c/pixelate} \\
\input{figures/cityscapes_c/shot_noise} \\
\input{figures/cityscapes_c/snow} \\
\input{figures/cityscapes_c/zoom_blur} \\
\caption{We here compare the performance of \acp{dum} and of the baselines under different corruption types and severities applied on the Cityscapes-C dataset. We show the \ac{miou}, \ac{auroc} and \ac{raulc} (vertical axis) for each method depending on the corruption severity (horizontal axis) of the following corruption types (listed from top to bottom): motion blur, pixelate, shot noise, snow, zoom blur.}
\label{fig:corruptions_cityscapes_c_3}
\vspace{-2mm}
\end{figure*}

%% file: figures/supplementary_corrupted/carla_c.tex
\begin{figure*}[!h]
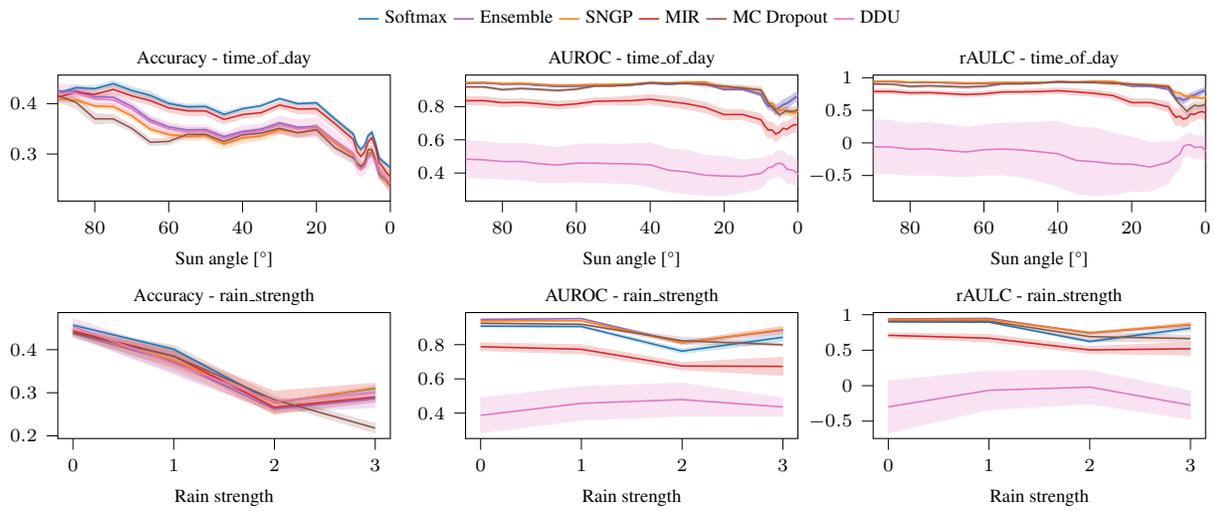

% \centering
\input{figures/carla_c/time_of_day} \\
\input{figures/carla_c/rain_strength}%
\caption{We here compare the performance of \acp{dum} and of the baselines under different corruption types and severities applied on the Carla-C dataset. We show the \ac{miou}, \ac{auroc} and \ac{raulc} (vertical axis) for each method depending on the corruption severity (horizontal axis) of the following corruption types (listed from top to bottom): time of day, rain.}
\label{fig:corruptions_carla_c}
\vspace{-2mm}
\end{figure*}